\documentclass[runningheads]{llncs}

\usepackage{eccv}

\usepackage{eccvabbrv}

\usepackage{graphicx}
\usepackage{booktabs}

\usepackage[accsupp]{axessibility}  

\usepackage{hyperref}

\usepackage{orcidlink}

\usepackage{graphicx}
\usepackage{duckuments}
\usepackage{cuted}
\usepackage{tabularray}
\usepackage{rotating}
\usepackage[table]{xcolor} 
\usepackage{multirow} 
\usepackage{array} 
\usepackage{bm}
\usepackage{xcolor}
\usepackage{marvosym}

\newcommand{\blue}[1]{\textcolor{blue}{{#1}}}

\newcommand{\magenta}[1]{\textcolor{magenta}{{#1}}}
\newcommand{\orange}[1]{\textcolor{orange}{{#1}}}

\newcommand{\titlecolor}[1]{\gradientRGB{#1}{188, 67, 186}{0, 99, 178}}
\usepackage{gradient-text}

\newcommand{\tc}[1]{\textcolor{blue!70!black}{#1}}
\newcommand{\td}[1]{\textcolor{orange!85!black}{#1}}
\newcommand{\toh}[1]{\textcolor{gray!80!black}{#1}}
\newcommand{\setdeg}[1]{\textcolor{purple!80!black}{\textbf{#1}}}
\newcommand{\setpaca}[1]{\textcolor{teal!70!black}{\textbf{#1}}}
\newcommand{\setcap}[1]{\textcolor{blue!75!black}{\textbf{#1}}}
\newcommand{\setstage}[1]{\textcolor{red!70!black}{\textbf{#1}}}

\begin{document}

\title{\titlecolor{MagnifiQ}: Patch-aware Text Guided Progressive Upscaling for High-Resolution Image Restoration}

\titlerunning{MagnifiQ: Patch-Aware Progressive Upscaling for Image Restoration}


\author{ Mahesh Reddy\textsuperscript{\Letter} \and Yashesh Savani \and Antoine Mercier \and Hong Cai \and Fatih Porikli \and Guillaume Berger\textsuperscript{\Letter} }


\authorrunning{M.~Reddy et al.}

\institute{ Qualcomm AI Research\thanks{Qualcomm AI Research is an initiative of Qualcomm Technologies, Inc.}\\ \email{\{mahkri,guilberg\}@qti.qualcomm.com} }

\maketitle

\begin{figure*}
  \centering
  \includegraphics[width=0.8\linewidth]{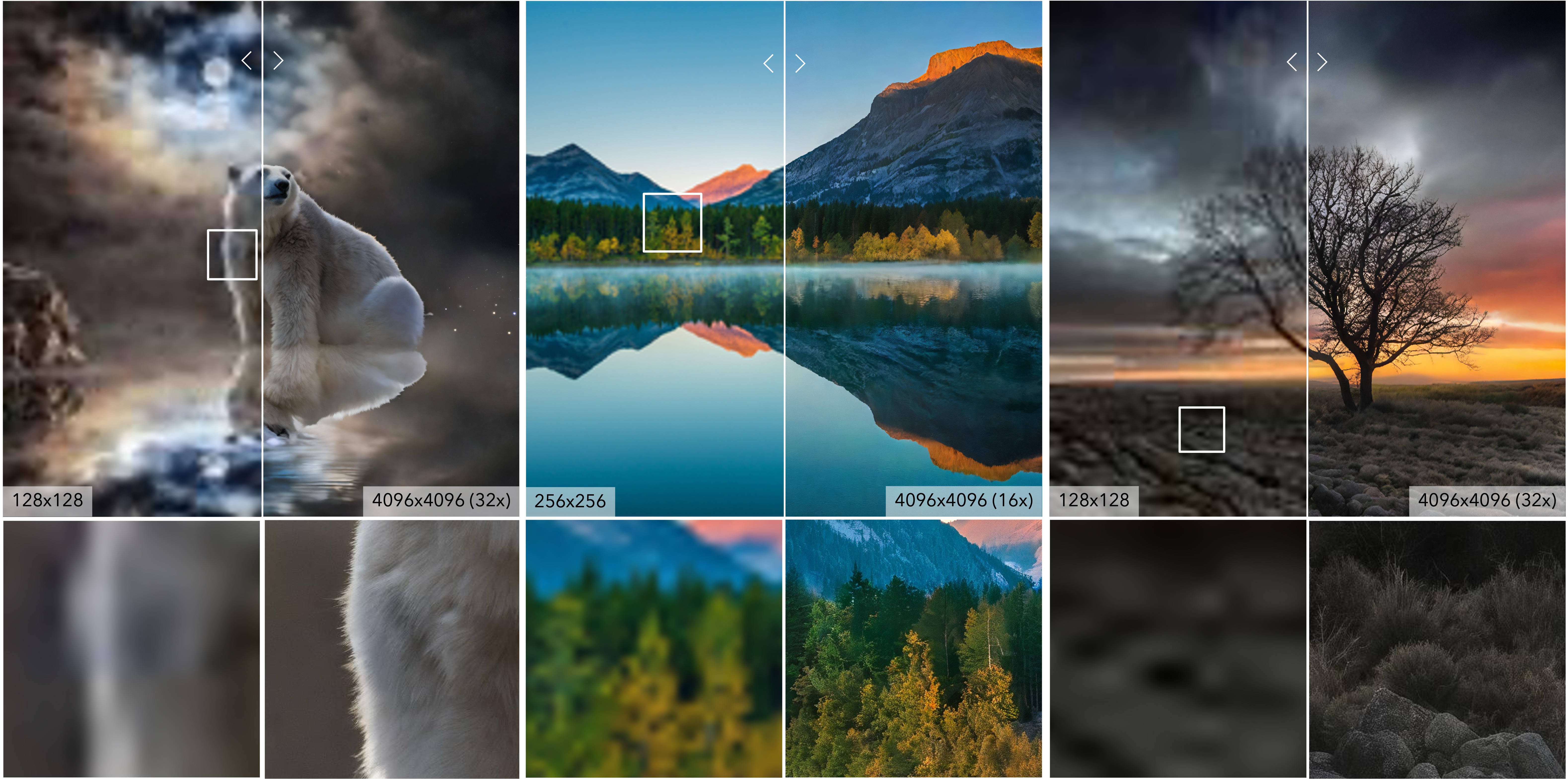}
   \caption{\textbf{MagnifiQ} restores real-world low-quality inputs into coherent high-resolution outputs up to $4096 \times 4096$, supporting up to $32\times$ scaling while preserving global structure and recovering fine details. \textbf{Zoom in for details.}}
  \label{fig:teaser}
  \vspace{-0.75cm}
\end{figure*}


\begin{abstract}
High-resolution image restoration from degraded inputs is challenging because it must preserve global structural consistency while recovering fine-grained local details, especially at 4K resolution where direct diffusion-based restoration is computationally expensive and prone to repeated or inconsistent textures. In this work, we introduce \textbf{MagnifiQ}, an image restoration framework that progressively upscales and restores images across resolutions, e.g., from $1024 \times 1024$ to $4096 \times 4096$. Our approach leverages a pre-trained text-to-image diffusion model such as SDXL and adapts it for more scalable high-resolution inference by replacing its original self-attention layers with convolutional operations whose computational cost grows linearly with image resolution. We further propose a progressive upscaling strategy that iteratively restores images over multiple
resolution stages, refining each intermediate output rather than directly hallucinating the final 4K image, thereby improving global coherence and reducing high-resolution artifacts. To enhance local details while controlling content drift, MagnifiQ uses patch-specific text prompts that provide spatially localized semantic guidance during restoration. Extensive experiments on synthetic and real-world degraded images show that MagnifiQ outperforms prior diffusion-based restoration methods in perceptual quality and human preference, producing sharper textures and more coherent 4K results while offering practical speed--quality trade-offs through its scalable backbone and progressive design.
\end{abstract}    
\section{Introduction}
\label{sec:intro}

Real-world restoration of single low-quality images is a challenging problem due to the inherent difficulty of recovering lost details and textures from unknown degradations. Recent approaches~\cite{wang2024exploiting,wu2024seesr,yang2024pixel,yu2024scaling,lin2024diffbir} leverage the strong generative capabilities of large-scale pre-trained text-to-image diffusion models, enabling the synthesis of restored images with superior perceptual quality and realistic content compared to GAN-based counterparts~\cite{wang2021real,liang2021swinir}. 

Despite these improvements, most existing approaches~\cite{wu2024seesr,yang2024pixel,yu2024scaling,lin2024diffbir} are constrained to moderate upscaling factors, typically up to $4\times$. This limitation restricts their applicability in real-world scenarios such as recovering image quality lost due to heavy zooming, close-up cropping, or low-resolution capture. StableSR~\cite{wang2024exploiting} introduces a pipeline for 4K high-resolution restoration based on Mixture-of-Diffusers~\cite{jimenez2023mixture}. While effective, this approach is computationally expensive because it denoises each image patch independently at every timestep. Furthermore, although the generated images maintain global structural consistency, they often lack fine-grained local details due to the absence of text guidance, which is critical for high-resolution restoration.

\begin{figure}[!t]
    \centering
    \includegraphics[width=0.9\linewidth]{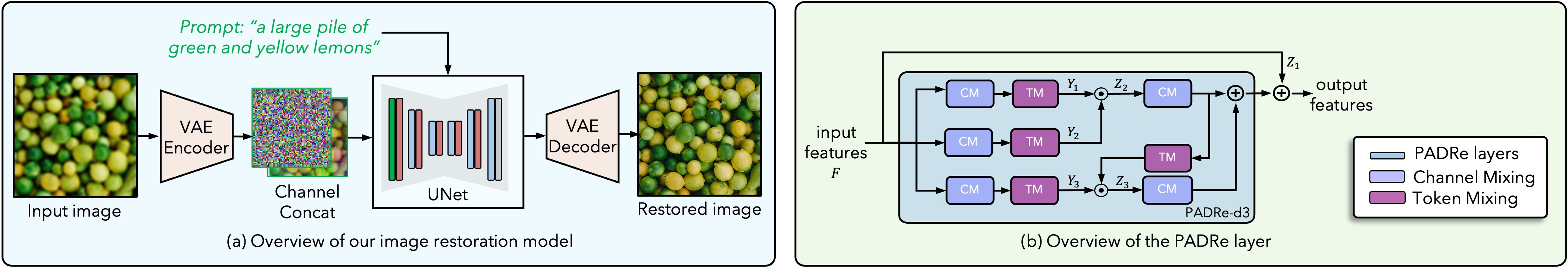}
    \captionof{figure}{Overview of our restoration backbone. (a) We adapt pre-trained SDXL~\cite{podell2023sdxl} for image restoration by modifying the input convolution, following InstructPix2Pix~\cite{brooks2023instructpix2pix}, to condition on the low-quality input image. (b) We replace self-attention layers in the SDXL U-Net with efficient PADRe layers, improving high-resolution efficiency and restoration quality.}

    \label{fig:padre_block}
    \vspace{-0.5cm}
\end{figure}

To address these challenges, we propose $\textbf{MagnifiQ}$, a patch-aware, text-guided progressive upscaling framework that iteratively restores low-quality images to high resolution $4096 \times 4096$ with scaling factors of up to $32\times$, as shown in Fig.~\ref{fig:teaser}. MagnifiQ ensures coherent global structure by progressively conditioning the restoration model on intermediate outputs, while enhancing local details through patch-level text prompts generated from intermediate images. Specifically, we introduce Patch-Aware Cross-Attention (PACA), a lightweight modification of the cross-attention layers in a pre-trained restoration model, enabling cross-attention between latent patches and their corresponding patch-specific text prompts.
Our progressive upscaling framework uses a Stable Diffusion XL (SDXL)~\cite{podell2023sdxl} model adapted and fine-tuned for image restoration, requiring minimal architectural changes compared to prior works~\cite{yu2024scaling}. Additionally, we address the limitations of standard self-attention for high-resolution restoration by replacing it with PADRe~\cite{letourneau2024padre}, an efficient alternative with convolutional operations to improve computational efficiency and generate sharper textures at high resolutions.


\begin{figure}[!t]
    \centering
    \includegraphics[width=0.9\linewidth]{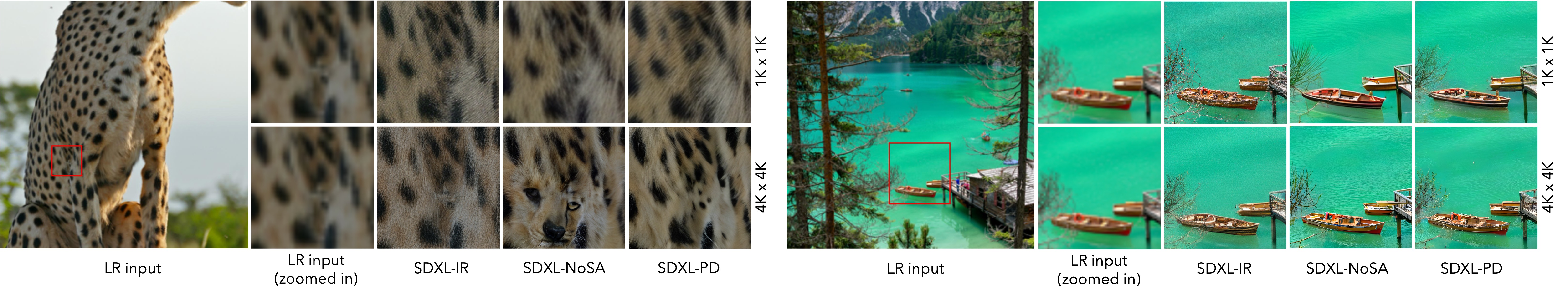}
    \captionof{figure}{Preliminary analysis of three SDXL-based restoration variants, SDXL-IR, SDXL-NoSA, and SDXL-PD, at $1024 \times 1024$ and $4096 \times 4096$. At higher resolution, SDXL-IR preserves structure but produces grainy artifacts, while SDXL-NoSA yields sharper textures but introduces hallucinated details. SDXL-PD provides a better balance between structural consistency and texture sharpness. \textbf{Zoom in for details.}}
    \vspace{-0.5cm}
    \label{fig:analysis_sdxl_vs_sdxlnosa}
\end{figure}

\section{Related Work}
\label{sec:related_work}

\textbf{Generative Image Restoration.} The goal of image restoration is to map a degraded image to a high-quality counterpart~\cite{fan2020neural, jinjin2020pipal, zhang2022accurate, zhang2017learning, zhang2019residual}. Early methods focused on single degradations~\cite{liang2021swinir}, while later approaches addressed complex, real-world scenarios~\cite{zhang2021designing, wang2021real}. Recent advances in restoration and super-resolution increasingly leverage pre-trained text-to-image models~\cite{lin2024diffbir, yu2024scaling, wu2024seesr, yang2024pixel, wang2024exploiting}, often using ControlNet~\cite{zhang2023adding} to condition on low-quality inputs. For example, PASD~\cite{yang2024pixel} and DiffBIR~\cite{lin2024diffbir} augment ControlNet with additional modules for input enhancement, while SPIRE~\cite{qi2024spire} combines ControlNet with synthetic prompts. SUPIR~\cite{yu2024scaling} integrates SDXL~\cite{podell2023sdxl} with a degradation-aware encoder and a streamlined ControlNet adapter. MMSR~\cite{mei2025power} departs from adapter-based designs by conditioning on depth, edges, and segmentation maps. In contrast, our approach achieves high-quality, high-resolution restoration by adapting and fine-tuning pre-trained text-to-image model without complex architectural modifications.


\textbf{High-resolution image restoration.} Most diffusion-based techniques~\cite{lin2024diffbir,yu2024scaling,wu2024seesr,yang2024pixel,mei2025power} are not optimized for high-resolution outputs. StableSR~\cite{wang2024exploiting} extends Stable Diffusion~\cite{rombach2022high} beyond $4\times$ scaling using overlapping latent patches denoised independently and merged via Gaussian-weighted aggregation~\cite{jimenez2023mixture}. While effective, this approach is computationally expensive and lacks text guidance, resulting in less detailed outputs. FaithDiff~\cite{chen2025faithdiff} is an SDXL‑based image restoration method that generates high‑resolution outputs through VAE tiling combined with a single global text prompt. Because it relies solely on global prompting without region‑specific guidance, the generated images tend to lack fine‑grained local details.
4KAgent~\cite{zuo20254kagent} introduces a multi-agent framework for $16\times$ restoration but is limited to inputs above $250\times250$ pixels. In contrast, our method enables up to $32\times$ scaling even for small inputs. Vivid‑VR \cite{bai2025vivid} focuses on video restoration and generates per‑tile captions using a vision‑language model to guide diffusion-based enhancement at high resolution. However, it follows a MultiDiffusion‑style denoise‑then‑blend pipeline \cite{bar2023multidiffusion}, where each tile is processed independently at every timestep. In contrast, our method injects per‑patch text embeddings directly into the cross‑attention layers while the rest of the network operates on the full latent, enabling globally coherent high‑resolution generation.

\textbf{Efficient Attention.} Efficient attention mechanisms aim to reduce the quadratic complexity of standard self-attention~\cite{vaswani2017attention} while maintaining performance. The PADRe framework~\cite{letourneau2024padre} unifies several efficient approaches, such as Hyena~\cite{poli2023hyena}, Mamba~\cite{gu2024mamba}, and SimA~\cite{koohpayegani2024sima}, under a polynomial attention formulation. This replaces softmax with hardware-friendly Hadamard products, delivering significant speed and memory improvements without compromising accuracy.

\begin{figure}[!t]
\centering
\begin{minipage}[t]{0.43\linewidth}
    \centering
    \includegraphics[width=\linewidth]{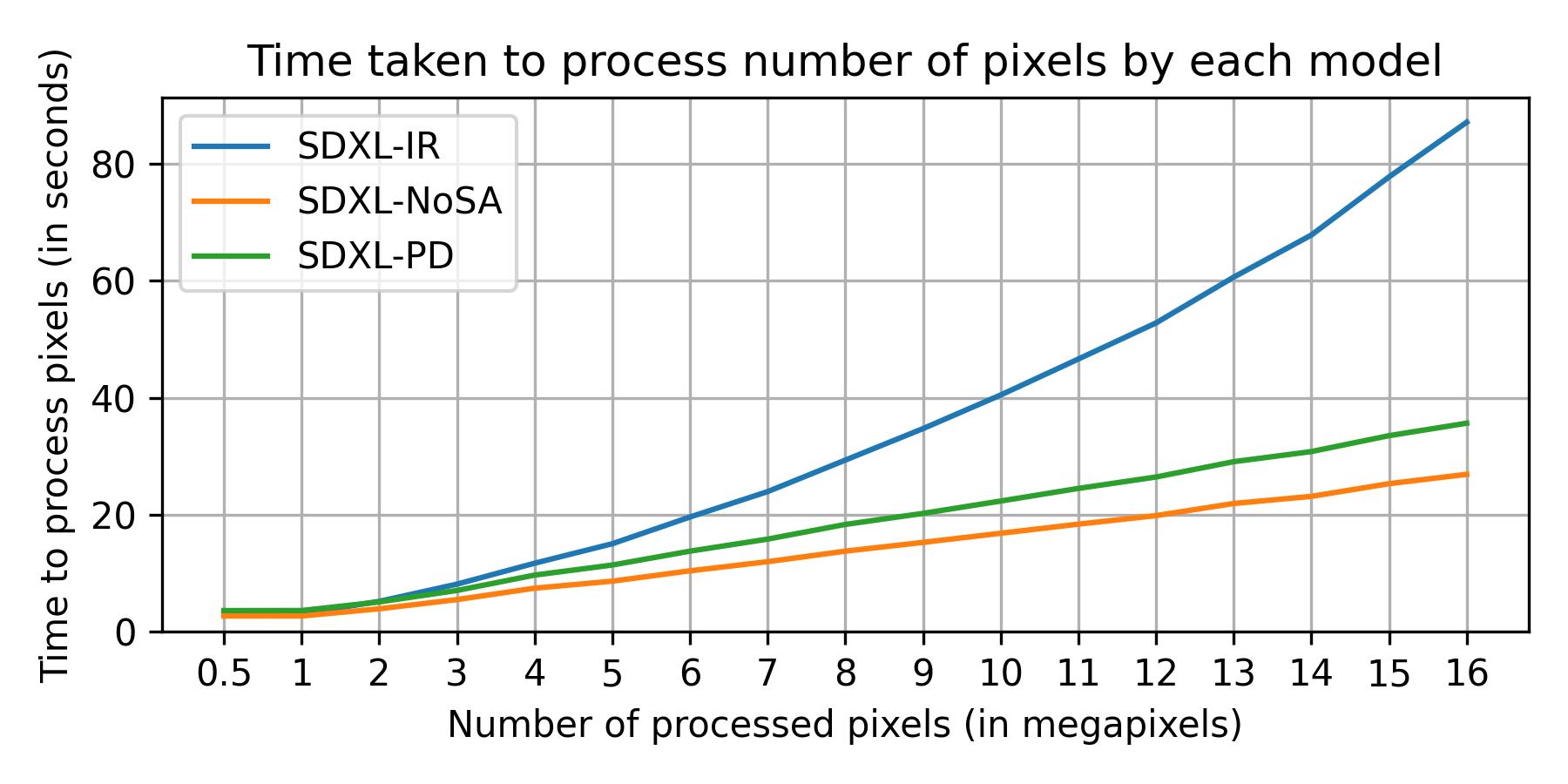}
    \caption{A100 runtime comparison of SDXL-IR, SDXL-NoSA, and SDXL-PD across increasing resolutions. All models perform similarly up to 3 MP, after which SDXL-IR shows a sharp increase due to quadratic cost of self-attention layers.}
    \label{fig:runtime}
\end{minipage}\hfill
\begin{minipage}[t]{0.45\linewidth}
    \centering
    \includegraphics[width=\linewidth]{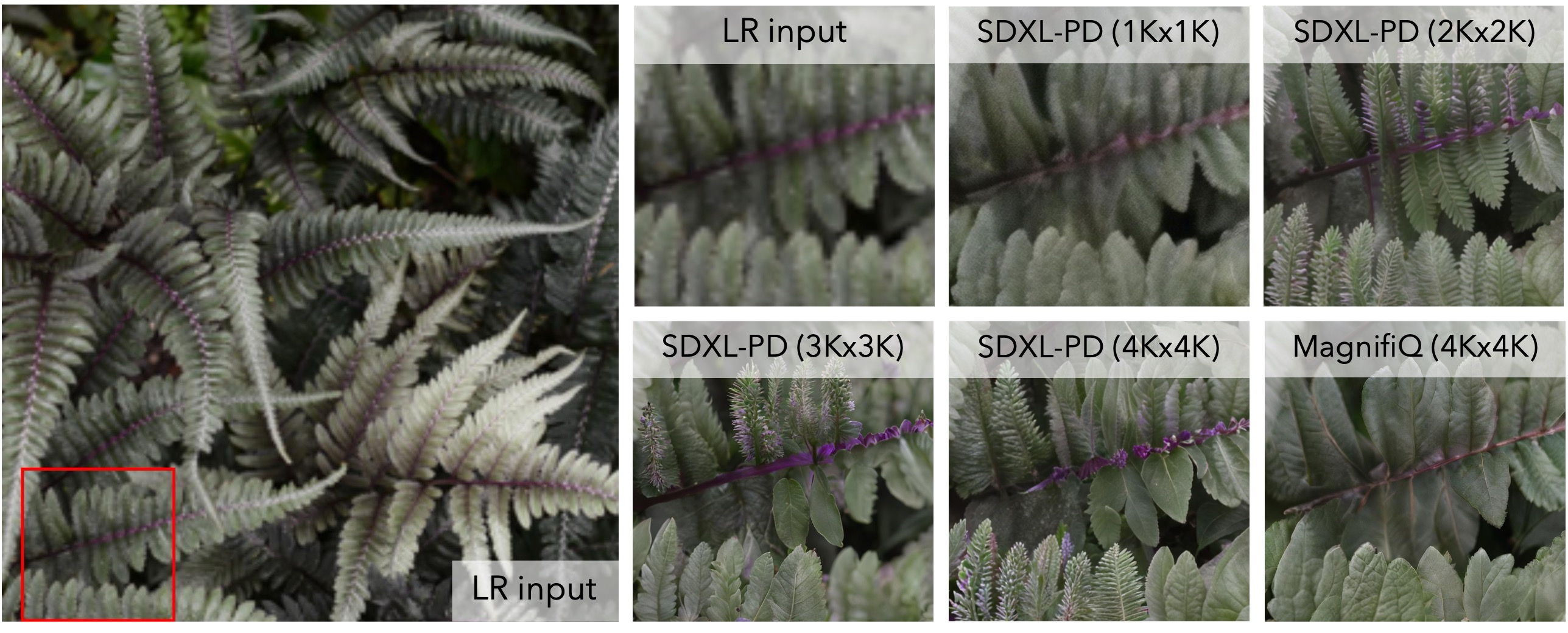}
    \caption{Direct high-resolution restoration with SDXL-PD often creates duplicated textures and texture repetition artifacts. \textbf{Zoom in to view fine details.}}
    \label{fig:duplications}
\end{minipage}
\vspace{-1.5em}
\end{figure}

\section{Observations on Image Restoration}
\label{sec:method}



Single-image restoration aims to reconstruct a high-resolution image from its low-resolution or degraded counterpart using an image restoration model. In recent years, text-conditioned generative approaches have become increasingly prominent, where a textual prompt serves as an auxiliary conditioning signal that provides semantic guidance to the model.  
Recent approaches~\cite{wu2024seesr,yang2024pixel,wang2024exploiting,lin2024diffbir} have adopted variants of Stable Diffusion~\cite{rombach2022high} as backbone architectures for image restoration. 
More recently, SUPIR~\cite{yu2024scaling} leveraged SDXL~\cite{podell2023sdxl} to generate restored images at resolutions up to 1024×1024, demonstrating impressive restoration quality at this scale. However, SUPIR remains limited to 1024×1024--the native pre-training resolution of SDXL--raising the question of whether restoration can be extended beyond this resolution while preserving global structure and visual fidelity.

\textbf{SDXL-based Image Restoration} To investigate high-resolution restoration, we adapt a pre-trained SDXL model by modifying its input convolution layer, similar to InstructPix2Pix~\cite{brooks2023instructpix2pix}, to take as input a low-quality conditioning image concatenated with the noisy latents, and fine-tune the network for restoration at $1024 \times 1024$, as illustrated in Fig.~\ref{fig:padre_block}a. At this native resolution, the model performs as expected, producing higher-quality reconstructions. However, when applied to resolutions beyond $1024 \times 1024$, this adapted model ``SDXL-IR'' preserves global structural consistency but exhibits noticeable degradations in local details (e.g., graininess), as illustrated in Fig.~\ref{fig:analysis_sdxl_vs_sdxlnosa}.

\textbf{Self-Attention vs. High-Resolution} We attribute these artifacts to the self-attention layers in  SDXL, which are inherently tuned to the training resolution, a limitation also noted in prior work~\cite{wang2024exploiting}. Specifically, self-attention exploits spatial correlation patterns that vary with resolution. As resolution increases, the number of tokens grows substantially, diluting the model’s ability to attend to salient features and ultimately leading to outputs with diminished visual fidelity. Furthermore, self-attention mechanisms face fundamental constraints in high-resolution image synthesis due to their quadratic computational complexity as shown in Fig.~\ref{fig:runtime}. 

To further examine this behavior, we fine‑tune SDXL‑IR with all self‑attention layers removed, denoted as ``SDXL‑NoSA''. As shown in Fig.~\ref{fig:analysis_sdxl_vs_sdxlnosa}, this variant produces noticeably sharper high‑frequency details with fewer grainy or blurry artifacts, and also reduces runtime substantially (Fig.~\ref{fig:runtime}). The absence of self‑attention, however, limits the model’s ability to capture long‑range spatial dependencies, which weakens global structural coherence and can lead to hallucinated regions. Despite this, fine‑tuning remains stable because the convolutional ResNet blocks and cross‑attention layers provide strong local texture priors, multi‑scale features, and conditioning signals, forming a sufficiently capable architecture for optimization. In our image‑restoration setup, much of the global structure is already present in the input, so self‑attention primarily serves to maintain global coherence rather than reconstruct it. This motivates exploring whether more efficient mechanisms could offer similar global consistency at lower computational cost.

\section{High-resolution Image Restoration}
\label{sec:magnifiq}



\subsection{Efficient SDXL-based Image Restoration}

To improve both computational efficiency and visual quality at high resolution, we replace self-attention with efficient yet expressive convolutional blocks constructed using nonlinear polynomials, inspired by PADRe~\cite{letourneau2024padre}. PADRe provides a unified framework that encompasses several instances of efficient attention designs~\cite{poli2023hyena,koohpayegani2024sima,gu2024mamba}.

Specifically, we replace standard self-attention with large-kernel convolutions (token mixing), pointwise convolutions (channel mixing), and Hadamard products. These operations construct polynomial functions over the input tensor, with nonlinear (degree $i > 1$) multiplicative interactions. This avoids the quadratic cost of self-attention while effectively capturing both local and larger-scale structures via large-kernel convolutions, scaling linearly in time and memory with resolution.

Following PADRe~\cite{letourneau2024padre}, given an input feature $F$, a linear transformation is applied using token mixing $A_i$ and channel mixing $B_i$ operations to get $Y_i$:
\begin{equation}
    Y_i = A_i F  B_i,
    \label{eq:linear_trasnformation}
\end{equation}
where channel mixing implements a pointwise convolution layer, and token mixing implements a convolution layer with kernel size of $k \times k$. Nonlinear terms $Z_i$ of degree $i$ is then computed via:

\begin{align}
  Z_{i} &= C_{i-1} Y_{i-1} D_{i-1} \odot Y_{i}, 
\end{align}

where $\odot$ denotes element-wise multiplication (Hadamard product), 
and $C_{i-1}$ and $D_{i-1}$ are additional token and channel mixing operations on $Y_{i-1}$. The final output is obtained by summing all $Z_i$ terms, forming a polynomial of degree $i$. A degree-3 PADRe block is shown in Fig.~\ref{fig:padre_block}b.

We refer to this variant as ``SDXL-PD''. As shown in Fig.~\ref{fig:analysis_sdxl_vs_sdxlnosa}, replacing self-attention with PADRe enhances visual quality at higher resolutions, producing images with sharper high-frequency details compared to ``SDXL-IR'' and fewer hallucinations compared to ``SDXL-NoSA'', effectively recovering the global coherence lost in the self-attention-free model while retaining its improved local fidelity. Furthermore, SDXL-PD achieves runtime comparable to SDXL-NoSA, as shown in Fig.~\ref{fig:runtime}, confirming that the efficiency gains of removing self-attention are preserved.


\begin{figure*}[!t]
    \centering
    \includegraphics[width=0.7\linewidth]{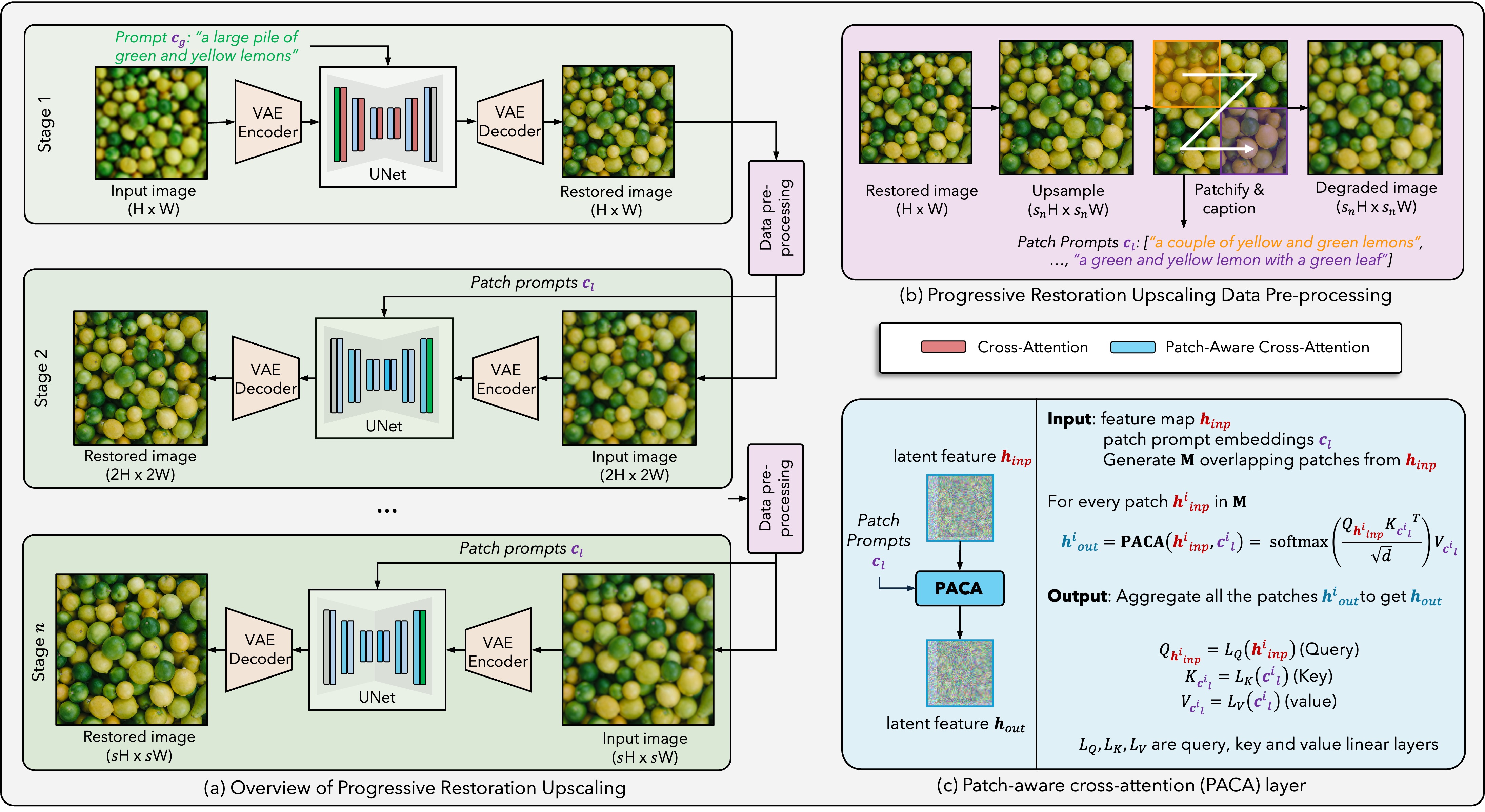}
    \captionof{figure}{Overview of MagnifiQ's progressive upscaling and patch-aware guidance. (a) High-resolution restoration is performed through $n$ iterative stages. (b) At each stage, we upscale the previous output, extract patch-level text prompts with LLaVA~\cite{liu2024improved}, and apply light degradation before restoration. (c) Patch-Aware Cross-Attention (PACA) routes each patch prompt to its corresponding latent patch during cross-attention.}
    \label{fig:high_resolution_architecture}
\end{figure*}

\begin{figure*}[!t]
    \centering
    \includegraphics[width=0.75\linewidth]{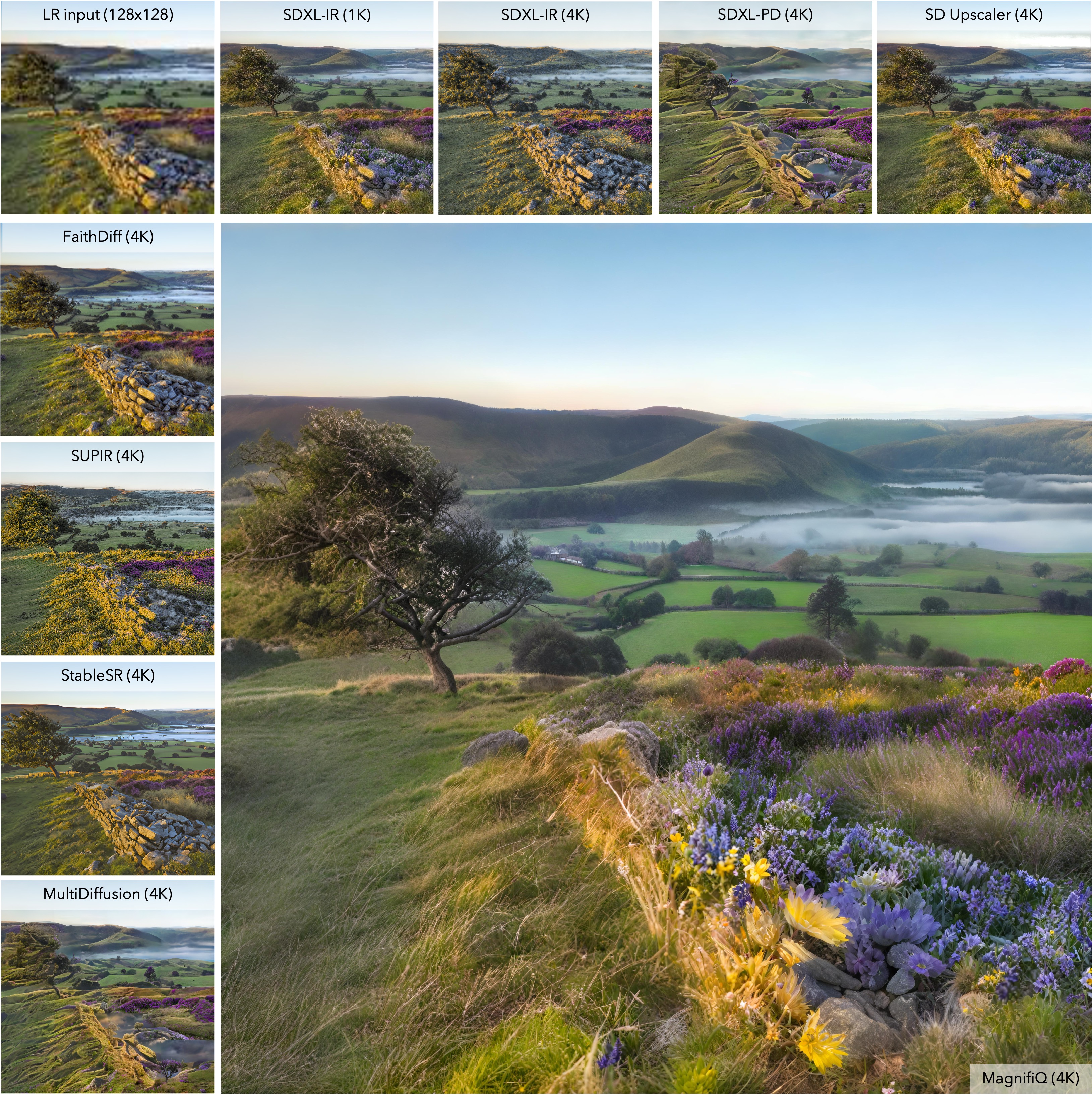}
    \caption{Overview of $4096 \times 4096$ ($32\times$) images generated from $128 \times 128$ low-quality inputs, comparing StableSR, SUPIR, SDXL-IR, SDXL-PD, SD Upscaler (with SDXL-PD), MultiDiffusion (with SDXL-PD), and MagnifiQ. Our MagnifiQ framework restores images to high resolution with realistic details and superior visual quality. \textbf{Zoom in to view fine details.}}
    \label{fig:qual_4k_v1}
    \vspace{-0.5cm}
\end{figure*}

\subsection{Restoration-aware Progressive Upscaling}


Although SDXL-PD achieves a more favorable balance between image quality and computational efficiency, restoring images at extreme scales ($16\times$ or $32\times$) still introduces artifacts such as local texture repetitions and object duplications, as illustrated in Fig.~\ref{fig:duplications}. Similar artifacts are well-documented in text-to-image generation~\cite{rombach2022high,podell2023sdxl}, where they arise from the model's limited ability to maintain global coherence across large spatial extents.

In our restoration setting, a key contributing factor is the direct upscaling of the degraded input to the target resolution. At extreme scale factors, this aggressive upscaling erodes structural cues in the input, causing the model to shift toward generative behavior, relying predominantly on the text prompt to synthesize local details rather than reconstructing them from the low-quality image. This effectively reduces the restoration problem to conditional generation, where repetition artifacts are a known failure mode.

To address texture repetitions and object duplications in high-resolution restoration, we propose a progressive upscaling paradigm that leverages a pre-trained image restoration model SDXL-PD ($f_{\theta}$). 
Given a degraded input image $\bm{x}_{lr} \in \mathbb{R}^{c \times H \times W}$ and a scaling factor $s$, our goal is to generate a restored image $\bm{x}_{hr} \in \mathbb{R}^{c \times sH \times sW}$ through $n$ progressive stages.


We first upscale the low-quality image to the model’s native resolution (e.g., $1024 \times 1024$) using an interpolation method $\phi$, and apply the restoration model 
with a global text prompt $\bm{c}_{g}$ for stage $n=1$. For each subsequent stage $n>1$, we: (i) upscale the restored image $\bm{x}^{n-1}_{hr}$ from the previous stage $n-1$ to the target resolution defined by scale $s_{n}$ for current stage $n$; (ii) generate local prompts $\bm{c}_{l}$ based on overlapping patches of the current upscaled image; (iii) apply a light degradation (e.g., resampling) before finally (iv) running the restoration model with local prompts. The overall iterative process can be expressed as:

\begin{align}
        \bm{x}^{1}_{hr} &= f_\theta\big(\phi(\bm{x}_{lr}), \bm{c}_{g}\big), & n = 1,\\
        \bm{x}^{n}_{hr} &= f_\theta\big(\phi({\bm{x}^{n-1}_{hr}},\  s_{n}), \bm{c}_{l})\big), & n > 1.
\end{align}


By refining the image over multiple stages, we avoid directly upscaling the original degraded input to the final resolution. Instead, each stage builds upon the previous output, as shown in Fig.~\ref{fig:high_resolution_architecture}a, preserving structural cues and reducing artifacts such as texture duplication, as shown in Fig.~\ref{fig:duplications}.

\subsection{Patch-aware Cross Attention (PACA)}
The text prompt plays an important role in conditioning the restoration model to generate fine-grained details at high resolutions. To leverage this signal more effectively, we introduce patch-level text prompts, with latent patches attending to their corresponding prompts through cross-attention during progressive upscaling.


During a restoration-aware progressive upscaling stage $n>1$, given an input image $\phi(\bm{x}_{hr}^{n-1}, s_n)$, we extract $\mathbf{M}$ overlapping patches, computed as $\mathbf{M} = (({H-\Delta_{h}})/{\delta_{h}} + 1) \times (({W-\Delta_{w}})/{\delta_{w}} + 1)$, where $H$ and $W$ denote the upscaled image's height and width, $\Delta_h$ and $\Delta_w$ represent the vertical and horizontal patch sizes, and $\delta_h$ and $\delta_w$ are the respective strides. For each patch $i \in \mathbf{M}$, we generate a patch-level text prompt $\bm{c}^{i}_{l}$ using LLaVa~\cite{liu2024improved}, as illustrated in Fig.~\ref{fig:high_resolution_architecture}b. The complete set of text prompts is denoted as $\bm{c}_{l} = \{ \bm{c}^{i}_{l} \mid i = 1, \dots, \mathbf{M} \}$.
To incorporate these patch-level text prompts $\bm{c}_{l}$, we propose to modify the cross-attention layers of the restoration model $f_\theta$ to make them patch-aware, such that each latent patch attends only to its corresponding patch-level prompt. We dub this Patch-Aware Cross Attention (PACA) and requires no fine-tuning. 
Given an intermediate latent representation $\bm{h}_{inp}$ for the cross-attention layer, we generate $\mathbf{M}$ overlapping latent patches, denoted as $\bm{h}_{inp} = \{ \bm{h}^{i}_{inp} \mid i = 1, \dots, \mathbf{M} \}$, by adjusting the patch size $\Delta$ and stride $\delta$ to match the latent's height and width. The PACA output is $\bm{\hat{h}}_{out} = \{ \bm{\hat{h}}^{i}_{out} \mid i = 1, \dots, \mathbf{M} \}$, obtained by applying cross-attention to each pair $(\bm{h}^{i}_{inp}, \bm{c}^i_l)$, which can be formulated as:

\begin{align}
    \bm{\hat{h}}^{i}_{out} = \text{PACA}(\bm{h}^{i}_{inp}, \bm{c}^{i}_{l}) = \text{softmax}\Biggl(\dfrac{Q_{\bm{h}^{i}_{inp}}K_{\bm{c}^{i}_l}^{T}}{\sqrt{d}}\Biggr)V_{\bm{c}^{i}_l},
\end{align}



where $Q_{\bm{h}^{i}_{inp}} = L_{Q}(\bm{h}^{i}_{inp})$, $K_{\bm{c}^{i}_{l}} = L_{K}(\bm{c}^{i}_{l})$, and $V_{\bm{c}^{i}_{l}} = L_{V}(\bm{c}^{i}_{l})$. Here, $Q$ is computed from the input latent patch $\bm{h}^{i}_{inp}$, while $K$ and $V$ are derived from the corresponding text embedding $\bm{c}^{i}_{l}$ for that patch. The term $d$ denotes the scaling factor used in the cross-attention mechanism.

The final representation $\bm{h}_{out}$ is produced by averaging the overlapping regions across all patches in $\bm{\hat{h}}_{out}$, ensuring it matches the dimensions of $\bm{h}_{inp}$, as shown in Fig.~\ref{fig:high_resolution_architecture}c. Importantly, only the PACA layer performs patch-level processing at each denoising step; all other layers are left unchanged.

\begin{table}[!t]
\centering
\scriptsize
\setlength{\tabcolsep}{2.6pt}
\renewcommand{\arraystretch}{0.95}
\begin{tabular}{@{}lcccc@{}}
\toprule
Method / Config. & MANIQA$\uparrow$ & CLIP$\uparrow$ & Aes.$\uparrow$ & Time$\downarrow$ \\
\midrule
\multicolumn{5}{@{}l}{\textit{Comparison with high-resolution restoration methods}} \\
StableSR       & .332 & .356 & 3.10 & 608 \\
SUPIR          & .308 & .461 & 4.14 & 466 \\
FaithDiff      & \textbf{.372} & .577 & 4.32 & 85 \\
SDXL-IR        & .326 & .559 & 4.31 & 109 \\
SDXL-PD        & .330 & .573 & 4.29 & 45 \\
MultiDiff.     & .348 & .586 & 4.30 & 190 \\
SD Upscaler    & .339 & .571 & 4.08 & 68 \\
\rowcolor[gray]{0.9}MagnifiQ & \underline{.363} & \textbf{.622} & \textbf{4.84} & 393 \\
\midrule
\multicolumn{5}{@{}l}{\textit{MagnifiQ captioner/stride ablation}} \\
DAPE, $\delta{=}0$      & .338 & .591 & 4.73 & 103 \\
DAPE, $\delta{=}512$    & .340 & .590 & 4.73 & 158 \\
LLaVA, $\delta{=}0$     & .360 & \underline{.619} & \underline{4.83} & 197 \\
LLaVA, $\delta{=}512$   & \underline{.363} & \textbf{.622} & \textbf{4.84} & 393 \\
\bottomrule
\end{tabular}
\caption{No-reference Aesthetic-4K evaluation on 50 images. We report MANIQA, CLIP-IQA, LAION-Aesthetic, and runtime in seconds on one NVIDIA H100 GPU. $\delta$ denotes PACA stride.}
\label{tab:aesthetic4k_compact}
\end{table}

\section{Experiments and Results}



\begin{figure}[!t]
\centering
\begin{minipage}[t]{0.35\linewidth}
    \centering
    \includegraphics[width=\linewidth]{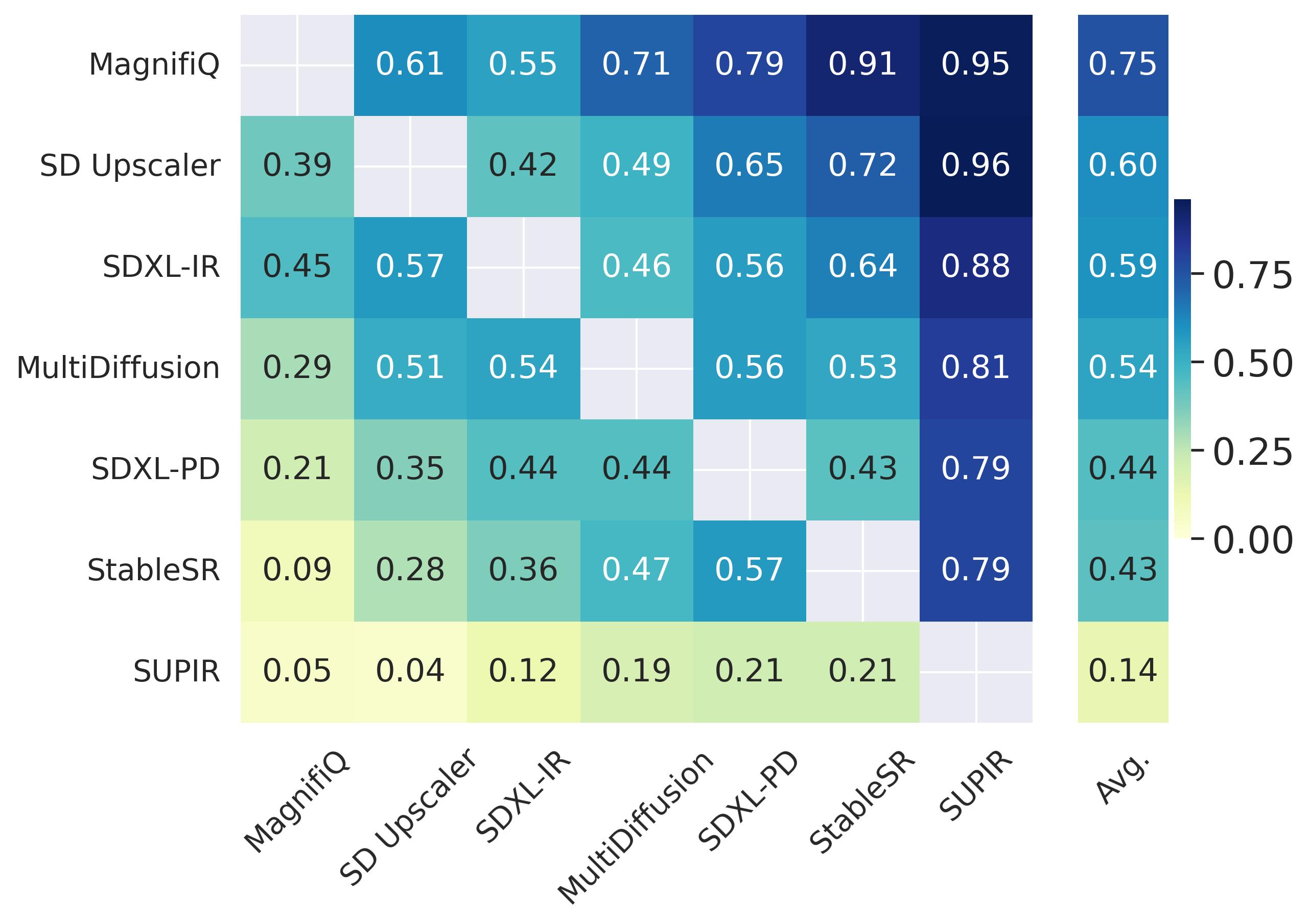}
    \caption{User study results for restoring images to $4096 \times 4096$ resolution. MagnifiQ achieves the highest preference, with $75\%$ of users favoring it over other methods on average.}
    \label{fig:userstudy}
    \vspace{-0.5cm}
\end{minipage}
\hfill
\begin{minipage}[t]{0.6\linewidth}
    \centering
    \includegraphics[width=\linewidth]{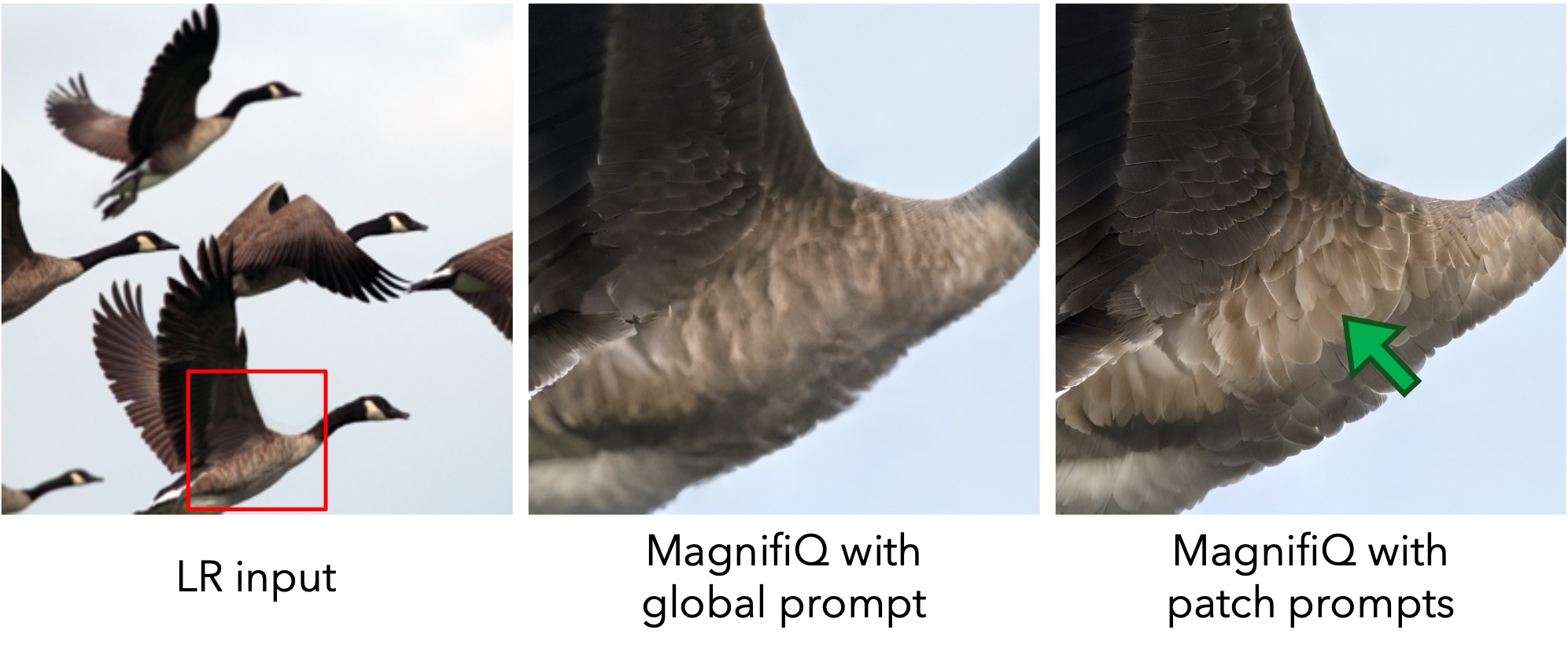}
    \caption{Comparison of local patch prompts vs. a single global prompt at $4096 \times 4096$. Patch prompts yield sharper fine-grained details (see {\color{green}green} arrows). Refer to Sec.~\ref{sec:abl_prompts} for prompt details. \textbf{Zoom in to view fine details.}}
    \label{fig:abl_prompts}
\end{minipage}
\end{figure}

\begin{figure}
    \centering
    \includegraphics[width=0.6\linewidth]{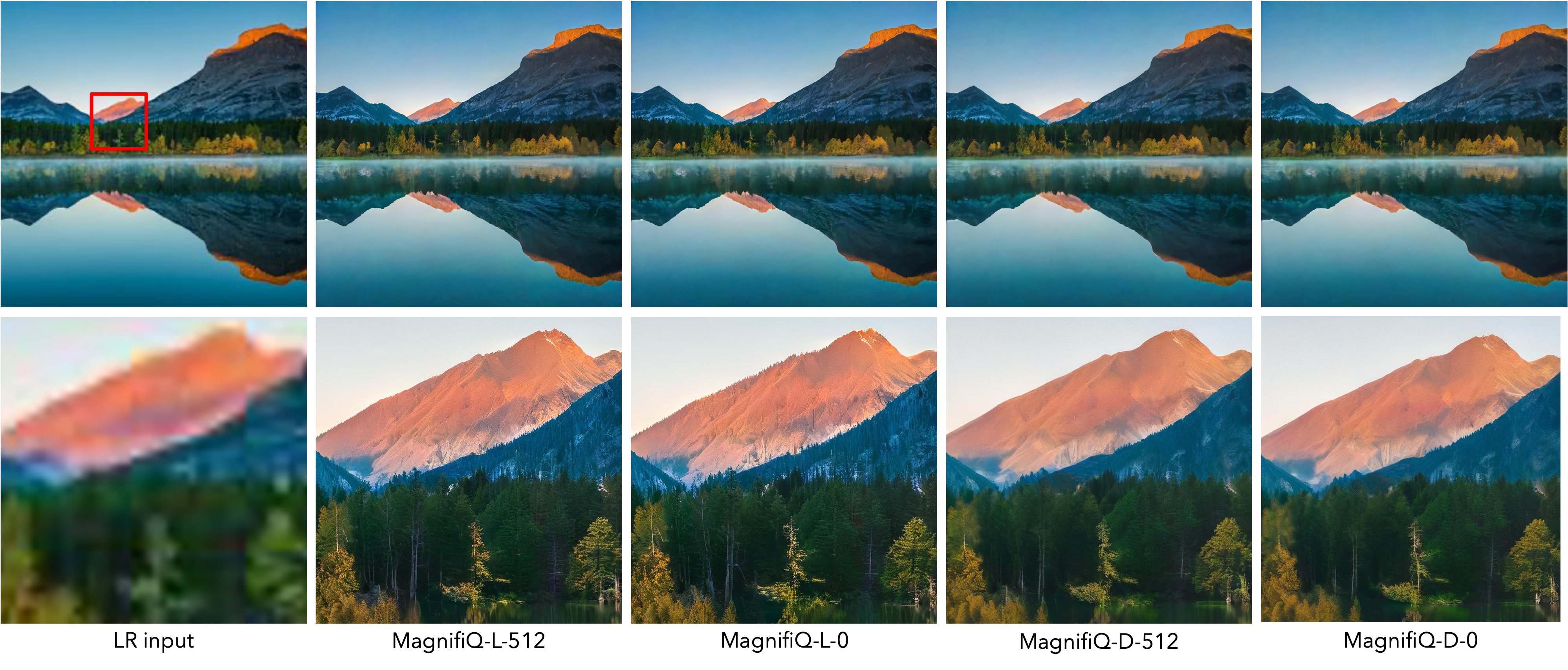}
    \caption{Comparing generated images for different captioning models and patch‑stride settings. We compare LLaVA‑1.5 (L) \cite{liu2024improved} and DAPE (D) \cite{wu2024seesr} for patch‑level caption generation, and evaluate two stride configurations corresponding to overlapping patches (512 px) and no overlap (0 px). MagnifiQ with LLaVA‑based patch captions tends to produce images with richer local details, reflecting the more descriptive semantic prompts provided by LLaVA compared with DAPE at the patch level. \textbf{Zoom in to view fine details.}}
    \label{fig:abl_caption_stride}
    \vspace{-0.5cm}
\end{figure}

\subsection{MagnifiQ}
\textbf{Evaluation details.} In our experiments, we set PADRe degree $i = 3$ using $1 \times 1$ convolutions for channel mixing, and a $5 \times 5$ kernel for token mixing. For our MagnifiQ pipeline, we use $n = 4$ progressive upscaling stages and generate patches with a size of $\Delta_h = 1024$, $\Delta_w = 1024$, using strides of $\delta_h = 512$ and $\delta_w = 512$. We use $50$ denoising steps and a text guidance scale of $4$ across all experiments. For more details, please refer to the supplementary material. To evaluate MagnifiQ, we perform $16\times$ upscaling and compare the generated $4096 \times 4096$ outputs from $256 \times 256$ degraded inputs against several restoration baselines: StableSR~\cite{wang2024exploiting}, SUPIR~\cite{yu2024scaling}, MultiDiffusion~\cite{bar2023multidiffusion}, SD Upscaler~\cite{rombach2022high}, SDXL-IR, and SDXL-PD. For StableSR, we use its official high-resolution generation pipeline. SUPIR, SDXL-IR, and SDXL-PD perform direct $16\times$ generation by resizing the input image to $4096 \times 4096$.  MultiDiffusion is evaluated by integrating the SDXL-PD model within its framework. For SD Upscaler, we first generate a $4\times$ upscaled output using SDXL-PD and then apply an additional $4\times$ upscaling with SD Upscaler to obtain the final $4096 \times 4096$ image.

\textbf{4K Quantitative evaluation.}
As shown in Tab.~\ref{tab:aesthetic4k_compact}, we evaluate all methods on 50 images sampled from the Aesthetic-4K dataset using no-reference metrics alongside runtime. MagnifiQ outperforms on CLIP-IQA, and LAION-Aesthetic, demonstrating consistently superior perceptual quality. While its runtime (393s) is higher than single-pass methods such as SDXL-PD (45s) and FaithDiff (85s), MagnifiQ remains substantially faster than StableSR (608s) and SUPIR (466s), both of which it outperforms on all four metrics. This favorable quality-to-runtime tradeoff validates that the additional cost of progressive multi-resolution upscaling with per-patch semantic conditioning is well justified by the consistent gains in perceptual quality.

\textbf{4K User study.}~\label{sec:user_study} To evaluate perceptual quality, we conducted a user study on the $4096 \times 4096$ outputs produced by all models using images from the DIV2K~\cite{agustsson2017ntire} dataset. For more details on the user-study setup, please refer to the supplementary material. Fig.~\ref{fig:userstudy} shows the results of our user study, where participants were presented with pairwise comparisons of 4K outputs and asked to select the image they preferred. Our proposed MagnifiQ achieves the best performance with an average user preference of $75\%$, demonstrating its robustness and reliability relative to all baselines. Among the other methods, SD Upscaler and MultiDiffusion (with SDXL-PD) achieve average preferences of $60\%$ and $54\%$, respectively. This suggests that the $4\times$-upscaled $1024 \times 1024$ output from SDXL-PD both conditions SD Upscaler effectively and integrates well within the MultiDiffusion framework. Interestingly, direct inference at $4096 \times 4096$ using SDXL-IR scores $59\%$, outperforming SDXL-PD ($44\%$). As illustrated in Fig.~\ref{fig:duplications}, this is partly because the single-pass SDXL-PD inference exhibits a stronger tendency toward duplicated textures at high resolution. 
StableSR and SUPIR receive significantly lower preference scores, with SUPIR averaging only $14\%$, reflecting a substantial performance gap compared to MagnifiQ. Qualitative comparisons in Fig.~\ref{fig:qual_4k_v1} further show that MagnifiQ consistently produces sharper local textures while maintaining strong global structural consistency.

\begin{figure}[!t]
\centering
\begin{minipage}[t]{0.4\linewidth}
    \centering
    \includegraphics[width=\linewidth]{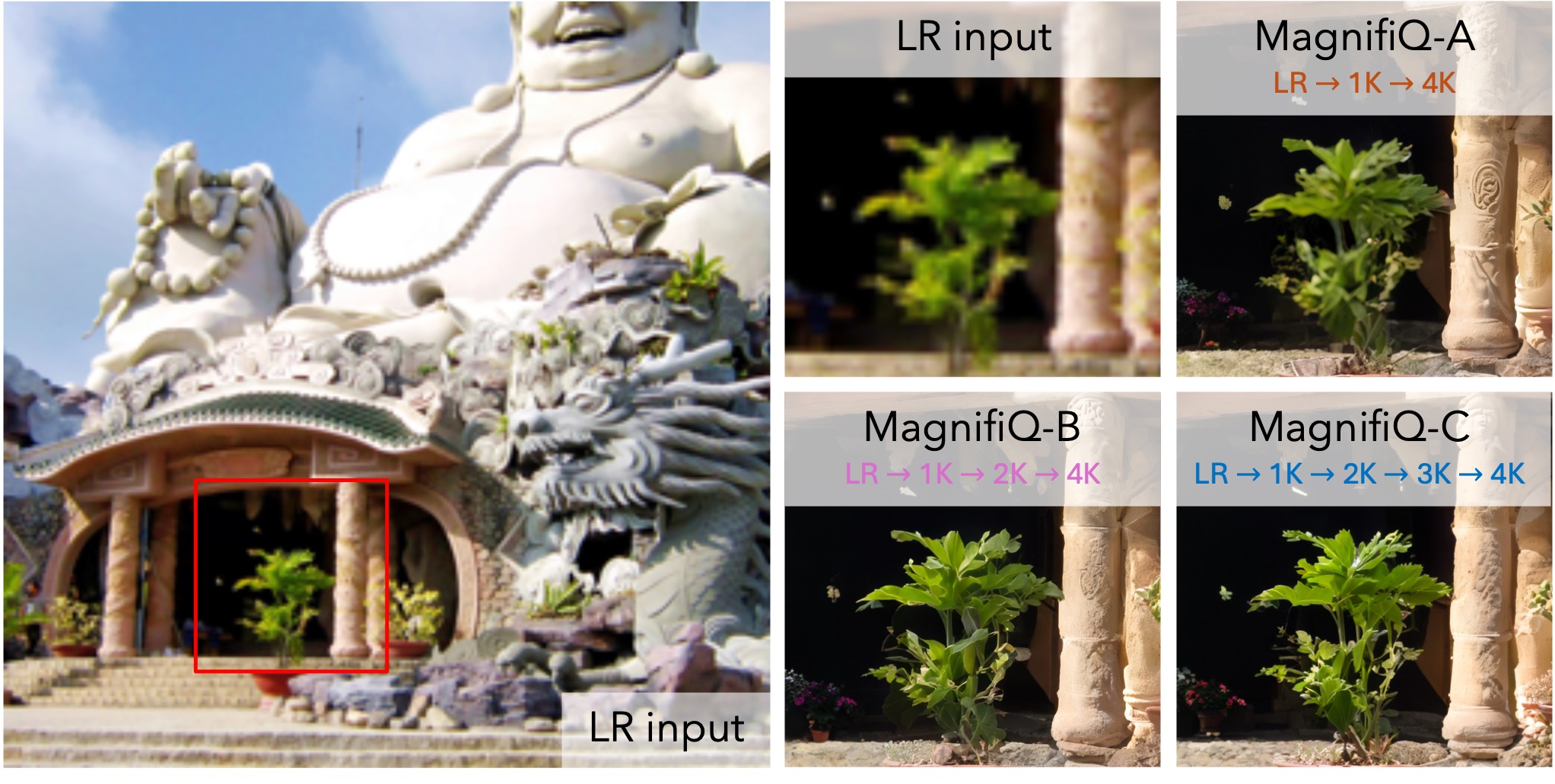}
    \caption{Comparison of $4096 \times 4096$ images generated with three MagnifiQ configurations (A: $n = 2$, B: $n = 3$, and C: $n = 4$). More stages yield better restorations, with C being the most detailed.
    See Sec.~\ref{sec:abl_configuration} for details. \textbf{Zoom in to view fine details.}}
    \label{fig:abl_stages}
\end{minipage}
\hfill
\begin{minipage}[t]{0.50\linewidth}
    \centering
    \includegraphics[width=\linewidth]{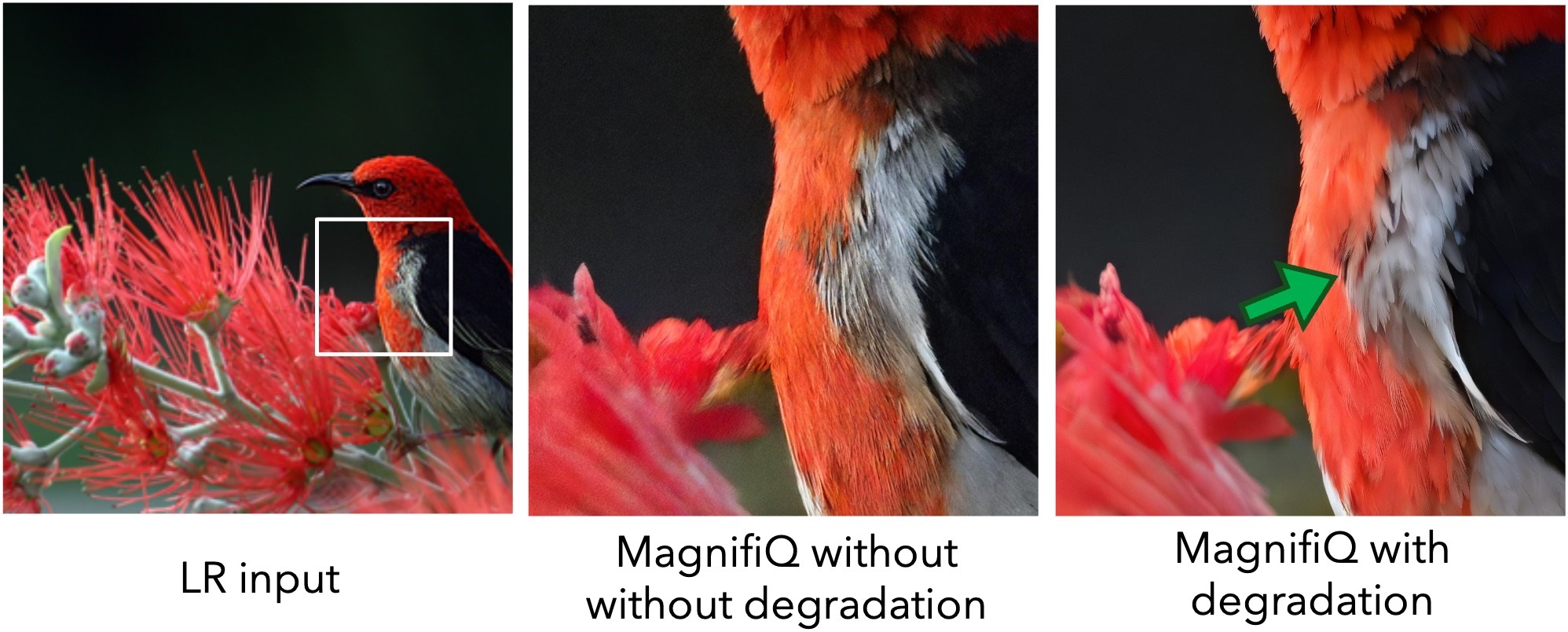}
    \caption{Adding a downsampling–upsampling degradation before each stage ($n > 1$) improves fine-detail restoration. For more details, please refer to Sec.~\ref{sec:abl_configuration}. \textbf{Zoom in to view fine details.}}
    \label{fig:abl_degradation}
    \vspace{-0.5cm}
\end{minipage}
\end{figure}

\textbf{Effect of Patch-Aware Prompts.}  
\label{sec:abl_prompts}
Fig.~\ref{fig:abl_prompts} compares using a single \textit{global} text prompt versus \textit{local} patch-aware prompts during progressive upscaling stages. Global prompts are generated from the full image, and captioning models such as LLaVA~\cite{liu2024improved} often produce high-level descriptions that omit fine-grained details and provide insufficient context for synthesizing rich local details. In contrast, generating prompts for individual image patches captures localized information, offering better guidance for restoring intricate structures. For example, in Fig.~\ref{fig:abl_prompts}, the global prompt states: \textit{``...the image features a large flock of geese flying together in the sky...''}, whereas the local prompt for the corresponding patch containing the zoomed-in bird reads: \textit{``...the bird's wings are the main focus of the image, highlighting its size and grace}...''. 
Additional results are provided in the supplementary material.

\begin{table}[!t]
  \centering
  \scriptsize
  \setlength{\tabcolsep}{2.2pt}
  \renewcommand{\arraystretch}{0.9}
  \begin{tabular}{@{}lccccccc@{}}
  \toprule
  Method & PSNR$\uparrow$ & SSIM$\uparrow$ & LPIPS$\downarrow$ & MQ$\uparrow$ & MSQ$\uparrow$ & PI$\downarrow$ & NIQE$\downarrow$ \\
  \midrule
  \multicolumn{8}{@{}l}{\textit{DIV2K with R-ESRGAN degradations}} \\
  R-ESRGAN & \textbf{19.95} & \textbf{.518} & .480 & .277 & 51.77 & 5.40 & 5.49 \\
  StableSR & 19.12 & .473 & .488 & .298 & 55.49 & 4.96 & 5.38 \\
  SeeSR & 19.54 & .499 & \underline{.448} & .342 & 61.30 & 4.59 & 4.85 \\
  PASD & 19.30 & .492 & .524 & .289 & 51.88 & 5.62 & 5.64 \\
  SUPIR & \underline{19.55} & \underline{.501} & .485 & .317 & 57.15 & 4.87 & 4.87 \\
  \rowcolor[gray]{0.9} SDXL-IR & 19.09 & .424 & .451 & \underline{.371} & 64.44 & \textbf{3.41} & \underline{3.97} \\
  \rowcolor[gray]{0.9} SDXL-NoSA & 18.87 & .445 & \textbf{.446} & .361 & \textbf{69.80} & \underline{3.53} & \textbf{3.67} \\
  \rowcolor[gray]{0.9} SDXL-PD & 19.26 & .455 & .456 & \textbf{.373} & \underline{68.44} & 3.70 & 4.03 \\
  \midrule
  \multicolumn{8}{@{}l}{\textit{DIV2K}} \\
  R-ESRGAN & 21.72 & \underline{.579} & .415 & .277 & 53.73 & 5.55 & 5.50 \\
  StableSR & 20.23 & .525 & .392 & .324 & 59.99 & 4.82 & 5.10 \\
  SeeSR & 21.33 & .552 & .379 & .339 & 61.56 & 4.61 & 4.72 \\
  PASD & 20.84 & .536 & .446 & .289 & 54.19 & 5.23 & 5.10 \\
  SUPIR & \textbf{23.00} & \textbf{.641} & \textbf{.293} & \textbf{.359} & \underline{66.03} & 3.98 & \underline{3.91} \\
  \rowcolor[gray]{0.9} SDXL-IR & 21.65 & .520 & .344 & \underline{.353} & 61.92 & \underline{3.85} & 4.67 \\
  \rowcolor[gray]{0.9} SDXL-NoSA & \underline{21.74} & .566 & \underline{.324} & .338 & \textbf{67.33} & \textbf{3.77} & \textbf{3.82} \\
  \rowcolor[gray]{0.9} SDXL-PD & 21.65 & .568 & .338 & .324 & 63.33 & 4.02 & 4.13 \\
  \midrule
  \multicolumn{8}{@{}l}{\textit{RealSR}} \\
  R-ESRGAN & 22.49 & \underline{.664} & .336 & .282 & 51.90 & 6.02 & 6.08 \\
  StableSR & 21.33 & .617 & .322 & .329 & 56.64 & 5.39 & 5.75 \\
  SeeSR & 22.48 & .647 & \textbf{.309} & \textbf{.368} & \underline{60.53} & 5.04 & 5.29 \\
  PASD & 21.94 & .629 & .352 & .302 & 51.52 & 5.71 & 5.74 \\
  SUPIR & \textbf{23.09} & \textbf{.679} & .339 & .331 & 58.38 & 4.95 & 5.12 \\
  \rowcolor[gray]{0.9} SDXL-IR & \underline{22.50} & .548 & .485 & .330 & 56.18 & \underline{4.15} & 5.12 \\
  \rowcolor[gray]{0.9} SDXL-NoSA & 21.82 & .596 & .366 & \underline{.340} & \textbf{64.35} & \textbf{4.12} & \textbf{4.35} \\
  \rowcolor[gray]{0.9} SDXL-PD & 22.48 & .625 & .370 & .316 & 59.61 & 4.69 & \underline{4.90} \\
  \bottomrule
  \end{tabular}
  \caption{Quantitative comparison of SDXL-IR, SDXL-NoSA, and SDXL-PD with prior restoration methods on $4\times$ downscaled DIV2K with R-ESRGAN
  degradations, clean DIV2K, and RealSR. MQ and MSQ denote MANIQA and MUSIQ, respectively. Baselines include R-ESRGAN~\cite{wang2021real},
  StableSR~\cite{wang2024exploiting}, SeeSR~\cite{wu2024seesr}, PASD~\cite{yang2024pixel}, and SUPIR~\cite{yu2024scaling}.}
  \label{tab:qual_1k}
  \end{table}

\textbf{Effect of the captioning model and stride.}
In Tab.~\ref{tab:aesthetic4k_compact}, we ablate runtime and quality for two captioning models: LLaVA-1.5 (L)~\cite{liu2024improved} and DAPE (D)~\cite{wu2024seesr} and two patch-overlap sizes (``stride''): 0\,px (non-overlapping) and 512\,px (large overlap).
Switching from LLaVA to DAPE yields substantial runtime reductions of \textbf{60\%} at fixed stride (e.g., 393$\rightarrow$158\,s at stride~512), while decreasing the overlap from 512\,px to 0\,px provides additional savings of \textbf{50\%} (e.g., 393$\rightarrow$197\,s for L).
Despite these runtime differences, the stride setting has only a minor effect on perceptual quality. 
This is expected because patching is applied \emph{only} inside the cross-attention pathway, whereas the remainder of the diffusion network processes the full latent globally; reducing overlap therefore changes the redundancy of semantic patch tokens without altering the model’s global context. 
The denoising process naturally smooths boundary inconsistencies, leading to negligible differences between stride values.
In contrast, the choice of captioning model has a more pronounced impact on quality: LLaVA produces richer and more descriptive visual-text embeddings, while DAPE prioritizes efficiency and generates more compact but less expressive representations. 
This establishes a clear trade-off: LLaVA offers higher perceptual and aesthetic fidelity at a substantially higher computational cost, whereas DAPE provides the fastest configuration with only modest quality degradation. In Fig.~\ref{fig:abl_caption_stride}, we illustrate the impact of the captioning model and patch‑stride setting on the generated high‑resolution image.

\textbf{Effect of MagnifiQ configurations.} 
\label{sec:abl_configuration}We analyze the impact of varying MagnifiQ configurations for generating $4096 \times 4096$ images. Specifically, we compare three configurations: MagnifiQ-{\orange{\textbf{A}}}, MagnifiQ-{\magenta{\textbf{B}}} and MagnifiQ-{\blue{\textbf{C}}}. In {\orange{\textbf{A}}}, stages $n = 2$ and the generation sequence is ${\bm{x}_{lr}} \rightarrow 1K \times 1K \rightarrow 4K \times 4K$, in {\magenta{\textbf{B}}}, stages $n = 3$ and the sequence is ${\bm{x}_{lr}} \rightarrow 1K \times 1K \rightarrow 2K \times 2K \rightarrow 4K \times 4K$, and in {\blue{\textbf{C}}}, stages $n = 4$ with a sequence of ${\bm{x}_{lr}} \rightarrow 1K \times 1K \rightarrow 2K \times 2K \rightarrow 3K \times 3K \rightarrow 4K \times 4K$. Fig.~\ref{fig:abl_stages} compares the results for all three configurations. The best looking results are produced by {\blue{\textbf{C}}}, as using more stages allows the model to progressively restore finer details. In contrast, configurations {\orange{\textbf{A}}} and {\magenta{\textbf{B}}} yield globally consistent high-resolution images but lack intricate local details, as evident in the zoomed-in regions. For instance, in Fig.~\ref{fig:abl_stages}, the plant exhibits richer texture and leaf details for {\blue{\textbf{C}}} with higher-stage settings. These observations highlight the strong influence of progressive stages on generating high-quality, detailed restorations. 
Additional results are provided in the supplementary material.




\begin{figure*}[!t]
    \centering
    \includegraphics[width=0.6\linewidth]{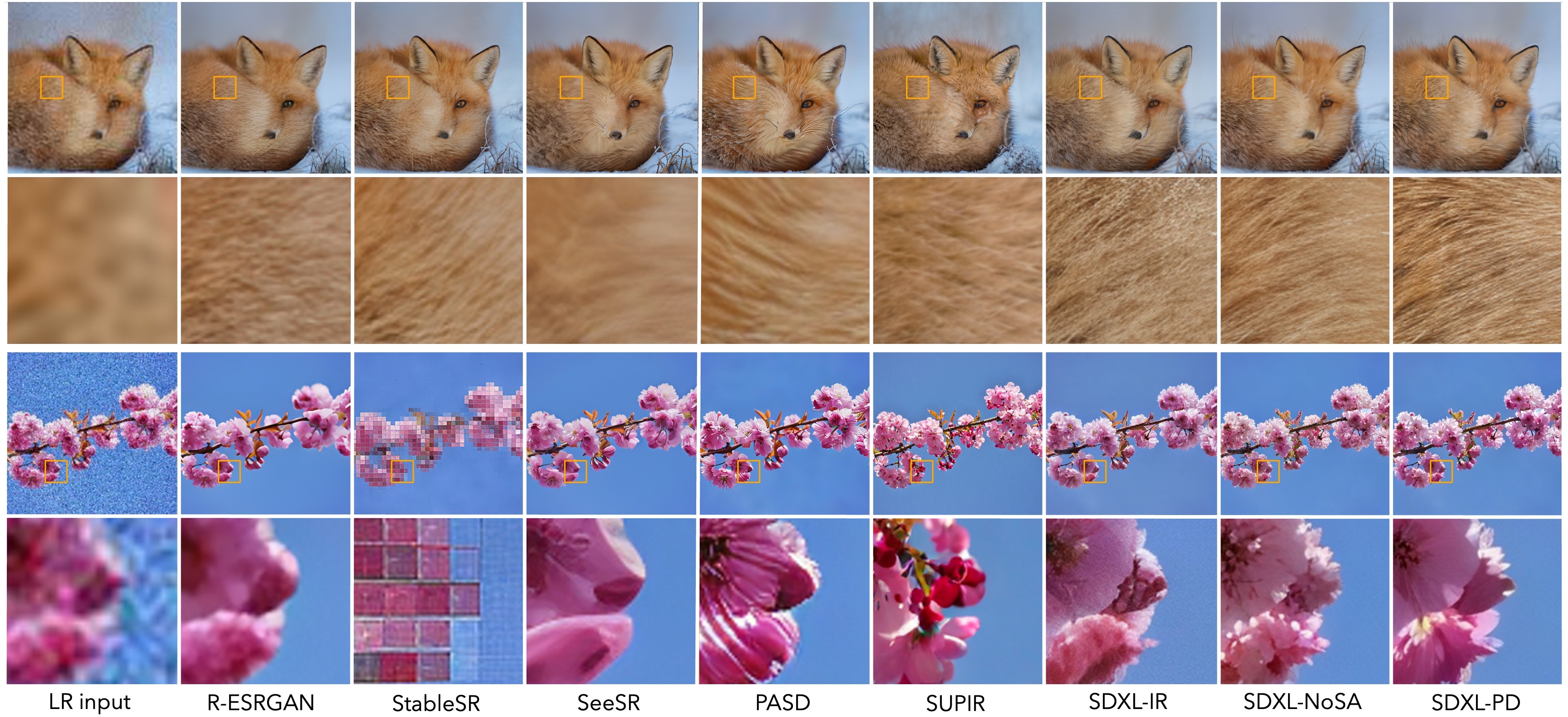}
  \caption{Qualitative results on the DIV2K~\cite{agustsson2017ntire} dataset show that SDXL-IR, SDXL-NoSA, and SDXL-PD effectively handle complex degradations, performing on par with recent state-of-the-art methods that rely on more complex architectures.}
    \label{fig:qual_1k}
    \vspace{-0.5cm}
\end{figure*}

\textbf{Effect of Progressive Upscaling Degradation.} We examine the impact of applying image degradation at stages $n \in \{2, 3, 4\}$ during progressive upscaling. The degradation involves resampling via a downscale–upscale step to 
$\frac{512}{H} \cdot H \times \frac{512}{W} \cdot W$. As shown in Fig.~\ref{fig:abl_degradation}, this improves visual quality. This is because our model is trained to restore degraded inputs; without this additional image degradation, only minor artifacts arise from upscaling the previous stage’s output. For example, in Fig.~\ref{fig:abl_degradation}, the red bird’s feathers appear more natural when degradation is applied.


\subsection{Image Restoration Evaluation} 
\label{sec:image_restoration_evaluation_1k}

In this section, we evaluate the fine-tuned SDXL-based image restoration models (SDXL-IR, SDXL-NoSA, and SDXL-PD) without the progressive upscaling component of MagnifiQ and compare them against recent approaches, including R-ESRGAN~\cite{wang2021real}, StableSR~\cite{wang2024exploiting}, SeeSR~\cite{wu2024seesr}, PASD~\cite{yang2024pixel}, and SUPIR~\cite{yu2024scaling}.

\textbf{Evaluation Details.} Following prior works~\cite{wang2024exploiting,yu2024scaling}, we report both reference-based metrics, including PSNR, SSIM, and LPIPS~\cite{zhang2018unreasonable} and no-reference metrics, including MANIQA (MQ)~\cite{yang2022maniqa}, MUSIQ (MSQ)~\cite{ke2021musiq}, Perceptual Index (PI)~\cite{blau20182018}, and NIQE~\cite{niqe}. Experiments are conducted on synthetic and real-world degraded datasets, including DIV2K~\cite{agustsson2017ntire} and RealSR~\cite{cai2019toward}. For DIV2K, we consider two setups: (i) original $(4\times)$ downscaled images and (ii) multiple synthetic degradations following R-ESRGAN~\cite{wang2021real}.

\textbf{Results.} Tab.~\ref{tab:qual_1k} shows quantitative comparisons. Unlike baselines that rely on complex architectural changes, such as adapter networks for low-quality inputs~\cite{yu2024scaling,wang2024exploiting}, our SDXL variants introduce only a simple modification to the input convolution layer yet achieve performance comparable to state-of-the-art SDXL-based methods like SUPIR~\cite{yu2024scaling}. SDXL-NoSA, which removes self-attention layers, performs strongly on no-reference metrics, indicating better perceptual quality, but prone to hallucinations. SDXL-PD strikes a balance between these variants, as shown in Fig.~\ref{fig:analysis_sdxl_vs_sdxlnosa}, and offers efficiency gains over SDXL-IR (Fig.~\ref{fig:runtime}). Qualitative results in Fig.~\ref{fig:qual_1k} further demonstrate SDXL-PD’s ability to generate sharp local details (e.g., fur), maintain perceptual quality (e.g., flowers), and preserve structural integrity (e.g., building architecture), highlighting its effectiveness for high-quality image restoration.

\begin{figure}[!t]
    \centering
    \includegraphics[width=0.45\linewidth]{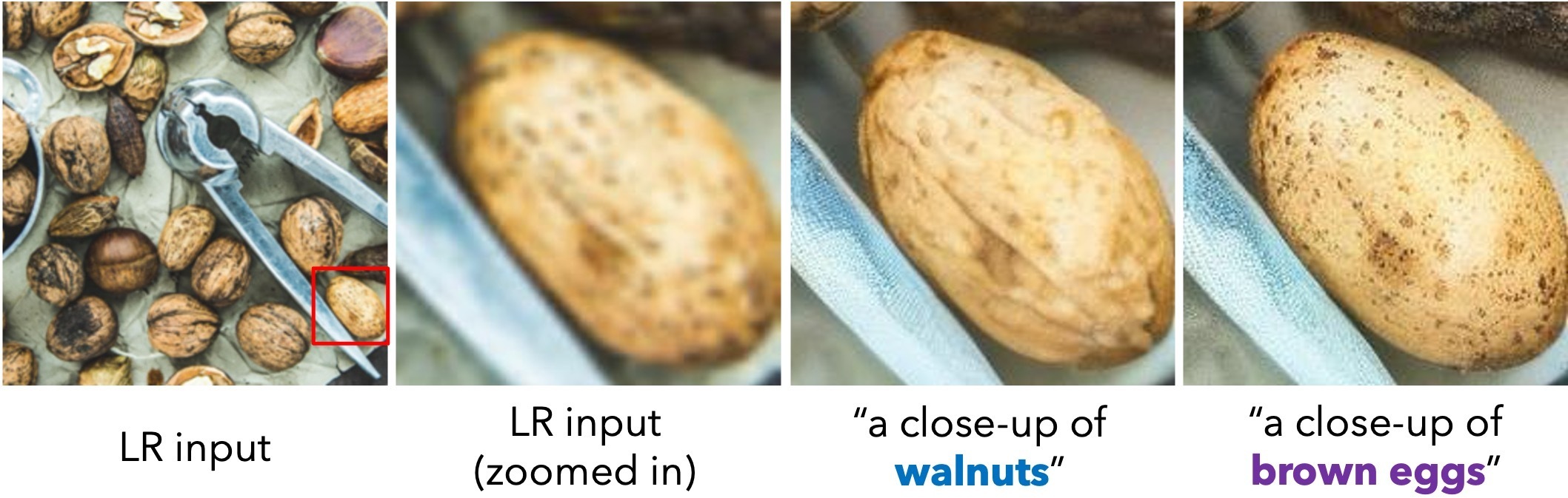}
    \caption{Controllable restoration with SDXL-PD: altering the prompt (e.g., ``walnut''$\rightarrow$``brown eggs'') changes the restored textures accordingly.}
    \label{fig:restoration_editing}
    \vspace{-0.5cm}
\end{figure}

\section{Conclusion and Discussion}
We propose MagnifiQ, a framework to restore low-quality images to resolutions up to $4096 \times 4096$ with scale factors of up to $32\times$ by using our fine-tuned efficient image restoration model. The framework introduces two key components:
(i) a training-free progressive upscaling paradigm that begins with global restoration using a text prompt and then iteratively applies patch-level prompts during upscaling to maintain structural coherence; and
(ii) a lightweight modification to the underlying pre-trained restoration model, incorporating patch-aware cross-attention to align patch-level prompts with corresponding latent patches, thereby enhancing fine details at high resolutions.
Furthermore, we explore a simple yet effective image restoration model without complex architectural changes by adapting and fine-tuning the SDXL architecture~\cite{podell2023sdxl} to condition the low-quality input image and replacing standard self-attention with efficient alternatives such as PADRe~\cite{letourneau2024padre}. These changes reduce computational overhead while improving generation quality at large scales compared to prior approaches. Although our implementation is based on SDXL, MagnifiQ is designed to be architecture-agnostic, enabling future exploration with transformer-based designs~\cite{esser2024scaling}. Our extensive analysis and evaluation demonstrate that MagnifiQ is effective at high-resolution image restoration with strong visual and structural fidelity.
The use of text prompts also enables controllable restoration, as illustrated in Fig.~\ref{fig:restoration_editing}, where local details can be refined based on textual guidance, bridging restoration and image editing. This observation suggests a promising future direction: integrating advanced image editing models to further enhance high-resolution restoration. Additionally, to further improve the runtime of our ultra-high-resolution image restoration framework, a potential future direction is to explore one‑ or few‑step~\cite{yin2024improved} image restoration approaches.

\bibliographystyle{splncs04}
\bibliography{main}

\newpage
\appendix

\section*{Supplementary Material Contents}
\begingroup
\small
\setlength{\parindent}{0pt}

\noindent\textbf{\hyperref[sec:implementation_supp]{A. Implementation Details}}\\
\hspace*{1em}\hyperref[sec:implementation_details]{A.1 SDXL-based Image Restoration Backbone}\\
\hspace*{1em}\hyperref[sec:magnifiq_impl]{A.2 MagnifiQ Progressive Upscaling}\\
\hspace*{1em}\hyperref[sec:capt_generation]{A.3 Caption Generation and Local Prompts}\\[0.35em]

\noindent\textbf{\hyperref[app:evaluation_details]{B. Evaluation Details}}\\
\hspace*{1em}\hyperref[sec:magnifiq_baselines_supp]{B.1 Datasets and Baselines}\\
\hspace*{1em}\hyperref[sec:user_study_details]{B.2 4K User Study Setup}\\
\hspace*{1em}\hyperref[sec:metrics_runtime_memory]{B.3 Metrics, Runtime, and Memory Measurement}\\[0.35em]

\noindent\textbf{\hyperref[app:additional_quant]{C. Additional Quantitative Results}}\\
\hspace*{1em}\hyperref[app:div8k_fr]{C.1 DIV8K Full-Reference Evaluation}\\
\hspace*{1em}\hyperref[app:aesthetic4k_ablation]{C.2 Aesthetic-4K Ablations}\\
\hspace*{1em}\hyperref[app:padre]{C.3 PADRe Ablations}\\
\hspace*{1em}\hyperref[app:runtime_memory]{C.4 Runtime, Memory, and Deployment Analysis}\\

\noindent\textbf{\hyperref[sec:additional_qual_supp]{D. Additional Qualitative Results}}\\
\hspace*{1em}\hyperref[sec:4k_comparisons_supp]{D.1 MagnifiQ 4K and 32x Results}\\
\hspace*{1em}\hyperref[sec:4kagent_supp]{D.2 Comparison with 4KAgent}\\
\hspace*{1em}\hyperref[sec:tuning_free_supp]{D.3 Comparison with Tuning-Free High-Resolution Methods}\\
\hspace*{1em}\hyperref[sec:stage_degradation_caption_supp]{D.4 Stage, Degradation, and Captioning Ablations}\\
\hspace*{1em}\hyperref[sec:sdxl_additional_analysis_supp]{D.5 SDXL-based Restoration Results}\\
\hspace*{1em}\hyperref[sec:failure_cases_supp]{D.6 Failure Cases}\\[0.35em]

\noindent\textbf{\hyperref[sec:prompt_details]{E. Prompt Details}}\\
\hspace*{1em}\hyperref[app:global_prompts_supp]{E.1 Global Prompts Used in MagnifiQ}

\endgroup

\clearpage

\section{Implementation Details}
\label{sec:implementation_supp}

\subsection{SDXL-based Image Restoration Backbone}
\label{sec:implementation_details}

All models (SDXL-IR, SDXL-NoSA, SDXL-PD) are initialized from the pre-trained Stable Diffusion XL~\cite{podell2023sdxl}. Following InstructPix2Pix~\cite{brooks2023instructpix2pix}, the UNet input convolution is modified by inflating the input channels so that the network can condition on low-resolution or degraded latents. Aside from this shared modification,
the only architectural difference between the three variants lies in the design of the attention layers:
\begin{itemize}
  \item \textbf{SDXL-IR}: Only the input channel concatenation is applied, and all self-attention layers from SDXL are kept unchanged.
  \item \textbf{SDXL-NoSA}: Input channel concatenation is applied, and all self-attention layers are removed without replacement.
  \item \textbf{SDXL-PD}: Input channel concatenation is applied, and all self-attention layers are replaced with randomly initialized PADRe blocks (degree $i = 3$) that use a $1 \times 1$ convolution for channel mixing and a $5 \times 5$ kernel for token mixing.
\end{itemize}

All variants share the same fine-tuning setup. Each model is optimized for 30k iterations with an effective batch size of 32 on eight NVIDIA A100 GPUs, using diffusion MSE loss with a constant learning rate of $1\times10^{-5}$ and a warmup of 500 steps. The training dataset consists of FFHQ (140K images)~\cite{karras2019style}, LSDIR (85K
images)~\cite{li2023lsdir}, and Unsplash-Lite (25K images)~\cite{unsplash}. All images are resized so that the shortest side is 1024 pixels, followed by random cropping to $1024 \times 1024$. Synthetic degradations are applied using the procedure from R-ESRGAN~\cite{wang2021real}, and text prompts are generated with LLaVA~\cite{liu2024improved}.

During inference, all models use $50$ denoising steps and a text guidance scale of $4$.

\subsection{MagnifiQ Progressive Upscaling}
\label{sec:magnifiq_impl}

For the MagnifiQ pipeline, we use SDXL-PD as the underlying image restoration model. Unless otherwise specified, MagnifiQ uses $n=4$ progressive upscaling stages, corresponding to the MagnifiQ-\textbf{\blue{C}} configuration discussed in Sec.~\ref{sec:abl_configuration}. At each stage, overlapping image patches are extracted from the current upscaled image for
local captioning and PACA guidance. The number of patches is given by $\mathbf{M}=(({H-\Delta_h})/{\delta_h}+1)\times(({W-\Delta_w})/{\delta_w}+1)$, where we set $\Delta_h=\Delta_w=1024$ and $\delta_h=\delta_w=512$.

Under these settings, $\mathbf{M}=9$ at stage $n=2$ for an intermediate resolution of $2048 \times 2048$, $\mathbf{M}=25$ at stage $n=3$ for $3072 \times 3072$, and $\mathbf{M}=49$ at stage $n=4$ for $4096 \times 4096$. Across all MagnifiQ experiments, the SDXL-PD restoration model uses $50$ denoising steps and a text guidance scale of $4$.

\subsection{Caption Generation and Local Prompts}
\label{sec:capt_generation}

Text captions are used in two parts of our framework: image-level captions for training the SDXL-based restoration backbone, and patch-level captions for local guidance in MagnifiQ. For training, the LLaVA-v1.5-7B~\cite{liu2024improved} model is used to generate captions for images from FFHQ~\cite{karras2019style}, LSDIR~\cite{li2023lsdir}, and Unsplash-
Lite~\cite{unsplash}. Following SUPIR~\cite{yu2024scaling}, we use the prompt: ``\texttt{Describe this image and its style in a very detailed manner}.'' Several training examples and their corresponding text captions are shown in Fig.~\ref{fig:training_dataset}. Although we use LLaVA in our default setting, the captioning module is not tied to a specific model
and can be replaced by other image captioning or tagging models.

During MagnifiQ inference, local captions are generated for overlapping image patches and used by PACA to route patch-specific semantic guidance to the corresponding latent regions. We primarily use LLaVA because it provides detailed local descriptions that improve perceptual quality. We also evaluate DAPE~\cite{wu2024seesr} as a lightweight alternative
captioner: DAPE produces concise semantic tags and is substantially faster, but its less detailed descriptions can reduce fine-texture recovery. This trade-off is analyzed in Sec.~\ref{sec:abl_configuration} and Tab.~\ref{tab:aesthetic4k_compact}.

\section{Evaluation Details}
\label{app:evaluation_details}

\subsection{Datasets and Baselines}
\label{sec:magnifiq_baselines_supp}

To evaluate MagnifiQ, we perform $16\times$ upscaling from $256 \times 256$ degraded inputs to $4096 \times 4096$ outputs and compare against several high-resolution restoration baselines: StableSR~\cite{wang2024exploiting}, SUPIR~\cite{yu2024scaling}, MultiDiffusion~\cite{bar2023multidiffusion}, SD Upscaler~\cite{rombach2022high}, SDXL-IR, and SDXL-PD. For
StableSR, we use the official high-resolution generation pipeline. SUPIR, SDXL-IR, and SDXL-PD perform direct $16\times$ restoration by resizing the input image to $4096 \times 4096$. MultiDiffusion is evaluated by integrating SDXL-PD into its tiled diffusion framework. For SD Upscaler, we first generate a $4\times$ output using SDXL-PD and then apply SD
Upscaler for another $4\times$ upscaling step to obtain the final $4096 \times 4096$ image.

\subsection{4K User Study Setup}
\label{sec:user_study_details}

As described in Sec.~\ref{sec:user_study}, we conduct a user study to evaluate MagnifiQ against the high-resolution restoration baselines in Sec.~\ref{sec:magnifiq_baselines_supp}. We randomly select 40 images from DIV2K~\cite{agustsson2017ntire} and generate corresponding $4096 \times 4096$ outputs using all methods. For $I=40$ images and $N=7$ methods, this gives
$\binom{N}{2}\cdot I = 840$ pairwise comparisons. The study includes 10 participants, with 84 randomly assigned pairs per participant. For each comparison, the interface shows the low-quality input image and two anonymous high-resolution outputs, denoted Image A and Image B. Participants select ``Image A,'' ``Image B,'' or ``No Preference'' based on overall
image quality, local fine details, and structural consistency with the low-quality input.

We additionally conduct a focused blind A/B study against FaithDiff as also described in Sec.~\ref{app:div8k_fr}, a strong recent diffusion-based restoration baseline. This study uses 50 DIV8K 4K restoration pairs and 5 participants, resulting in 250 total ratings. MagnifiQ receives \textbf{161/250 preferences (64.4\%)}, compared with 50/250 for FaithDiff (20.0\%) and 39/250 no-preference votes (15.6\%).
Among decisive choices, MagnifiQ is preferred in \textbf{76.3\%} of cases.

\subsection{Metrics, Runtime, and Memory Measurement}
\label{sec:metrics_runtime_memory}

We report both full-reference and no-reference metrics. For full-reference evaluation, we use PSNR, SSIM, LPIPS, and DISTS to measure fidelity with respect to the reference image. Since high-resolution restoration is perceptual and can be ill-posed, we also report no-reference perceptual metrics, including MANIQA, MUSIQ, CLIP-IQA, LAION-Aesthetic, PI, and
NIQE. For patch-level evaluation, we crop corresponding local regions from the restored and reference images and compute patch-level fidelity metrics, including LPIPS, DISTS, and DINO similarity, to assess local content preservation and drift.

Runtime is measured as wall-clock inference time per image on a single NVIDIA H100 GPU. For MagnifiQ, we further decompose runtime into captioning, denoising/restoration, and other overhead, including resizing, degradation, patch extraction, bookkeeping, and I/O. Memory is reported as peak allocated GPU memory during restoration/enhancement inference. Unless
otherwise specified, memory and parameter counts exclude external captioning models, so the comparison focuses on the restoration pipelines themselves rather than the choice of captioning backend.

\section{Additional Quantitative Results}
\label{app:additional_quant}

\begin{table}[!t]
\centering
\fontsize{6.25pt}{6.45pt}\selectfont
\setlength{\tabcolsep}{0.55pt}
\renewcommand{\arraystretch}{1}
\begin{tabular}{@{}lccccc!{\vrule width 0.7pt}cc@{}}
\toprule
 & \multicolumn{5}{c!{\vrule width 0.7pt}}{\textbf{DIV8K full-reference metrics}} & \multicolumn{2}{c}{\textbf{System cost}} \\
\cmidrule(lr){2-6}\cmidrule(l){7-8}
Method & SSIM$\uparrow$ & Global LPIPS$\downarrow$ & Global DISTS$\downarrow$ & Patch LPIPS$\downarrow$ & Patch DISTS$\downarrow$ & Params & Alloc. \\
\midrule
StableSR & \textbf{.625} & .566 & .379 & .545 & .360 & 2.47 & 12.7 \\
SD Up. & .462 & .521 & .283 & .503 & .309 & 4.48 & 26.2 \\
SDXL-IR & .440 & .555 & .320 & .539 & .328 & 3.50 & 13.7 \\
SDXL-PD & .479 & .631 & .264 & .613 & .337 & 3.61 & 14.0 \\
SUPIR & \underline{.597} & .598 & .383 & .575 & .355 & 3.98 & 14.6 \\
MultiDiff. & .506 & .657 & .304 & .637 & .360 & 3.61 & 9.9 \\
4KAgent & .526 & .521 & .229 & .497 & .294 & {-} & {-} \\
FaithDiff & .520 & \textbf{.407} & \textbf{.212} & \textbf{.391} & \textbf{.267} & 3.51 & 22.4 \\
\midrule
MagnifiQ & .536 & \underline{.504} & \underline{.222} & \underline{.485} & \underline{.287} & 3.61 & 18.0 \\
\bottomrule
\end{tabular}
\caption{\scriptsize Combined table with separate metric and cost blocks. Patch SSIM omitted. Params are restoration/enhancement params (B); Alloc. is peak allocated memory (GB), excluding captioning models.}
\label{tab:div8k_fr_rebuttal}
\label{tab:system_cost_rebuttal}
\end{table}

\subsection{DIV8K Full-Reference Evaluation}
\label{app:div8k_fr}

Full-reference metrics provide a complementary but incomplete view of perceptual restoration quality. As discussed by SUPIR~\cite{yu2024scaling}, pixel-aligned scores can favor conservative or over-smoothed reconstructions and may not always agree with human judgments; therefore, no-reference perceptual metrics and user studies are also important for
evaluating high-quality restoration. For completeness, we report DIV8K full-reference results for FaithDiff, 4KAgent, StableSR, SUPIR, and other baselines in Tab.~\ref{tab:div8k_fr_rebuttal}. FaithDiff achieves the best LPIPS/DISTS scores, indicating strong reference-level fidelity under these metrics, while MagnifiQ ranks second on global LPIPS, global DISTS,
patch LPIPS, and patch DISTS. MagnifiQ also outperforms 4KAgent on all reported full-reference metrics, including SSIM, global/patch LPIPS, and global/patch DISTS. Since FaithDiff is the strongest baseline under the full-reference perceptual metrics, we additionally conduct a blind A/B user study comparing MagnifiQ and FaithDiff on 50 DIV8K image pairs with 5
participants. MagnifiQ receives \textbf{161/250 preferences (64.4\%)}, compared with 50/250 for FaithDiff (20.0\%) and 39/250 no-preference votes (15.6\%); among decisive choices, MagnifiQ is preferred in \textbf{76.3\%} of cases. These results suggest that, while FaithDiff is favored by full-reference perceptual distances, human raters more often prefer
MagnifiQ's restored 4K outputs.

\begin{table}[!t]
\centering
\fontsize{5.65pt}{5.85pt}\selectfont
\setlength{\tabcolsep}{0.35pt}
\renewcommand{\arraystretch}{1}
\begin{tabular}{@{}lc cccc@{}}
\toprule
Ablation & Setting & MANIQA$\uparrow$ & CLIP-IQA$\uparrow$ & Aesthetic$\uparrow$ & Time cap./den./oth. \\
\midrule
Degradation & \setdeg{none} & .305 & .577 & 4.05 & 377 (\tc{243}/\td{114}/\toh{20}) \\
\midrule
PACA prompt & \setpaca{empty} & .300 & .451 & 3.94 & 95 (\tc{0}/\td{71}/\toh{24}) \\
PACA prompt & \setpaca{global} & .356 & .601 & 4.44 & 96 (\tc{0}/\td{72}/\toh{24}) \\
\midrule
Captioner & \setcap{DAPE} & .340 & .590 & 4.40 & 158 (\tc{3}/\td{130}/\toh{25}) \\
\midrule
Stages & \setstage{2} & .335 & .617 & 4.43 & 262 (\tc{167}/\td{82}/\toh{13}) \\
Stages & \setstage{3} & .347 & .615 & 4.44 & 314 (\tc{200}/\td{98}/\toh{16}) \\
Stages & \setstage{5} & \textbf{.368} & .620 & \textbf{4.85} & 571 (\tc{357}/\td{178}/\toh{36}) \\
\midrule
Full model & \setdeg{512}/\setpaca{local}/\setcap{LLaVA}/\setstage{S4} & .363 & \textbf{.622} & 4.84 & 393 (\tc{243}/\td{114}/\toh{36}) \\
\bottomrule
\end{tabular}
\caption{\scriptsize 4K aesthetic ablation. Full: degradation 512, local PACA, LLaVA, Stage 4. Time split is caption/denoise/other (\tc{blue}/\td{orange}/\toh{gray}).}
\label{tab:aesthetic4k_ablation_rebuttal}
\end{table}

\subsection{4K Ablations}
\label{app:aesthetic4k_ablation}

We further analyze the effect of patch-aware cross-attention (PACA) and the fidelity--creativity trade-off introduced by progressive 4K restoration. In Tab.~\ref{tab:aesthetic4k_ablation_rebuttal}, we isolate the text-conditioning strategy under the same 4K aesthetic evaluation setting by comparing three PACA prompt variants: an empty prompt, a single global
prompt, and local patch-level prompts routed through PACA. Moving from empty prompts to a global prompt improves MANIQA from .300 to .356 and CLIP-IQA from .451 to .601, showing that semantic text guidance is important for perceptual restoration. Replacing the global prompt with local PACA prompts further improves MANIQA to .363 and CLIP-IQA to .622,
indicating that spatially matched local descriptions better guide high-resolution detail synthesis than a single image-level description. The same ablation also supports the use of the light degradation/resampling step before restoration: compared with no degradation, the full degradation-512 setting improves MANIQA/CLIP-IQA/Aesthetic from .305/.577/4.05
to .363/.622/4.84.

\noindent Progressive restoration can also introduce local content drift because high-resolution restoration from severely degraded inputs is inherently ill-posed. To quantify this effect, we compare intermediate-stage outputs on the 50-image aesthetic subset. We resize the 4K outputs to 2048 and compute patch-level fidelity metrics against the corresponding
ground-truth patches. From Stage~2 to Stage~5, patch LPIPS/DISTS worsen from .377/.229 to .411/.250, and patch DINO similarity decreases from .639 to .588. The qualitative examples in Supp. Fig.~38 show the same trend: later stages add sharper and more plausible local details, but can gradually alter fine textures or identity-sensitive regions. Thus, the
number of stages acts as a controllable fidelity--creativity knob. Fewer stages better preserve local content and identity, while more stages increase perceptual detail at the cost of higher runtime and potential drift. In our default setting, Stage~4 provides a practical balance: it achieves strong no-reference perceptual scores while avoiding the additional
runtime and drift observed at Stage~5.

\noindent PACA itself is a training-free inference-time routing mechanism. It does not modify the SDXL-PD weights and can be disabled at inference by falling back to empty or global prompts. At each progressive stage, local patches are re-captioned from the current image estimate, and each caption is routed only to its corresponding latent region through
cross-attention. This keeps the low-resolution/current image as the primary spatial condition, while using text only to modulate local detail generation. Image-anchored latent initialization provides an additional possible control knob for reducing content drift, and we leave a systematic study of such drift-control strategies to future work.

\subsection{PADRe Ablations}
\label{app:padre}

PADRe configurations used in the SDXL-PD model are evaluated on the DIV2K~\cite{agustsson2017ntire} dataset with degradations by varying the PADRe degree $i$ and the token-mixing kernel size $k \times k$, as shown in Tab.~\ref{tab:qual_1k_padre_ablations}. All configurations are trained using the same setup described in Sec.~\ref{sec:implementation_details}.
The SDXL-PD model with $i = 3$ and a $5 \times 5$ kernel achieves the best overall performance across most metrics.

\begin{table}[!t]
    \centering
    \scriptsize
    \resizebox{\columnwidth}{!}{%
    \begin{tabular}{lcccccccc}
        \toprule
        Method & PSNR $\uparrow$ & SSIM $\uparrow$ & LPIPS $\downarrow$ & MQ $\uparrow$ & MSQ $\uparrow$ & PI $\downarrow$ & NIQE $\downarrow$ & Runtime (s)\\ 
        \midrule
        PADRe $i=2,k=5$ & 19.06 & 0.4490 & 0.\textbf{4461} & 0.3490 & 68.006 & \textbf{3.6626} & \textbf{3.8714} & 3.06 \\
        PADRe $i=2,k=7$ & 19.12 & 0.4487 & \underline{0.4497} & \underline{0.3540} & \underline{68.434} & \underline{3.6695} & 3.8751 & 3.06 \\
        PADRe $i=2,k=11$ & \underline{19.19} & \underline{0.4545} & 0.4498 & 0.3340 & 67.335 & 3.7535 & \underline{3.8717} & 7.18\\
        \rowcolor[gray]{0.9} PADRe $i=3,k=5$ & \textbf{19.26} & \textbf{0.4550} & 0.4558 & \textbf{0.3726} & \textbf{68.440} & 3.7030 & 4.0279 & 3.55\\
        PADRe $i=3,k=7$ & 19.18 & 0.4542 & 0.4613 & 0.3392 & 67.509 & 3.8337 & 4.1048 & 3.58\\
        \bottomrule
    \end{tabular}
    }
    \caption{Evaluation of the SDXL-PD model with different PADRe configurations by varying degree $i$ and token-mixing kernel size $k \times k$ on the Div2K~\cite{agustsson2017ntire} dataset with degradations. The best-performing setup ($i = 3$, $k = 5$) is adopted in our MagnifiQ framework for high-resolution image restoration via progressive upscaling.}
    \label{tab:qual_1k_padre_ablations}
\end{table}

\subsection{Runtime, Memory, and Deployment Analysis}
\label{app:runtime_memory}

We also report the computational cost of MagnifiQ to clarify its practical trade-offs. The full-quality configuration is not real-time: as shown in Tab.~\ref{tab:aesthetic4k_ablation_rebuttal}, the default 4K setting takes 393s per image, consisting of \tc{243}s for captioning, \td{114}s for denoising/restoration, and \toh{36}s for resizing, degradation, patch
bookkeeping, and I/O overhead. This breakdown shows that local caption generation is the dominant cost in the current implementation, rather than the SDXL-PD restoration backbone alone.

\noindent The pipeline exposes several speed--quality trade-offs. Replacing LLaVA with the lighter DAPE captioner reduces the captioning time from \tc{243}s to \tc{3}s and the total runtime from 393s to 158s, while reducing MANIQA/CLIP-IQA/Aesthetic from .363/.622/4.84 to .340/.590/4.40. Similarly, reducing the number of progressive stages lowers both
captioning and denoising cost: Stage~2 takes 262s, Stage~3 takes 314s, Stage~4 is the default 393s setting, and Stage~5 increases runtime to 571s. These results indicate that the method can be configured either for higher perceptual quality or lower latency depending on the application.

\noindent There are also clear acceleration opportunities. Since denoising accounts for \td{114}s in the default configuration, a distilled 50$\rightarrow$5 step denoiser would reduce this component to roughly \td{11}s under linear scaling. Combined with a fast captioner such as DAPE and the current \toh{36}s overhead, this would yield an estimated runtime of
approximately 50s per 4K image while preserving the same progressive, $32\times$-capable pipeline design.

\noindent Tab.~\ref{tab:system_cost_rebuttal} reports restoration/enhancement parameters and peak allocated memory, excluding captioning models so that the comparison focuses on the image restoration pipelines themselves. MagnifiQ uses the same 3.61B-parameter SDXL-PD restoration backbone as SDXL-PD and MultiDiffusion, while its peak allocated memory is
18.0GB. This is higher than SDXL-PD due to progressive 4K inference and PACA bookkeeping, but remains comparable to other large restoration systems such as SUPIR and FaithDiff. Overall, MagnifiQ is best viewed as a quality-oriented high-resolution restoration system, with lower-latency variants available through captioner choice, stage count, and future
denoising distillation.

\section{Additional Qualitative Results}
\label{sec:additional_qual}

\subsection{MagnifiQ 4K and 32x Results}
\label{sec:4k_comparisons_supp}

\begin{figure*}
    \centering
    \includegraphics[width=\linewidth]{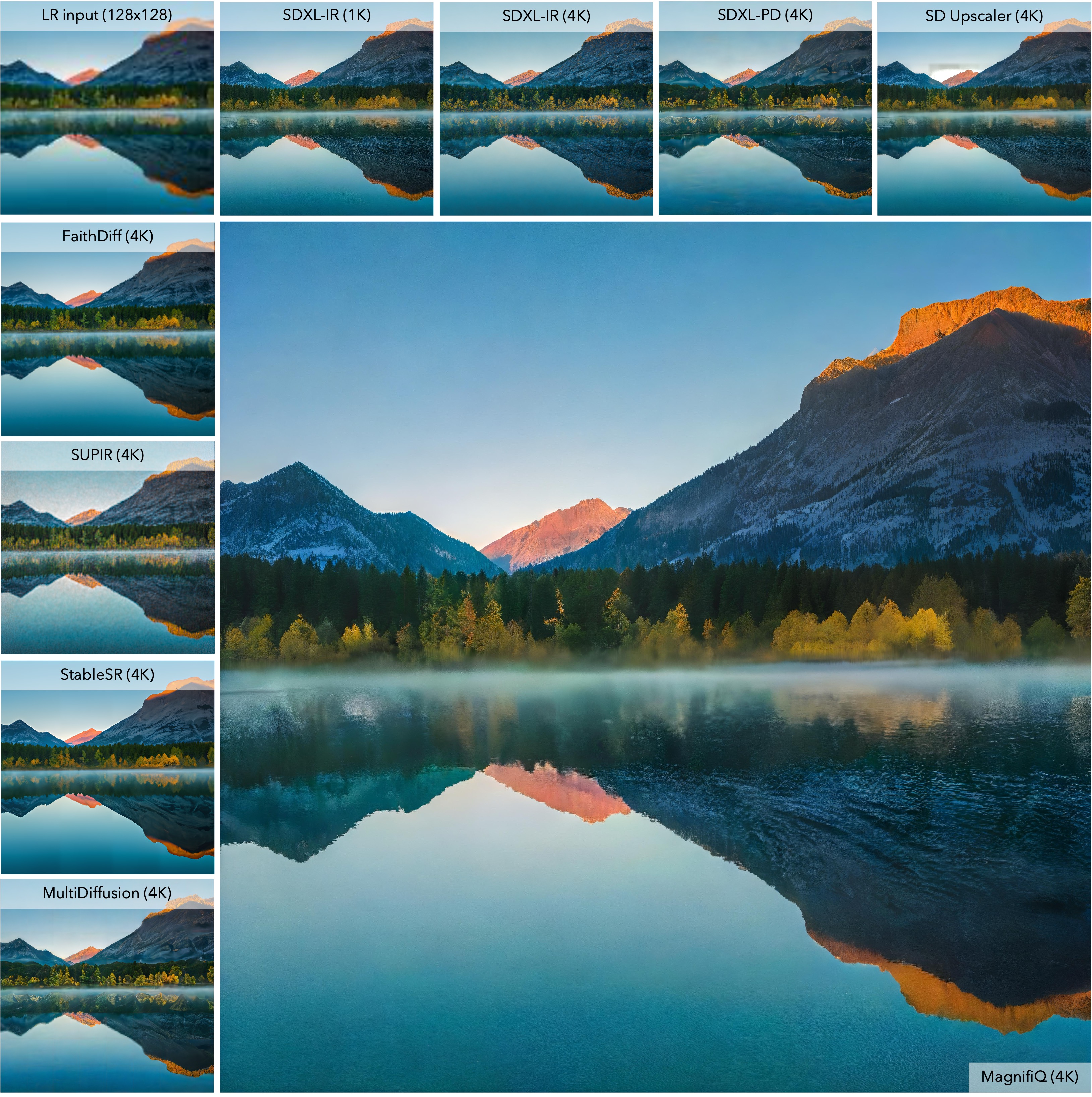}
    \caption{$128 \times 128 \rightarrow 4096 \times 4096$ ($32\times$) images from StableSR, SUPIR, FaithDiff, SDXL-IR, SDXL-PD, SD Upscaler (with SDXL-PD), MultiDiffusion (with SDXL-PD), and MagnifiQ are compared. The MagnifiQ framework restores images to high-resolution with realistic details and superior visual quality. \textbf{Zoom in to view fine details.}}
    \label{fig:qual_4k_baselines_supp_1}
\end{figure*}

\paragraph{4K comparisons.}
  In addition to Fig.~\ref{fig:qual_4k_v1}, Fig.~\ref{fig:qual_4k_baselines_supp_1} shows additional $128 \times 128 \rightarrow 4096 \times 4096$ ($32\times$) comparisons with high-resolution restoration baselines discussed in Sec.~\ref{sec:magnifiq_baselines_supp}.

\begin{figure*}[!h]
    \centering
    \includegraphics[width=0.95\linewidth]{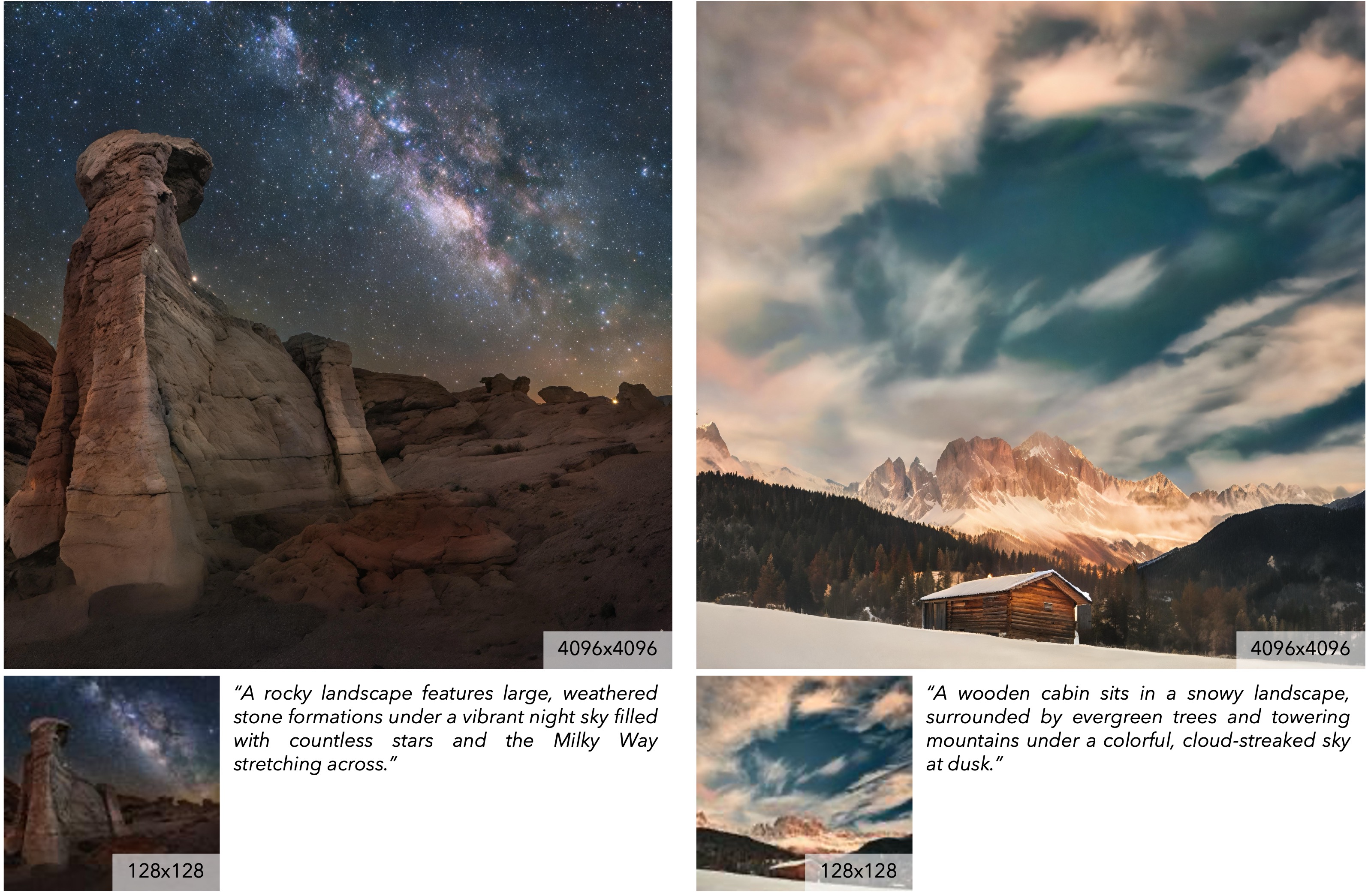}
    \includegraphics[width=0.95\linewidth]{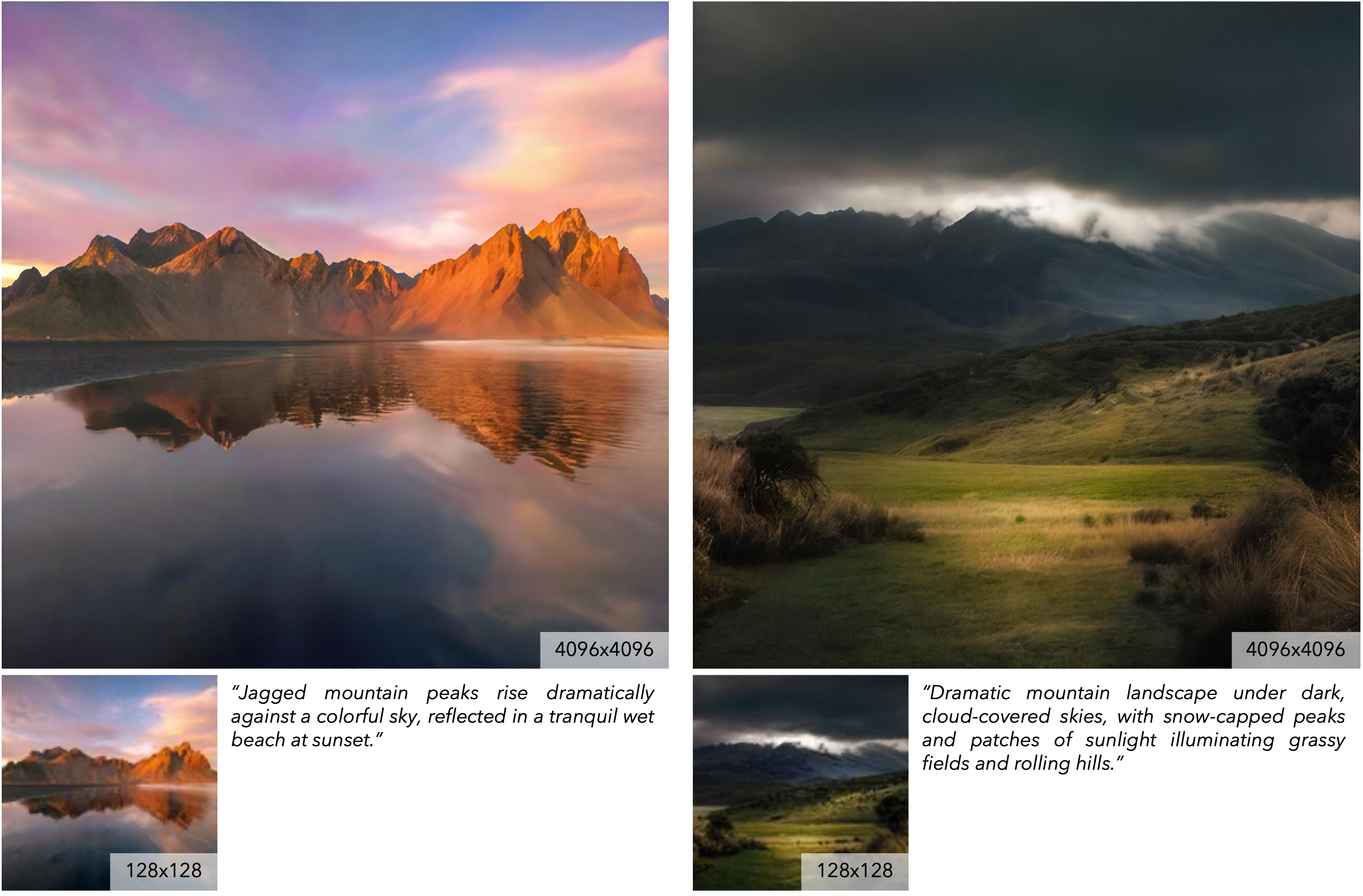}
    \caption{$128 \times 128 \rightarrow 4096 \times 4096$ ($32\times$) images generated using the MagnifiQ framework. Both the low-quality input image and the global text prompt are shown. \textbf{Zoom in to view fine details.}}
    \label{fig:aesthetic_32x_v1}
\end{figure*}

\begin{figure*}[!h]
    \centering
    \includegraphics[width=0.95\linewidth]{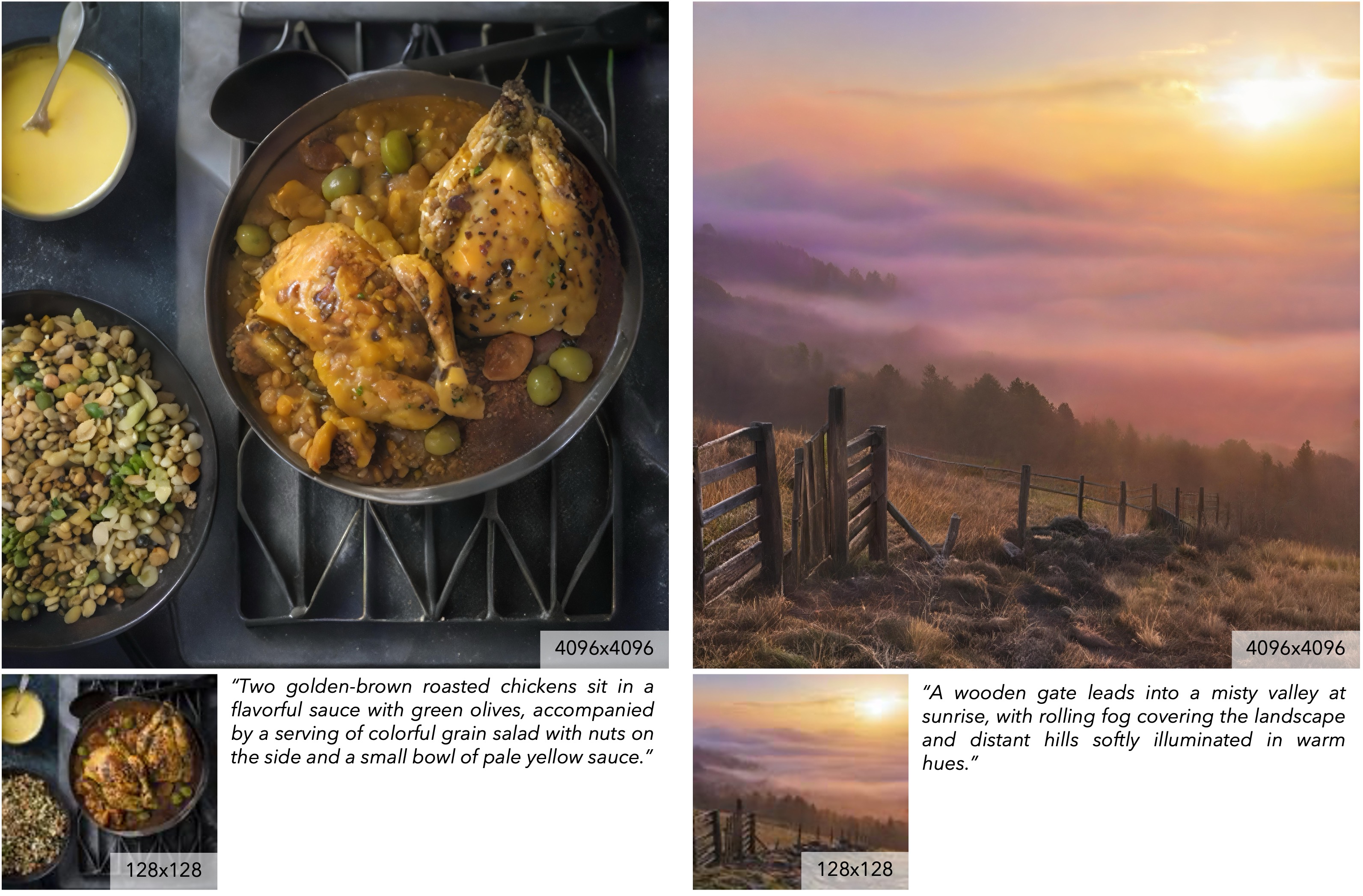}
    \includegraphics[width=0.95\linewidth]{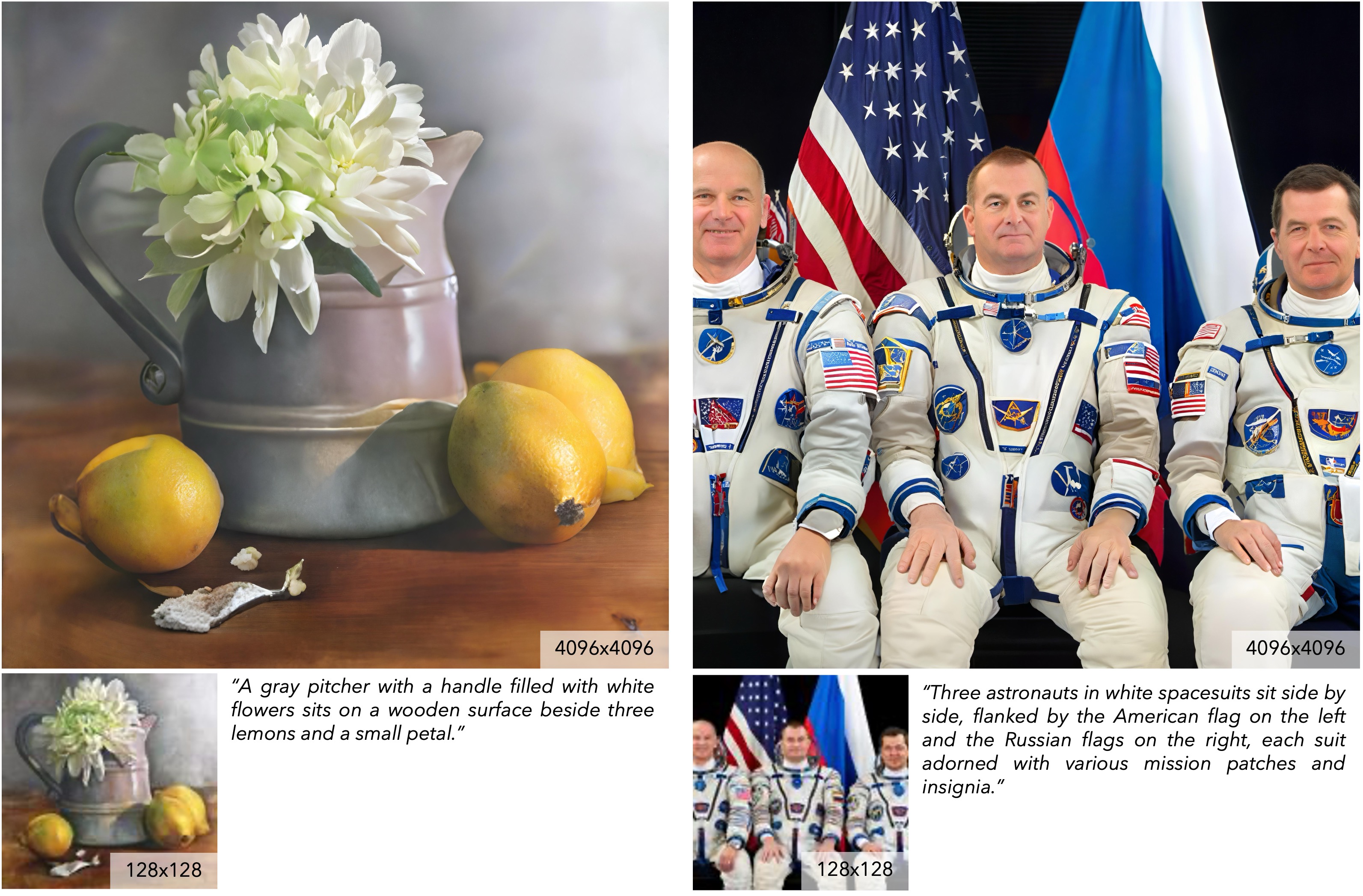}
    \caption{$128 \times 128 \rightarrow 4096 \times 4096$ ($32\times$) images generated using the MagnifiQ framework. Both the low-quality input image and the global text prompt are shown. \textbf{Zoom in to view fine details.}}
    \label{fig:aesthetic_32x_v5}
\end{figure*}

 \paragraph{MagnifiQ 4K Results.}
  Additional MagnifiQ results are shown in Figs.~\ref{fig:aesthetic_32x_v1}  and \ref{fig:aesthetic_32x_v5}. These examples are generated from $128 \times 128$ low-quality inputs to $4096 \times 4096$ outputs, demonstrating robust $32\times$ restoration across diverse degraded images.

 \subsection{Comparison with 4KAgent}
  \label{sec:4kagent_supp}

\begin{figure*}
    \centering
    \includegraphics[width=\linewidth]{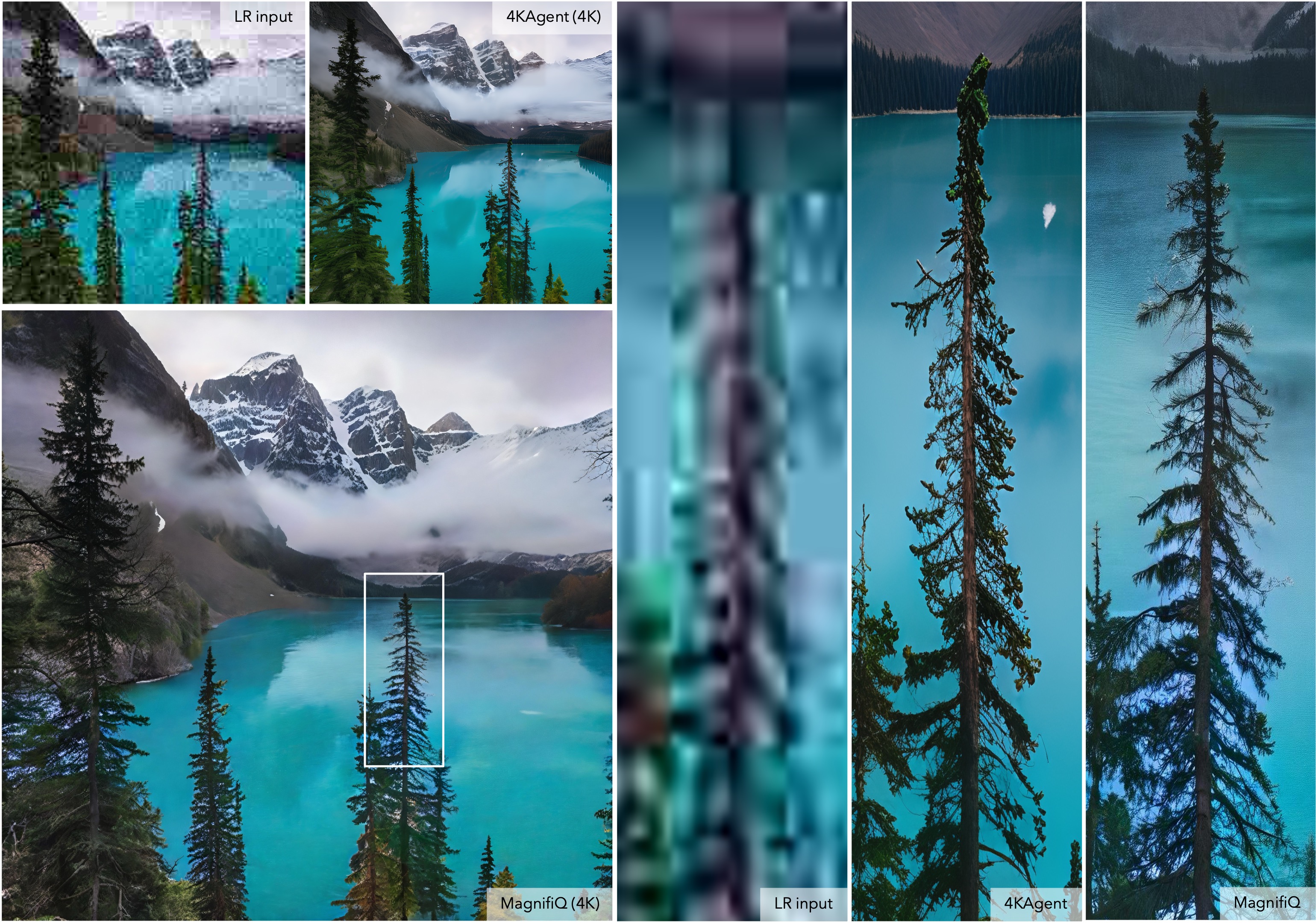}
    \caption{Comparing MagnifiQ with 4KAgent~\cite{zuo20254kagent} on low-quality images from DIV2K50~\cite{zuo20254kagent}. \textbf{Zoom in to view fine details.}}
    \label{fig:4kagent_div2k50_1}
\end{figure*}

\begin{figure*}
    \centering
    \includegraphics[width=\linewidth]{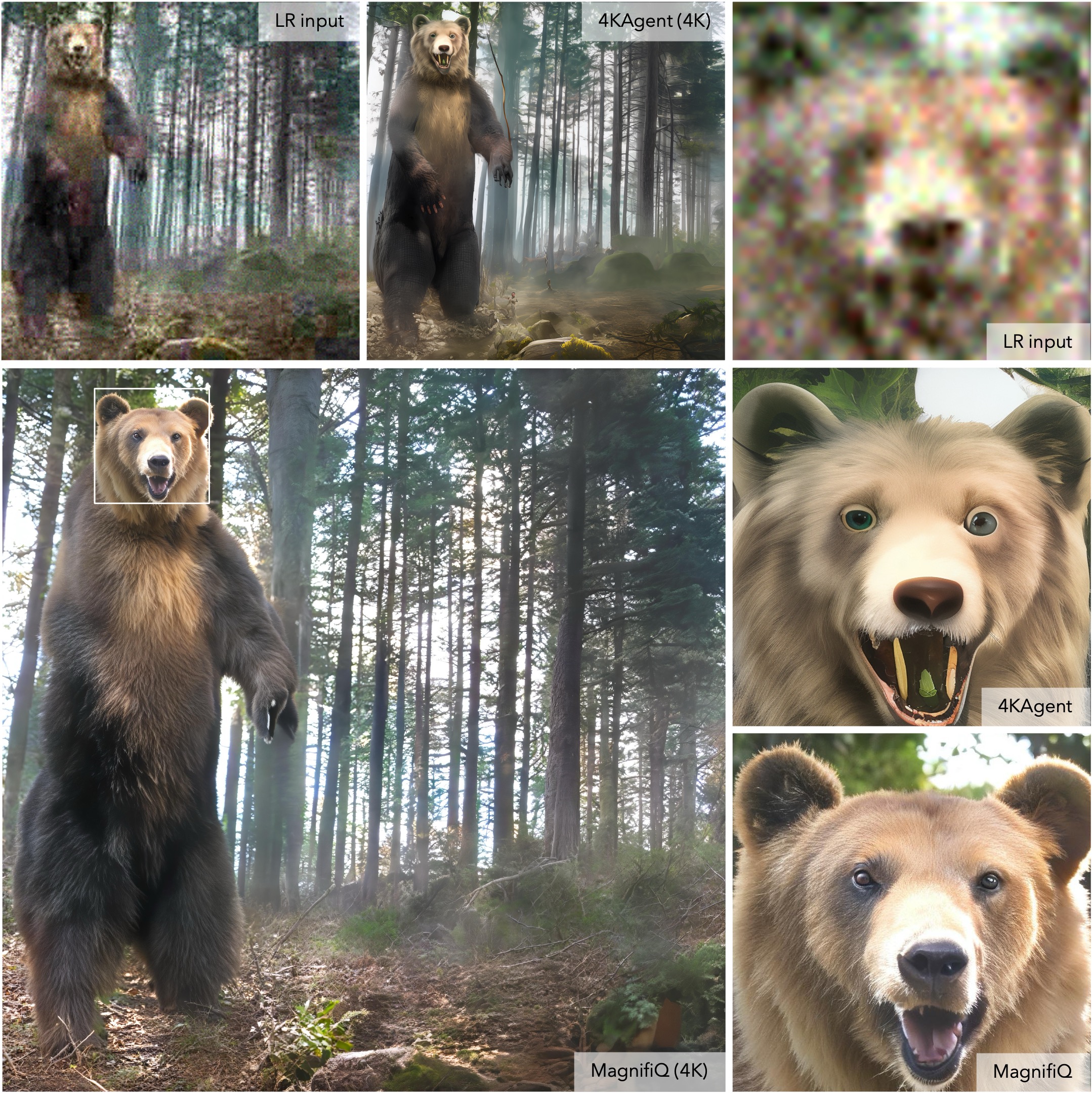}
    \caption{Comparing MagnifiQ with 4KAgent~\cite{zuo20254kagent} on low-quality images from DIV2K50~\cite{zuo20254kagent}. \textbf{Zoom in to view fine details.}}
    \label{fig:4kagent_div2k50_2}
\end{figure*}

\begin{figure*}
    \centering
    \includegraphics[width=\linewidth]{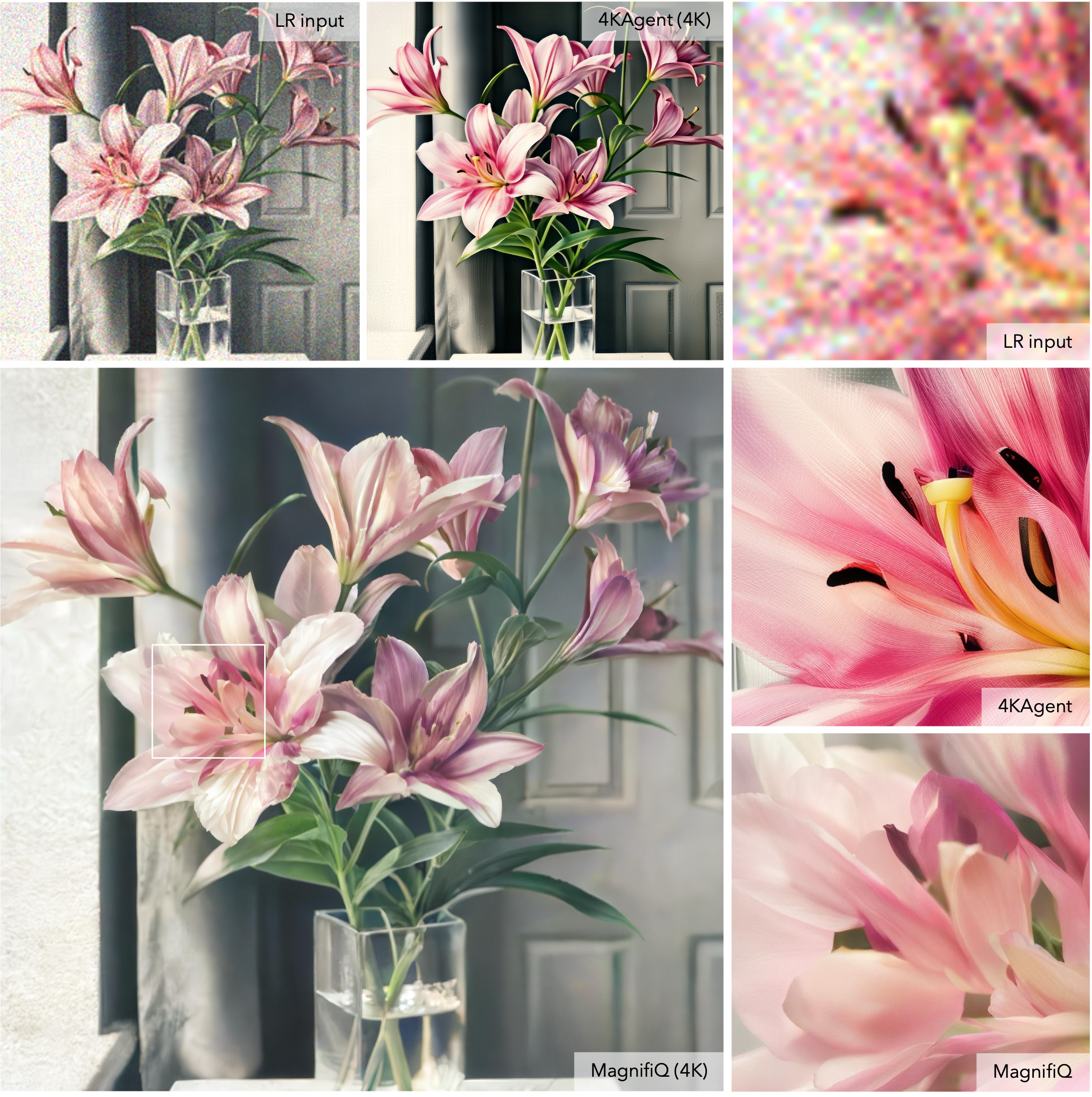}
    \caption{Comparing MagnifiQ with 4KAgent~\cite{zuo20254kagent} on low-quality images from DIV2K50~\cite{zuo20254kagent}. \textbf{Zoom in to view fine details.}}
    \label{fig:4kagent_div2k50_4}
\end{figure*}

\begin{figure*}
    \centering
    \includegraphics[width=\linewidth]{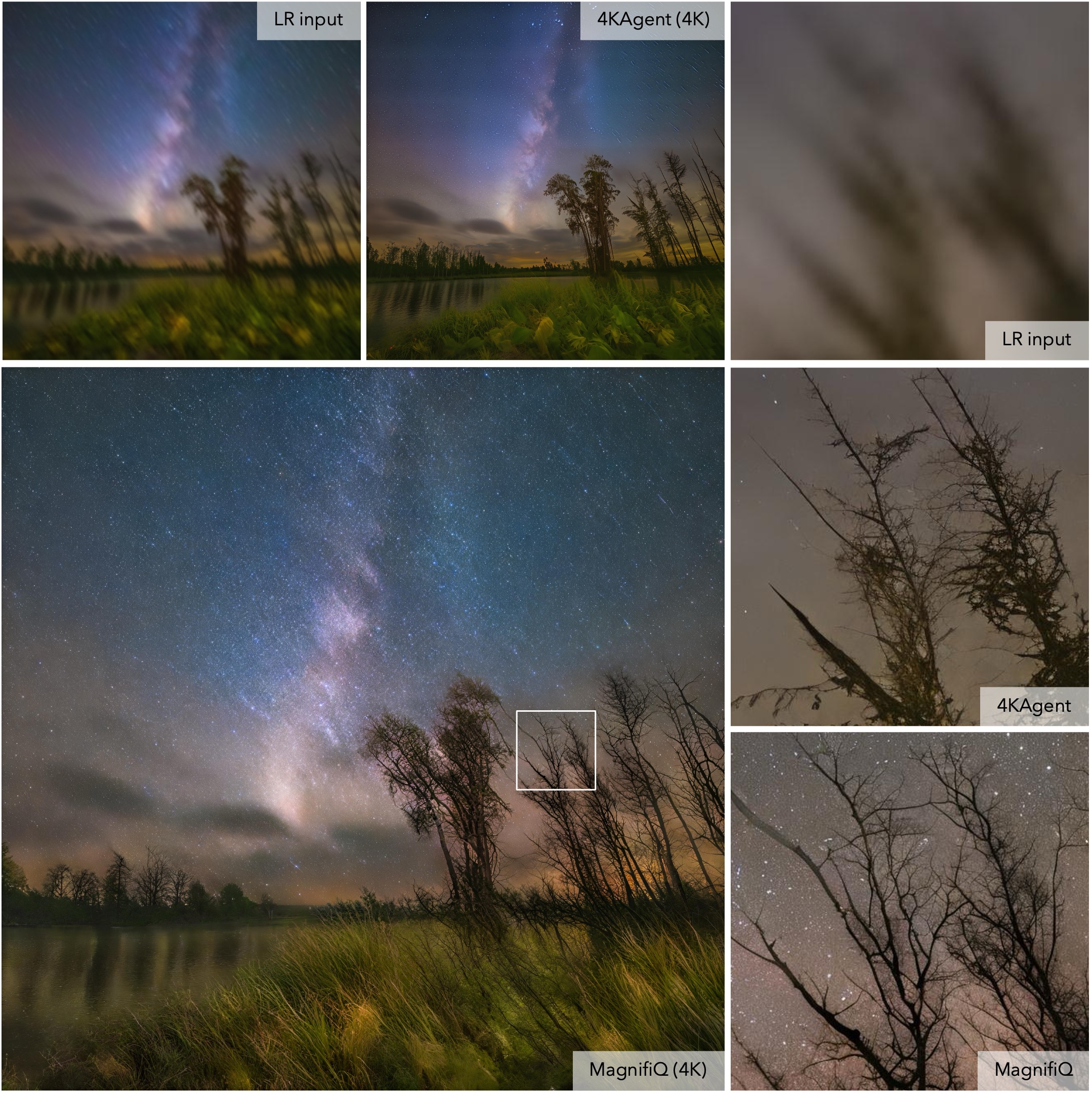}
    \caption{Comparing MagnifiQ with 4KAgent~\cite{zuo20254kagent} on high-resolution image restoration. \textbf{Zoom in to view fine details.}}
    \label{fig:4kagent_4}
\end{figure*}

\begin{figure*}
    \centering
    \includegraphics[width=\linewidth]{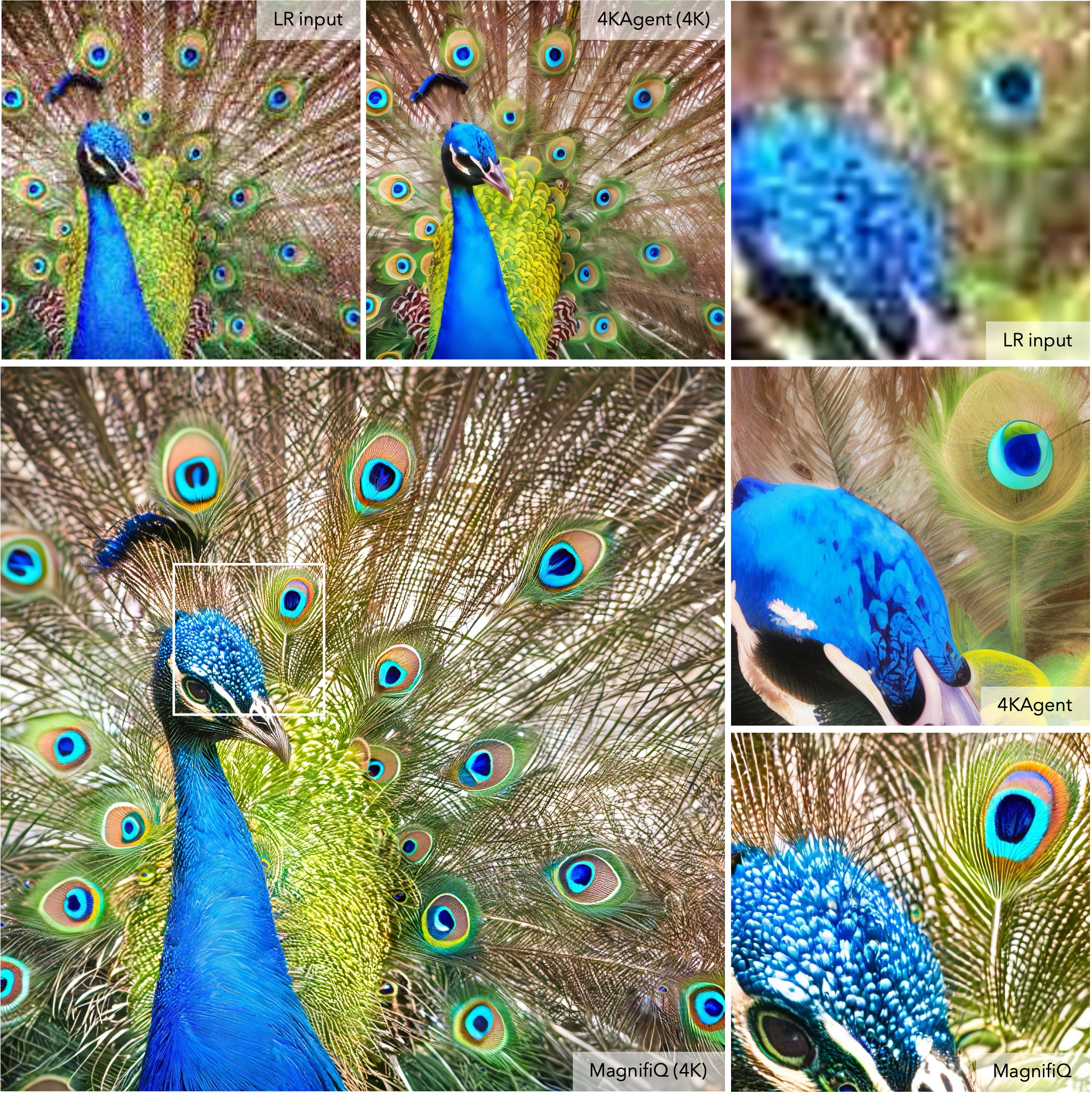}
    \caption{Comparing MagnifiQ with 4KAgent~\cite{zuo20254kagent} on high-resolution image restoration. \textbf{Zoom in to view fine details.}}
    \label{fig:4kagent_1}
\end{figure*}

  High-resolution restored images from MagnifiQ are compared with those from 4KAgent~\cite{zuo20254kagent}. 4KAgent uses agentic planning with multiple prior restoration tools, whereas MagnifiQ uses a simpler progressive diffusion-restoration pipeline. As shown in Figs.~\ref{fig:4kagent_div2k50_1}--\ref{fig:4kagent_1}, MagnifiQ produces sharper local details and more coherent restored textures.

\subsection{Comparison with Tuning-Free High-Resolution Methods}
  \label{sec:tuning_free_supp}

\begin{figure*}
    \centering
    \includegraphics[width=\linewidth]{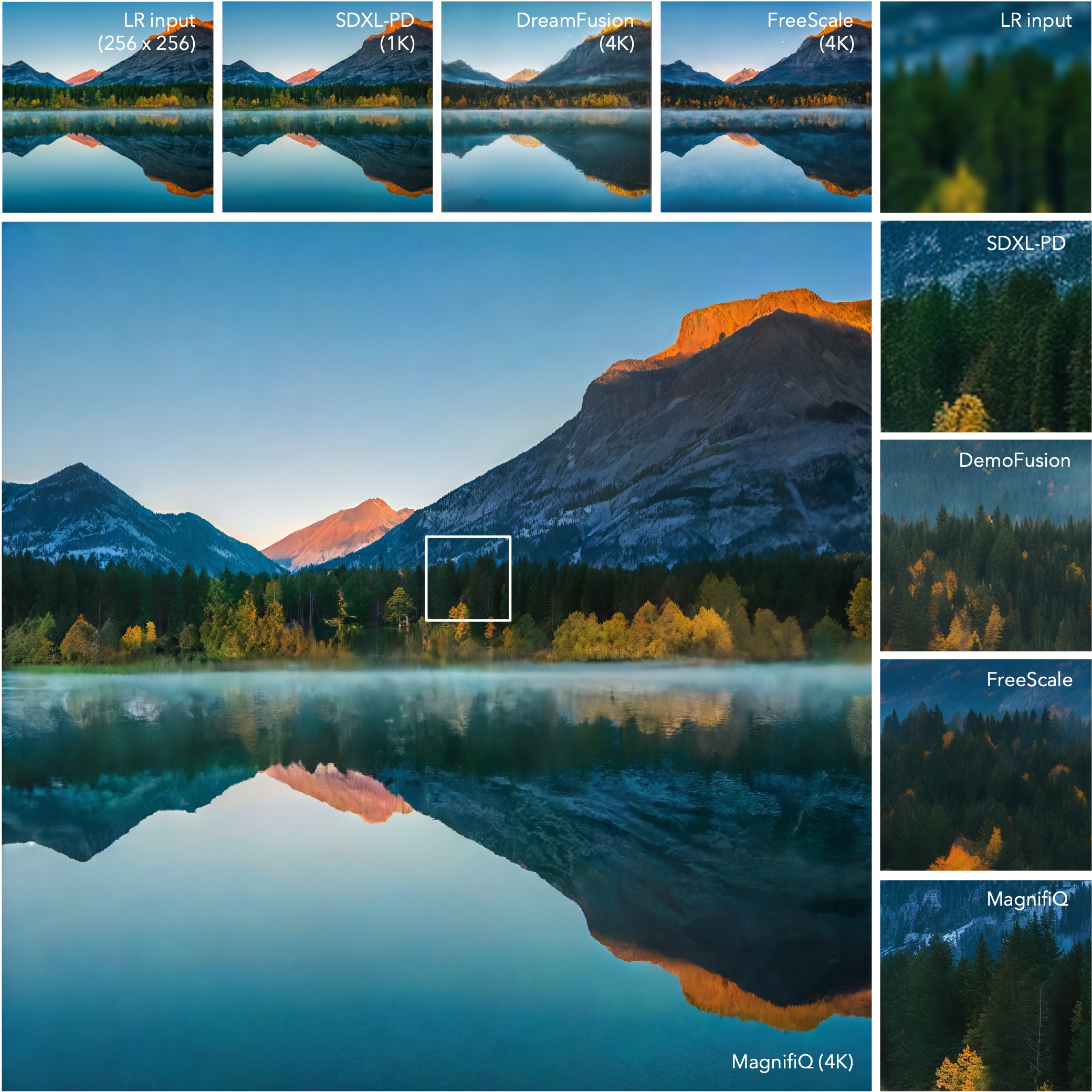}
    \caption{MagnifiQ is compared with other tuning-free high-resolution image generation approaches~\cite{du2024demofusion,qiu2024freescale}. The $1024 \times 1024$ output from SDXL-PD is used as the input for all three approaches to generate high-resolution $4096 \times 4096$ images. The prompt employed for SDXL-PD is:~\texttt{Majestic mountains rise against a clear sky, their peaks bathed in golden light, while a serene lake reflects their image surrounded by vibrant autumn foliage and mist hovering over the water.} \textbf{Zoom in to view fine details.}}
    \label{fig:magnifiq_vs_t2i_1}
\end{figure*}

\begin{figure*}
    \centering
    \includegraphics[width=\linewidth]{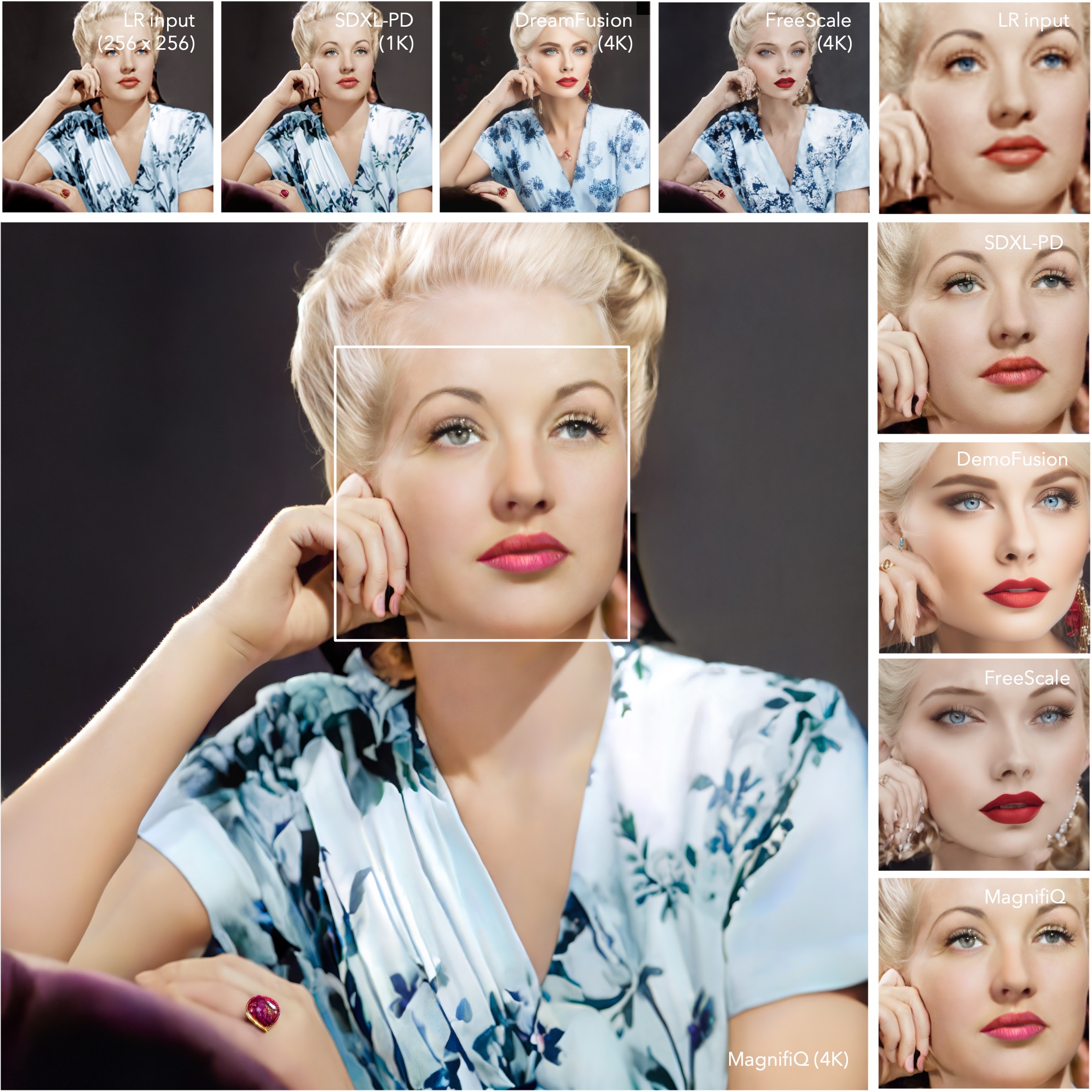}
    \caption{MagnifiQ is compared with other tuning-free high-resolution image generation approaches~\cite{du2024demofusion,qiu2024freescale}. The $1024 \times 1024$ output from SDXL-PD is used as the input for all three approaches to generate high-resolution $4096 \times 4096$ images. The prompt employed for SDXL-PD is:~\texttt{Blonde woman with elegantly styled hair and striking blue eyes rests her chin on her hand, wearing a floral blue dress and a red ring.} \textbf{Zoom in to view fine details.}}
    \label{fig:magnifiq_vs_t2i_2}
\end{figure*}

\begin{figure*}
    \centering
    \includegraphics[width=\linewidth]{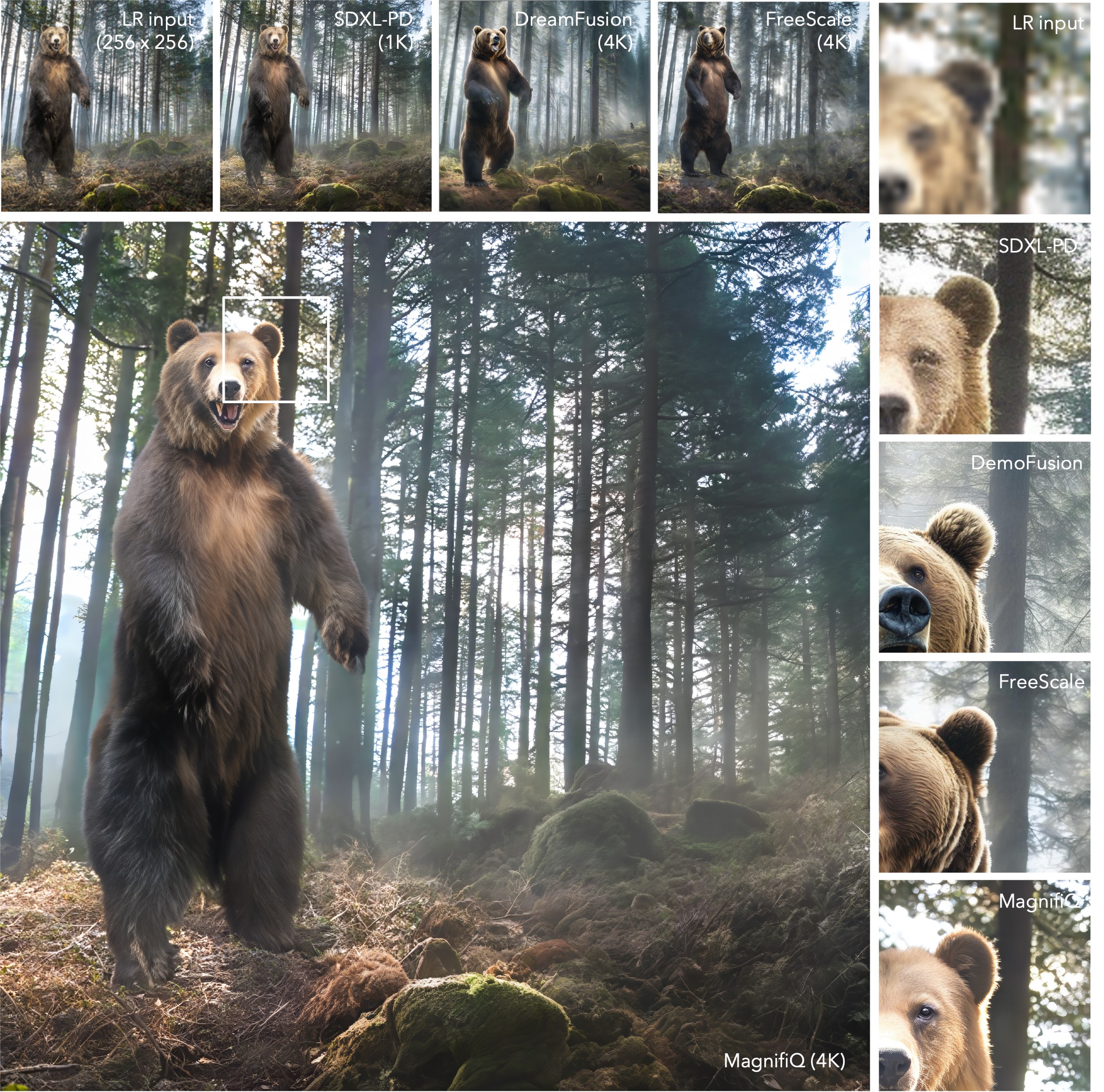}
    \caption{MagnifiQ is compared with other tuning-free high-resolution image generation approaches~\cite{du2024demofusion,qiu2024freescale}. The $1024 \times 1024$ output from SDXL-PD is used as the input for all three approaches to generate high-resolution $4096 \times 4096$ images. The prompt employed for SDXL-PD is:~\texttt{A towering brown bear stands on its hind legs in a misty forest, surrounded by tall trees and moss-covered rocks, with sunlight filtering through the foliage. } \textbf{Zoom in to view fine details.}}
    \label{fig:magnifiq_vs_t2i_3}
\end{figure*}

\begin{figure*}
    \centering
    \includegraphics[width=\linewidth]{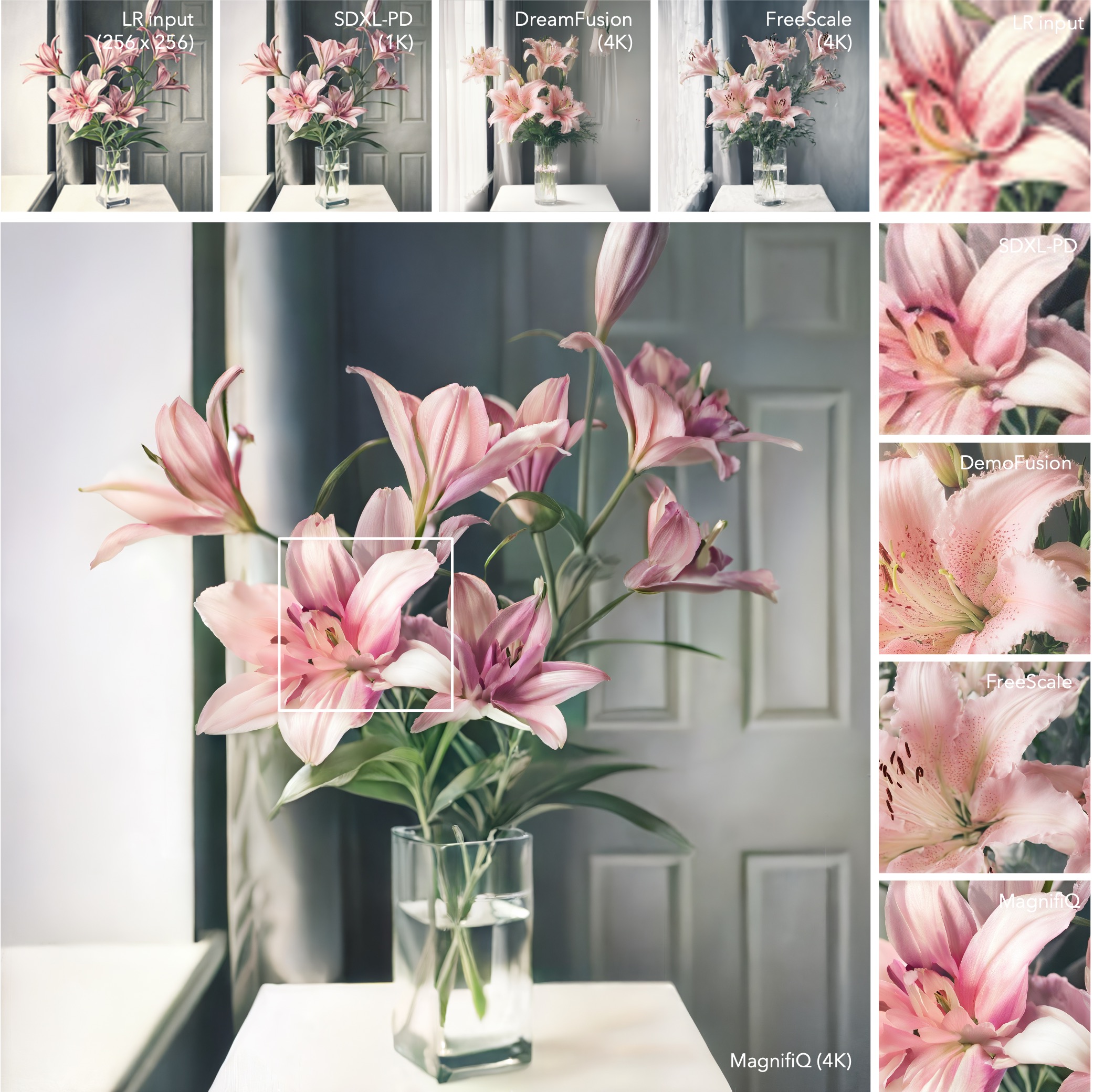}
    \caption{MagnifiQ is compared with other tuning-free high-resolution image generation approaches~\cite{du2024demofusion,qiu2024freescale}. The $1024 \times 1024$ output from SDXL-PD is used as the input for all three approaches to generate high-resolution $4096 \times 4096$ images. The prompt employed for SDXL-PD is:~\texttt{A bouquet of pink lilies is arranged in a clear vase filled with water, placed on a white side table next to a partially opened curtain and a closed door with vertical paneling, illuminated by soft light.} \textbf{Zoom in to view fine details.}}
    \label{fig:magnifiq_vs_t2i_4}
\end{figure*}

  We compare MagnifiQ with tuning-free text-to-image high-resolution generation methods, including DemoFusion~\cite{du2024demofusion} and FreeScale~\cite{qiu2024freescale} in Figs.~\ref{fig:magnifiq_vs_t2i_1}--~\ref{fig:magnifiq_vs_t2i_4}. For all methods, we first generate a $1024 \times 1024$ restored image using SDXL-PD and use it as the input for high-resolution generation. Although DemoFusion and FreeScale can synthesize 4K outputs, they often alter the structure of the conditioning image during denoising. In contrast, MagnifiQ preserves the conditioning image and denoises only the concatenated noisy latents, resulting in better structural fidelity to the input.

\begin{figure}
    \centering
    \includegraphics[width=\linewidth]{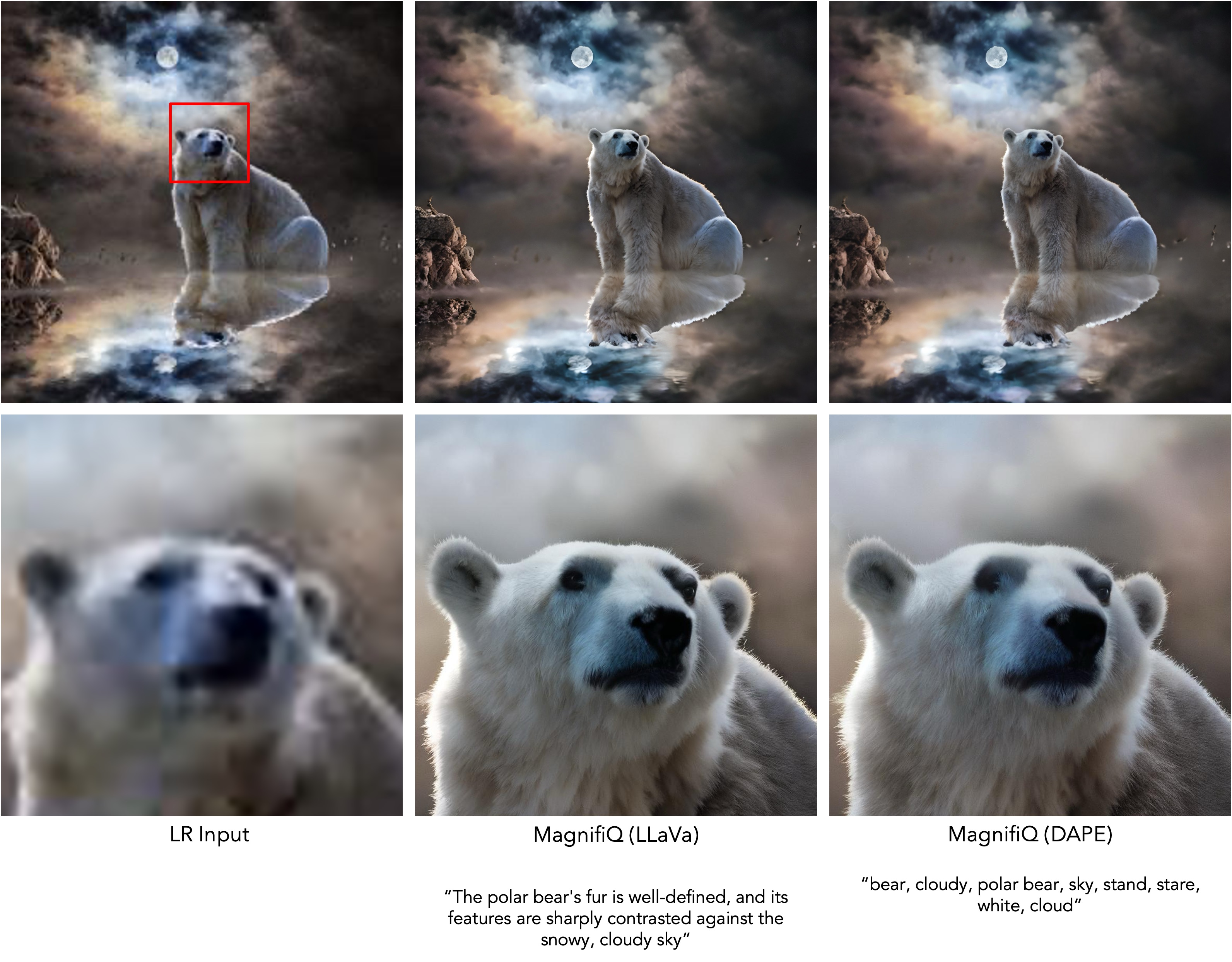}
    \caption{Comparison of MagnifiQ outputs using LLaVA or DAPE as the local patch captioning model with stride $\delta_h=\delta_w=512$.}
    \label{fig:llava_vs_dape_supp}
\end{figure}

\subsection{Stage, Degradation, and Captioning Ablations}
\label{sec:stage_degradation_caption_supp}

  \paragraph{Captioning model.}
  Fig.~\ref{fig:llava_vs_dape_supp} compares MagnifiQ outputs using LLaVA and DAPE as local patch captioners with stride $\delta_h=\delta_w=512$. LLaVA produces more detailed local descriptions, while DAPE produces concise tags and is substantially faster. DAPE therefore reduces runtime, as shown in Tab.~\ref{tab:aesthetic4k_compact}, but can miss fine
  details such as fur texture.

\begin{figure*}[!h]
    \centering
    \includegraphics[width=\linewidth]{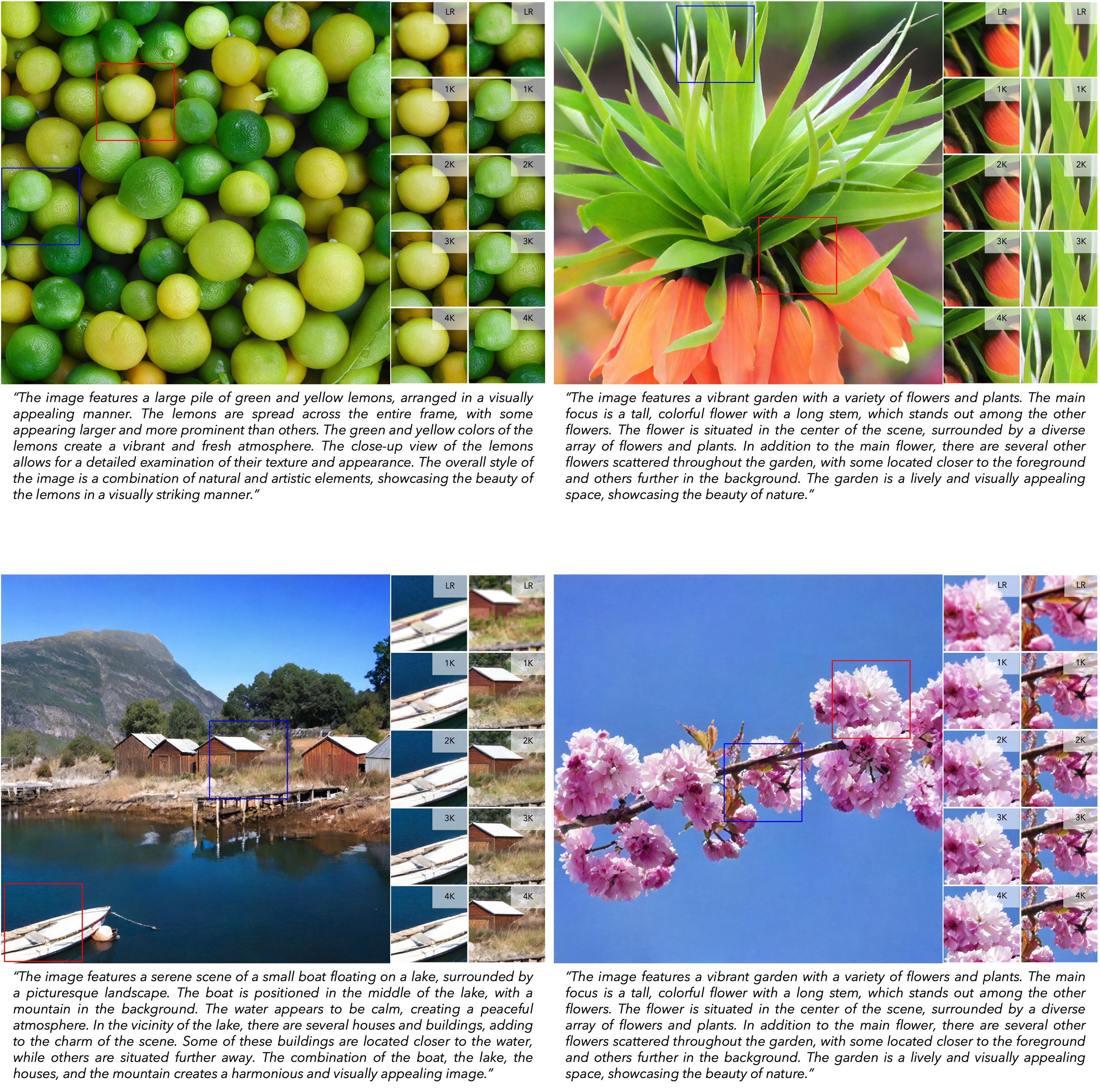}
    \caption{All intermediate restored images at each stage for $n = 4$ are compared. Using $n = 4$ demonstrates a clear, gradual improvement in generating the final $4096 \times 4096$ image. The caption shown corresponds to the global caption used at stage $n = 1$. \textbf{Zoom in to view fine details.}}
    \label{fig:div2k_16x_stages_v1}
\end{figure*}

\begin{figure}
    \centering
    \includegraphics[width=0.8\linewidth]{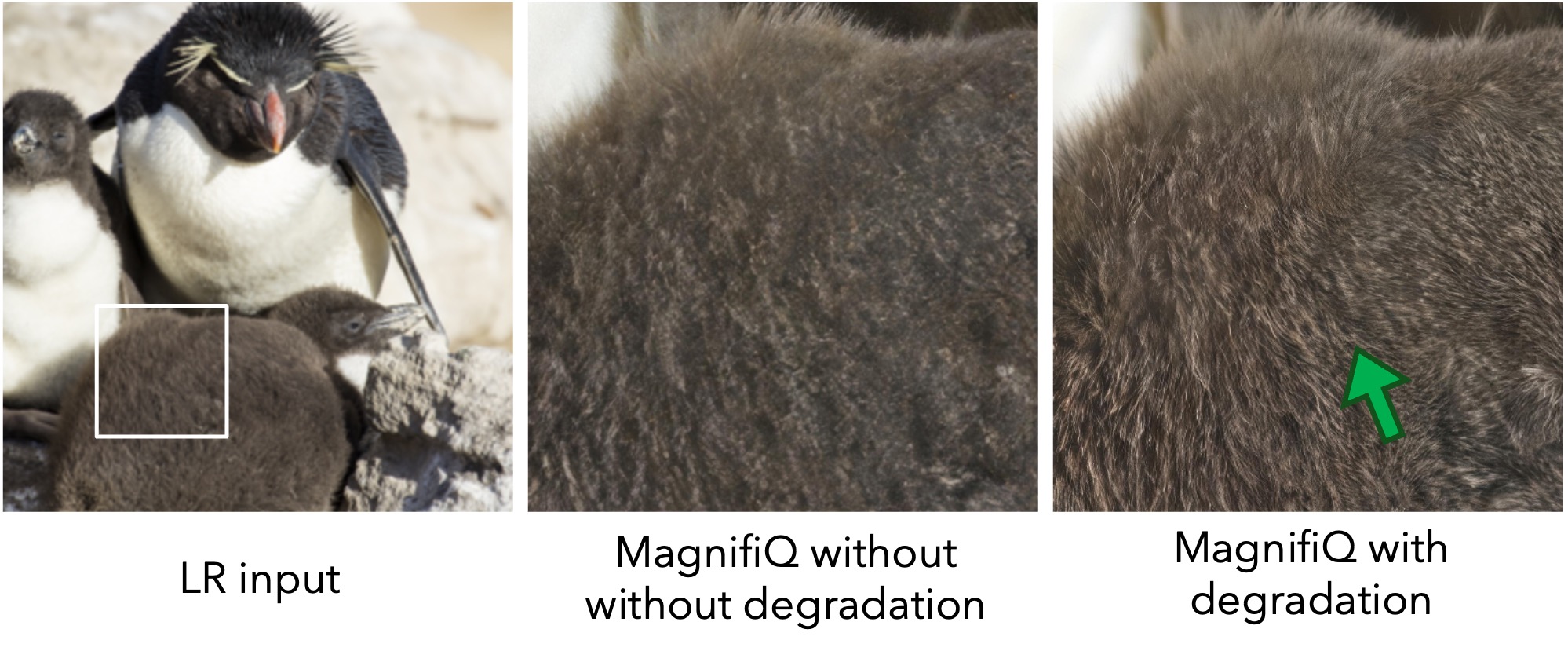}
    \caption{Adding a downsampling–upsampling degradation before each stage ($n > 1$) improves fine-detail restoration. For more details, please refer to Sec.~\ref{sec:abl_configuration}. \textbf{Zoom in to view fine details.}}
    \label{fig:degradations_supp}
\end{figure}

  \paragraph{Progressive stages and degradation.}
  Fig.~\ref{fig:div2k_16x_stages_v1} shows intermediate outputs from the MagnifiQ-\textbf{\blue{C}} configuration ($x_{lr}\rightarrow1K\rightarrow2K\rightarrow3K\rightarrow4K$), demonstrating gradual detail refinement across stages. Fig.~\ref{fig:degradations_supp} further shows that applying downsampling--upsampling degradation before each stage improves fine-
  detail restoration.

\subsection{SDXL-based Restoration Results}
  \label{sec:sdxl_additional_analysis_supp}

\begin{figure}
      \centering
      \includegraphics[width=\linewidth]{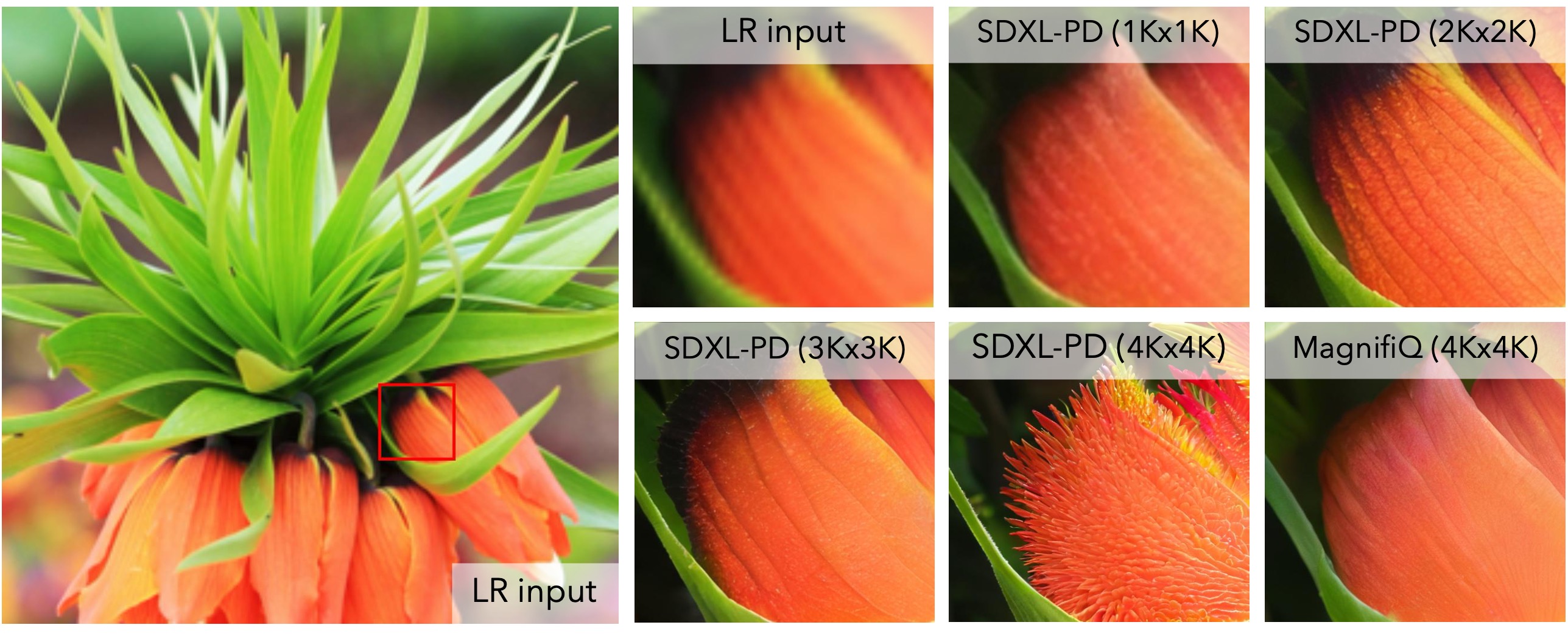}
      \caption{Direct high-resolution restoration with SDXL-PD often creates duplicated textures and texture repetition artifacts. \textbf{Zoom in for details.}}
      \label{fig:duplications_supp}
  \end{figure}

\begin{figure}
    \centering
    \includegraphics[width=0.75\linewidth]{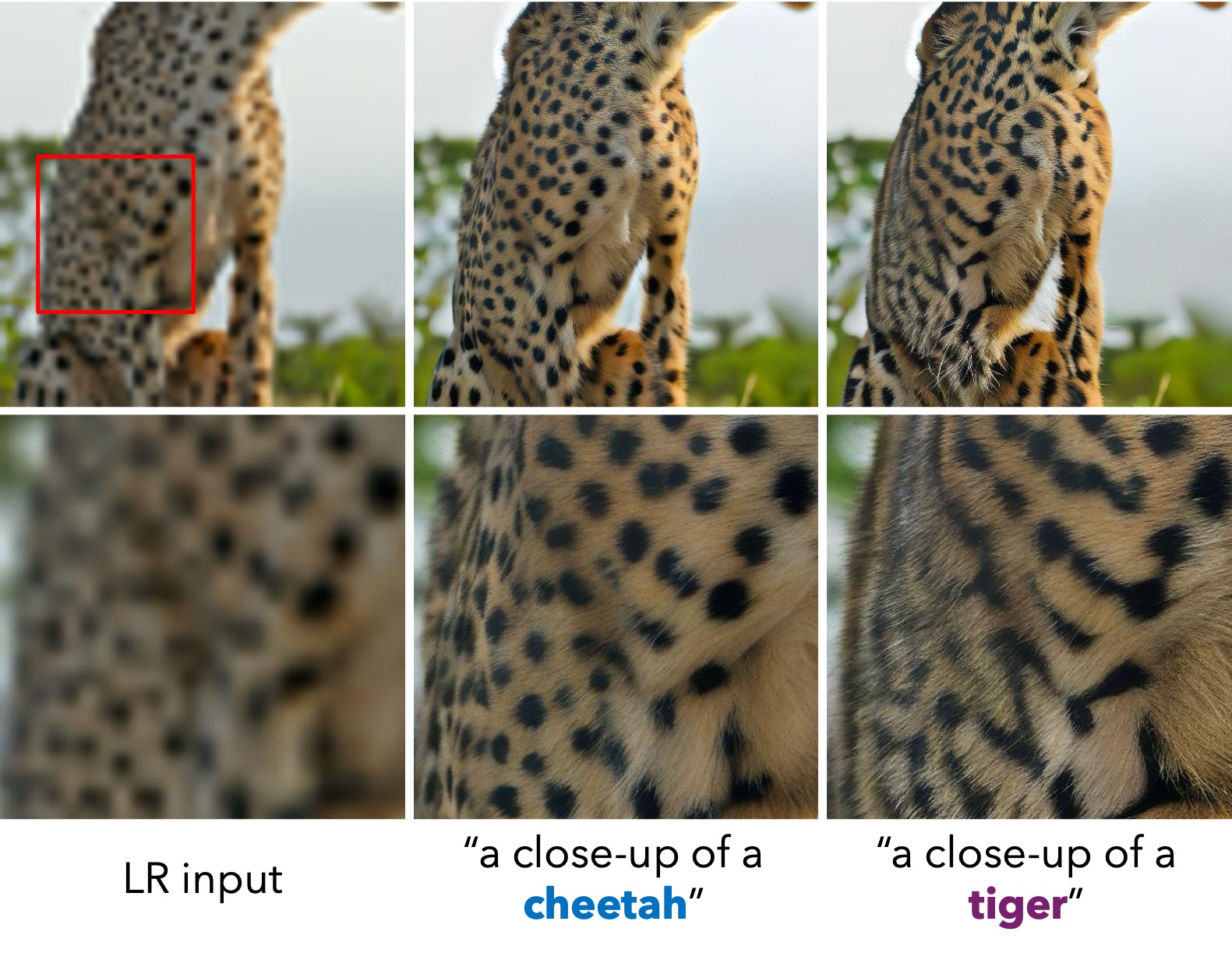}
    \caption{Controllable restoration with SDXL-PD: altering the prompt (e.g., ``cheetah''$\rightarrow$``tiger'') changes the restored textures accordingly. \textbf{Zoom in to view fine details.}}
    \label{fig:restoration_edit_supp}
\end{figure}

\begin{figure*}
    \centering
    \includegraphics[width=0.9\linewidth]{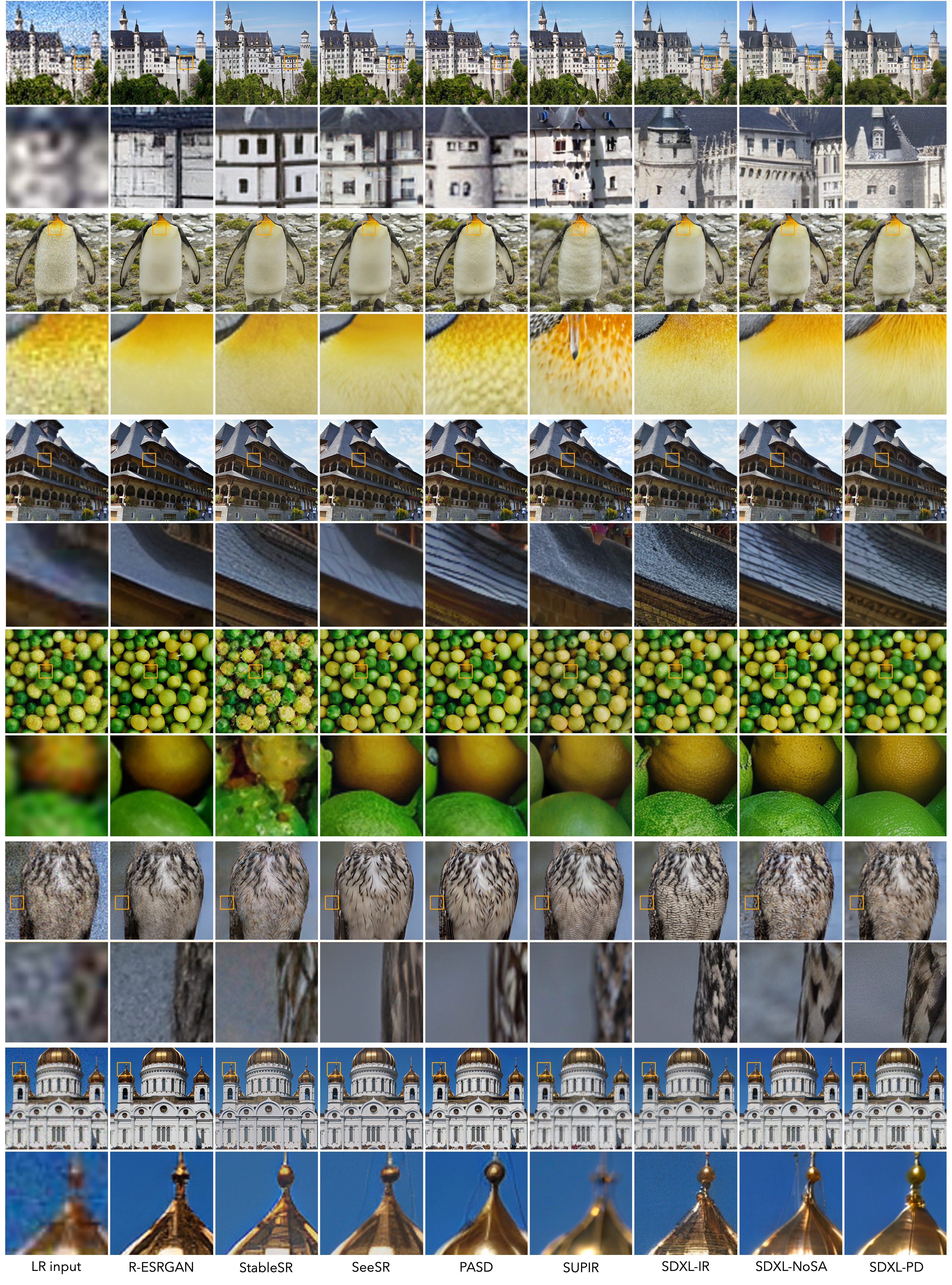}
    \caption{Qualitative results on the DIV2K~\cite{agustsson2017ntire} dataset indicate that SDXL-IR, SDXL-NoSA, and SDXL-PD effectively handle complex degradations, achieving performance comparable to recent state-of-the-art methods that employ more complex architectures.}
    \label{fig:qual_1k_supp}
\end{figure*}

  We provide additional qualitative analysis of the SDXL-based restoration backbone. Fig.~\ref{fig:duplications_supp} shows that direct high-resolution restoration with SDXL-PD can create duplicated texture patterns. Fig.~\ref{fig:restoration_edit_supp} shows controllable restoration, where changing the prompt from ``cheetah'' to ``tiger'' changes the restored
  texture. Fig.~\ref{fig:qual_1k_supp} provides additional DIV2K comparisons, showing that SDXL-PD remains competitive with recent restoration methods.

\subsection{Failure Cases}
  \label{sec:failure_cases_supp}

\begin{figure*}
    \centering
    \includegraphics[width=\linewidth]{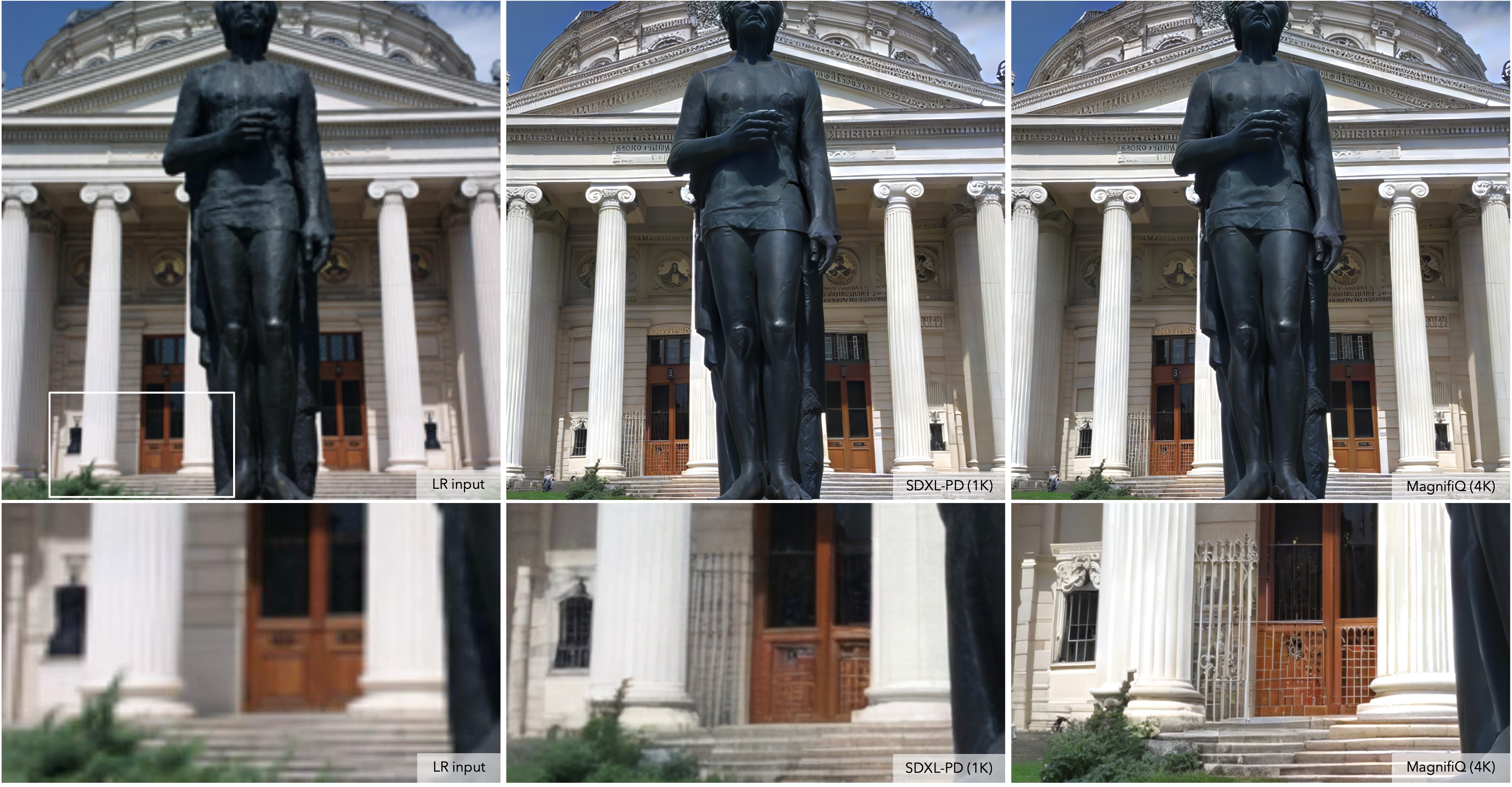}
    \includegraphics[width=\linewidth]{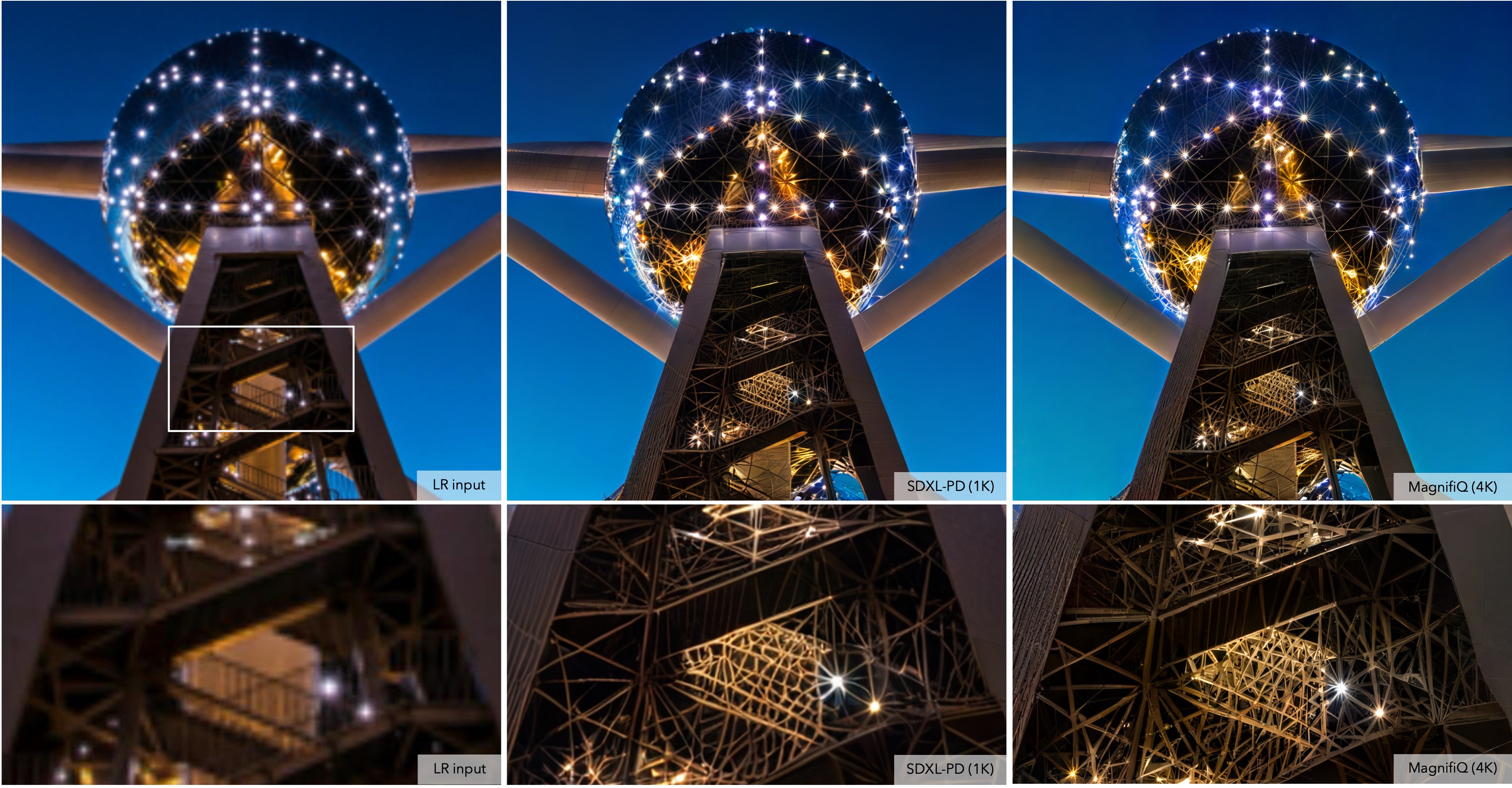}
    \caption{An overview of the failure cases of MagnifiQ in generating hallucinated textures that differ structurally from the input low-quality image is presented. These hallucinations are related to the underlying SDXL-PD image restoration process but are amplified through the progressive upscaling framework. \textbf{Zoom in to view fine details.}}
    \label{fig:limitations}
\end{figure*}

\begin{figure*}
    \centering
    \includegraphics[width=\linewidth]{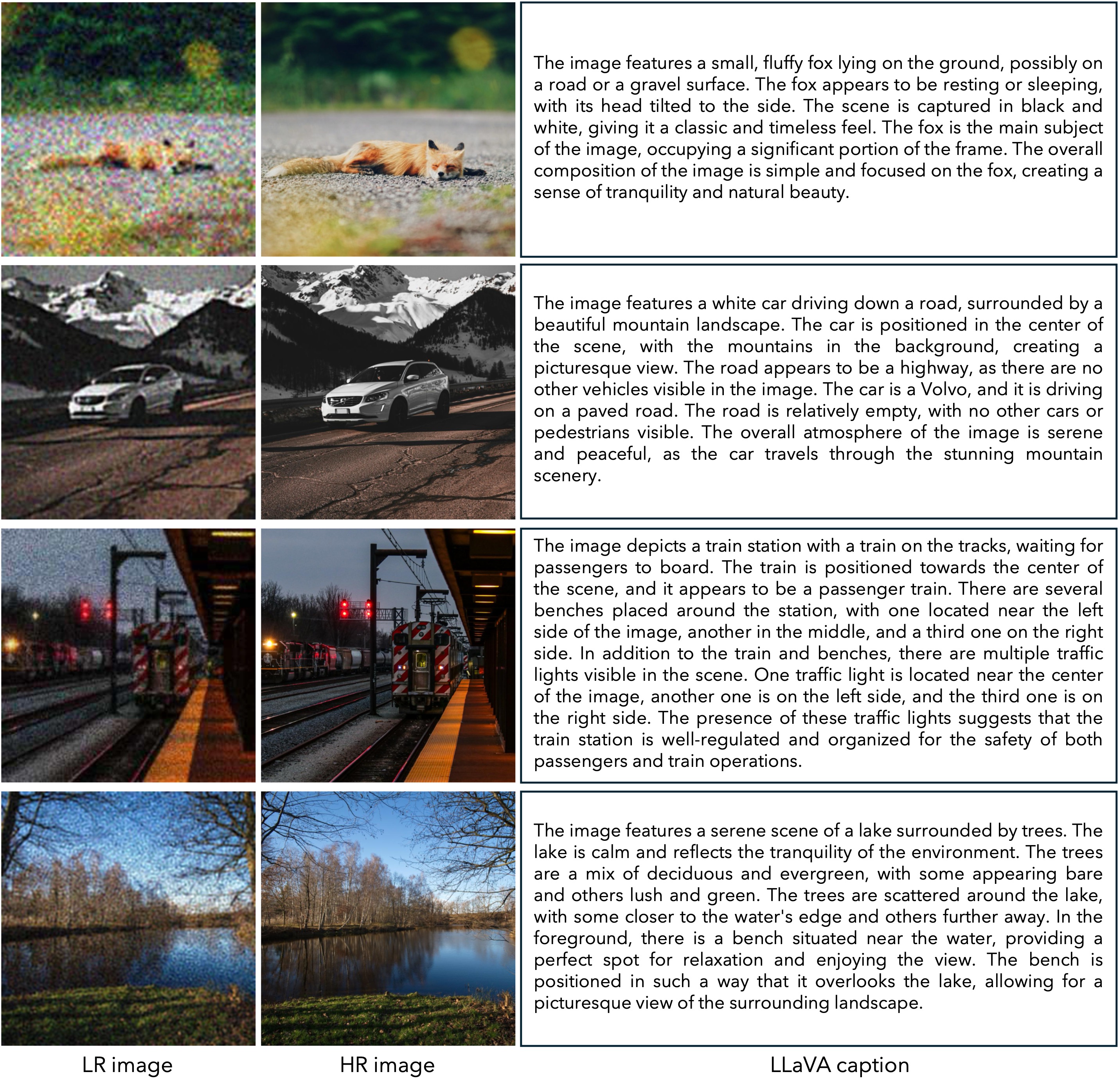}
    \caption{An overview of the training images is presented, including the low-resolution image generated using the degradation pipeline from R-ESRGAN~\cite{wang2021real}, the corresponding ground-truth high-resolution image, and the text caption generated by LLaVA~\cite{liu2024improved}.}
    \label{fig:training_dataset}
\end{figure*}

  In some images containing architectural elements, MagnifiQ can generate hallucinated textures that do not align with the low-quality input, as shown in Fig.~\ref{fig:limitations}. These artifacts originate from the underlying SDXL-PD restoration at stage $n=1$ and can be amplified by progressive upscaling. Future improvements may come from larger and more
  diverse restoration training data or newer restoration backbones such as SD3~\cite{esser2024scaling}.

\clearpage

\section{Prompt Details}
  \label{sec:prompt_details}

\subsection{Global Prompts Used in MagnifiQ}
\label{app:global_prompts_supp}
\begin{itemize}

     \item Fig.~\ref{fig:degradations_supp} - \texttt{The image features a group of penguins gathered on a rocky surface. There are three penguins in the scene, with one penguin standing prominently in the foreground, another penguin situated in the middle, and the third penguin located towards the right side of the image. The penguins are standing close to each other, creating a sense of togetherness and companionship. The rocky surface provides a natural habitat for these birds, and the scene captures the essence of their daily lives.}
    
     \item Fig.~\ref{fig:qual_4k_baselines_supp_1} - \texttt{Majestic mountains rise against a clear sky, their peaks bathed in golden light, while a serene lake reflects their image surrounded by vibrant autumn foliage and mist hovering over the water.}
    
     \item Fig.~\ref{fig:4kagent_div2k50_1} - \texttt{Turquoise lake surrounded by tall, dark green coniferous trees, with snow-capped mountains and misty clouds in the background.}
     \item Fig.~\ref{fig:4kagent_div2k50_2} - \texttt{A towering brown bear stands on its hind legs in a misty forest, surrounded by tall trees and moss-covered rocks, with sunlight filtering through the foliage.}
     \item Fig.~\ref{fig:4kagent_div2k50_4} - \texttt{A bouquet of pink lilies is arranged in a clear vase filled with water, placed on a white side table next to a partially opened curtain and a closed door with vertical paneling, illuminated by soft light.}

     \item Fig.~\ref{fig:limitations}(a) - \texttt{The image features a large statue of a man standing in front of a large building, possibly a church or a museum. The statue is positioned in the center of the scene, drawing attention to its impressive size. The building behind the statue has a prominent dome, adding to the grandeur of the scene. There are several people in the image, with one person standing near the left edge of the frame, another person on the right side, and a third person closer to the center. Additionally, there is a handbag placed on the ground near the center of the scene. The presence of these people suggests that the statue is a point of interest or a focal point in the area.}
     \item Fig.~\ref{fig:limitations}(b) - \texttt{The image features a large, round, illuminated structure, possibly a sphere or a dome, situated in a dark sky. The structure is surrounded by several smaller, illuminated spheres, creating a visually striking scene. The main structure is located towards the center of the image, while the smaller spheres are scattered around it, with some closer to the edges and others further in the background. The combination of the large structure and the smaller spheres creates a sense of depth and intrigue in the scene.}

     \item Fig.~\ref{fig:qual_1k_supp}(a) - \texttt{The image features a majestic castle with a tall tower, situated on a hill overlooking a lake. The castle is surrounded by lush greenery, including trees and bushes, creating a serene and picturesque scene. The castle's architecture is reminiscent of a medieval fortress, with its tall towers and crenelated walls. In the foreground, there is a small boat on the water, adding a sense of depth and scale to the scene. The overall style of the image is a harmonious blend of natural beauty and architectural grandeur, capturing the essence of a bygone era.}
     \item Fig.~\ref{fig:qual_1k_supp}(b) - \texttt{The image features a penguin standing on a rocky surface, possibly a beach or a rocky area. The penguin is the main focus of the scene, and it appears to be looking at the camera. The penguin is positioned towards the center of the image, with its body facing the viewer. The background of the image is characterized by a mix of grass and rocks, creating a natural and rugged environment. The grass can be seen in the foreground, while the rocks are scattered throughout the scene, adding depth and texture to the image.}
     \item Fig.~\ref{fig:qual_1k_supp}(c) - \texttt{The image features a large, old, and ornate building with a distinctive pointed roof. The building is adorned with a variety of potted plants, flowers, and vines, creating a lush and vibrant atmosphere. There are at least 13 potted plants placed throughout the scene, with some situated near the building's entrance and others scattered around the area. In addition to the plants, there are several people in the scene, with one person standing near the left side of the building, another person closer to the right side, and a third person near the center of the image. The presence of people and the beautifully maintained building suggest that this location might be a popular spot for visitors or a place of historical significance.}
     \item Fig.~\ref{fig:qual_1k_supp}(d) - \texttt{The image features a large pile of green and yellow lemons, arranged in a visually appealing manner. The lemons are spread across the entire frame, with some appearing larger and more prominent than others. The green and yellow colors of the lemons create a vibrant and fresh atmosphere. The close-up view of the lemons allows for a detailed examination of their texture and appearance. The overall style of the image is a combination of natural and artistic elements, showcasing the beauty of the lemons in a visually striking manner.}
     \item Fig.~\ref{fig:qual_1k_supp}(e) - \texttt{The image features a close-up of a large, fluffy owl with yellow eyes, sitting on a rock. The owl appears to be looking directly at the camera, capturing the viewer's attention. The owl's feathers are well-defined, showcasing its unique features. The background is blurred, focusing the viewer's attention solely on the owl. The composition of the image is simple and direct, highlighting the owl's striking appearance.}
     \item Fig.~\ref{fig:qual_1k_supp}(f) - \texttt{The image features a large, ornate white church with a gold dome, situated on a hill. The church has a prominent clock tower, and its architecture is reminiscent of a Russian Orthodox church. The church is surrounded by a beautiful garden, and there are several people walking around the area, enjoying the serene atmosphere. In the scene, there are at least 13 people visible, some of them carrying handbags. The church's clock is also visible, adding to the overall charm of the scene. The combination of the church's architectural beauty, the garden, and the people walking around creates a picturesque and peaceful setting.}
     
\end{itemize}

\end{document}